\documentclass[]{kling/kling}

\usepackage{microtype}
\usepackage{graphicx}
\usepackage{subcaption}
\usepackage{csquotes}
\usepackage{afterpage}
\usepackage{xcolor}

\usepackage{amsmath}
\usepackage{amssymb}
\usepackage{mathtools}
\usepackage{amsthm}
\usepackage{xspace}
\usepackage{colortbl}
\usepackage{multirow}
\usepackage{enumitem}
\usepackage{wrapfig}
\usepackage{float}
\usepackage{nicefrac}
\usepackage[font=small,labelfont=bf,justification=centering]{caption}

\usepackage{booktabs}
\usepackage{multirow}
\usepackage{graphicx}
\usepackage{xcolor}
\usepackage{makecell}
\usepackage{pifont} 

\usepackage{geometry}
\usepackage{tabularx}

\usepackage{array}
\newcolumntype{L}[1]{>{\raggedright\arraybackslash}p{#1}}
\newcolumntype{Y}{>{\raggedright\arraybackslash}X}
\definecolor{badbg}{HTML}{FDF3F2}
\definecolor{badborder}{HTML}{C84B31}
\definecolor{badtxt}{HTML}{8C2D19}

\usepackage{tikz}
\usetikzlibrary{arrows.meta,positioning,fit,backgrounds,calc}

\newcommand{\cmark}{\textcolor{green!60!black}{\ding{51}}}%
\newcommand{\xmark}{\textcolor{red}{\ding{55}}}%

\usepackage{amsmath}

\newtcolorbox{observationpanel}[1][]{
  enhanced,
  breakable,
  colback=black!3,
  colframe=black!14,
  boxrule=0.5pt,
  arc=3pt,
  left=10pt,
  right=10pt,
  top=8pt,
  bottom=8pt,
  borderline west={2.2pt}{0pt}{black!55},
  fonttitle=\sffamily\bfseries,
  coltitle=black!80,
  title={Observation},
  #1
}

\newtcolorbox{insightpanel}[1][]{
  enhanced,
  breakable,
  colback=black!3,
  colframe=black!14,
  boxrule=0.5pt,
  arc=3pt,
  left=10pt,
  right=10pt,
  top=8pt,
  bottom=8pt,
  borderline west={2.2pt}{0pt}{black!75},
  fonttitle=\sffamily\bfseries,
  coltitle=black!80,
  title={Insight},
  #1
}

\title{Behind Video Generation: Observations and Insights on Multimodal Understanding from Kling}
\headertext{Multimodal Understanding behind Kling Video Generation}

\author[]{Kling Team}

\abstract{
Video generation has progressed from isolated clips to coherent short-form
audiovisual sequences, and the bottleneck is shifting from rendering to semantic
control. This report summarizes the observations and insights on multimodal
understanding gained from successive generations of Kling. We
argue that generation quality is the projection of multimodal understanding
through the entire data--conditioning--alignment--evaluation chain, and we
examine this chain module by module. Kling-MLLM, a shared multimodal
foundation model that unifies image, video, audio, synchronized audio--video,
and interleaved inputs, is specialized for each downstream module. Because a DiT is trained on far fewer
training samples than a language model, we treat captioning as a semantic
contract that maximizes description density, and extend it to
reference-grounded captioning that makes reference bindings explicit. Prompt
enhancement translates user intent into this caption distribution: basic PE is
derived from captions of compelling videos and refined with text-level
and generator-in-the-loop reinforcement learning, while agentic PE refines the
story and retrieves compelling examples. We show that a reliable text encoder
comes from perceptual training rather than from stronger language ability, and that
candidates can be screened from text alone before end-to-end training. Reward
modeling pairs a general reward model for explainable, localized critiques
with a decision reward model for fast multidimensional scoring, treating
perceptual bandwidth as a first-class design parameter. Finally, we review
task-specific automatic evaluation, focusing on protocols that expose scoring
shortcuts and help trace failures to their upstream source. We hope these lessons offer practical
guidance for building the understanding stack behind controllable video
generation.
}

\metadata[Keywords]{Multimodal understanding, Video generation, Omni captioning, Prompt enhancement, Text encoder, Reward model, Evaluation \& benchmark}
\metadata[Date]{November 3, 2026}

\begin{document}

\maketitle

\vspace{10pt}
\noindent\begin{minipage}{\linewidth}
  \makebox[\linewidth][c]{\includegraphics{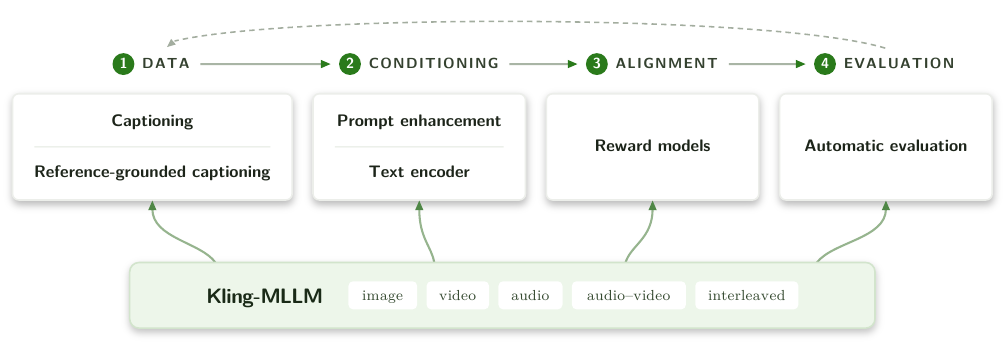}}
  \captionof{figure}{Multimodal understanding along the data--conditioning--alignment--evaluation chain behind Kling.}
  \label{fig:overview}
\end{minipage}

\newpage
{\small\tableofcontents}
\newpage

\section{Introduction}
\label{section:intro}

Over four years, video generation has passed through several distinct regimes.
In 2023, general-purpose video creation remained largely aspirational. In 2024,
early systems produced compelling demonstrations, but with limited reliability
and control. By 2025, leading models could generate production-usable shots and
creative assets. By 2026, frontier systems
had begun to produce usable short-form audiovisual sequences that coordinate
multiple shots, persistent subjects, performance, camera language, dialogue,
music, and sound. The unit by which progress is judged has expanded from the
frame to the clip, then to the shot, and now to the complete sequence.

This expansion changes the technical problem. Quality at the frame or clip level
can often be assessed locally, whereas usefulness at the sequence level is
compositional: constraints interact, errors accumulate over time, and decisions
in one shot affect the interpretation of the next. The system must infer a
latent creative specification from heterogeneous inputs, express it as
conditions the generator can execute, and evaluate the result at the same
semantic granularity. Multimodal understanding has evolved accordingly, from
coarse text--video correspondence to dense spatiotemporal description,
reference grounding, creative planning, and multidimensional judgment, and it
has helped make synthesis controllable and usable.

General-purpose multimodal benchmarks capture only part of this requirement.
High aggregate scores in visual question answering (VQA) do not imply reliable
understanding of camera language, left--right relations, composition, lighting,
shot boundaries, identity binding, or dense temporal change.
These capabilities are weakly represented in standard supervision, yet they determine whether a generated result is
controllable and editable; evaluation research has shifted in the same
direction, from superficial realism toward controllability, commonsense, and
physical faithfulness~\cite{zheng2025vbench2,bansal2025videophy2}. For video
generation, the question is therefore not whether a multimodal large language
model (MLLM) is broadly capable,
but which semantics reach the generator, through what representation, and with
what evidence of downstream benefit.

This perspective is grounded in our work on Kling 1.0 through Kling
4.0~\cite{team2025kling}, together with Kling AI
Avatar~\cite{team2025klingavatar} and Kling Motion
Control~\cite{team2026kling}. Across these releases, we specialized
Kling-MLLM, our shared multimodal foundation model, into captioners, prompt enhancement (PE)
models, text encoders, reward models, and automatic evaluators, as requirements
progressed from basic text--video correspondence to reference binding, identity
preservation, motion transfer, precise editing, audiovisual synchronization, and
complete video generation. As generation capabilities advanced,
progress depended increasingly on understanding what to generate, what to
preserve, how to organize it over time, and how to verify the result. Our
central thesis is:
\begin{quote}
\emph{Video generation quality is the projection of multimodal understanding
through the entire data--conditioning--alignment--evaluation chain.}
\end{quote}
An observed generation failure is not, by itself, a diagnosis. The same symptom
may originate from a caption that omits the relevant supervision, a conditioning
representation that discards temporal structure, a reward model that favors
smooth but static motion, or an evaluator that never tests the requested event.
We therefore treat multimodal understanding as a property of the entire chain
rather than as a score attached to a single model. This perspective leads to
four claims:
\begin{itemize}[leftmargin=*]
    \item \textbf{The bottleneck is shifting from rendering to semantic control.}
    Improvements in visual fidelity do not automatically yield correct actions,
    camera language, reference binding, or narrative continuity.
    Generation-oriented understanding is an independent axis of progress.

    \item \textbf{Data efficiency depends on description density.} A diffusion transformer (DiT) is
    trained on far fewer samples than a language model, and both training
    throughput and the supply of usable video preclude closing this gap
    through scaling. Each training video must therefore carry as much
    supervision as possible, which makes the density of textual description
    per unit of video duration the central measure of data efficiency.
    Dense, generation-oriented captions and reference-grounded captions with
    explicit bindings are designed to maximize this density, and a
    perception-tuned text encoder ensures that it actually reaches the
    generator.

    \item \textbf{A mature DiT has a sweet spot, and agentic methods help reach
    it.} Trained on dense captions, the generator responds best to conditions
    written in its caption distribution, and a mature model can usually realize
    the user's request from a description in this distribution. Basic PE translates
    requests into this distribution, but a single PE model has limited
    capacity and may fail to capture the vividness of compelling videos, while asking it to
    both preserve and repair a contradictory request places conflicting goals
    on one model.
    Agentic PE, such as story refinement that repairs contradictory requests
    and retrieval of captions of compelling videos, divides this work and
    improves reasonableness, faithfulness, and vividness together.

    \item \textbf{Evaluation should follow the task.}
    Because a failure can originate anywhere in the chain, a single aggregate
    score says little about its source. Generation tasks differ in their
    success criteria and in the shortcuts that inflate scores, so evaluation
    protocols should be task-specific and decomposed into directly observable
    criteria. Inexpensive proxies, such as text-only encoder screening and
    text-level PE verifiers, can filter candidates before costly end-to-end
    training. Together, these practices make failures easier to trace upstream.
\end{itemize}

The following sections trace this chain from Kling-MLLM
through standard and reference-grounded captioning, PE, text encoding, and reward
modeling, and conclude with a review of task-specific evaluation practices.
Each module section follows the same discipline: observation, mechanism, and
engineering implication. Our aim is to explain how
advances in multimodal understanding became deployable capabilities across
successive generations of Kling, and which aspects of that experience may
generalize beyond a single model family.

\section{Multimodal Understanding Foundation Models}
\label{sec:mm-understanding-foundation}

\subsection{Objective and Scope}

The multimodal understanding foundation model serves as the shared perception
and reasoning backbone for Kling's image-generation, video-generation,
image-editing, and video-editing tasks.  It is not tied to a single endpoint.
Instead, the foundation checkpoints are specialized through task-specific
post-training for use as captioners, PE models, reward models,
text encoders, and other experts.  This design centralizes visual, temporal,
and acoustic knowledge in a reusable base model while allowing each upstream
task to select an appropriate accuracy--latency operating point.  We refer to
this model as Kling-MLLM; it unifies images, videos, audio, synchronized
audio--video, and arbitrarily interleaved media in one autoregressive
interface.

\subsection{Kling-MLLM Architecture}
\label{subsec:kling-mllm-architecture}

The Kling-MLLM family comprises a dense 27B variant and two mixture-of-experts
(MoE) variants, 35B-A3B and 397B-A17B.  Across scales and development
iterations, the large language model (LLM) and vision backbones are
initialized from compatible checkpoints in the Qwen3.5--Qwen3.8 model
families \cite{qwen2026qwen35}.
Starting from these checkpoints, we add scale-dependent audio towers, described
in \emph{Audio} below.  We also redesign the modality-processing logic for images, videos, audio, synchronized
audio--video, and interleaved media, and conduct multimodal mid-training to
obtain the Kling-MLLM foundation models.

\begin{figure}[htp]
  \centering
  \includegraphics[width=\linewidth]{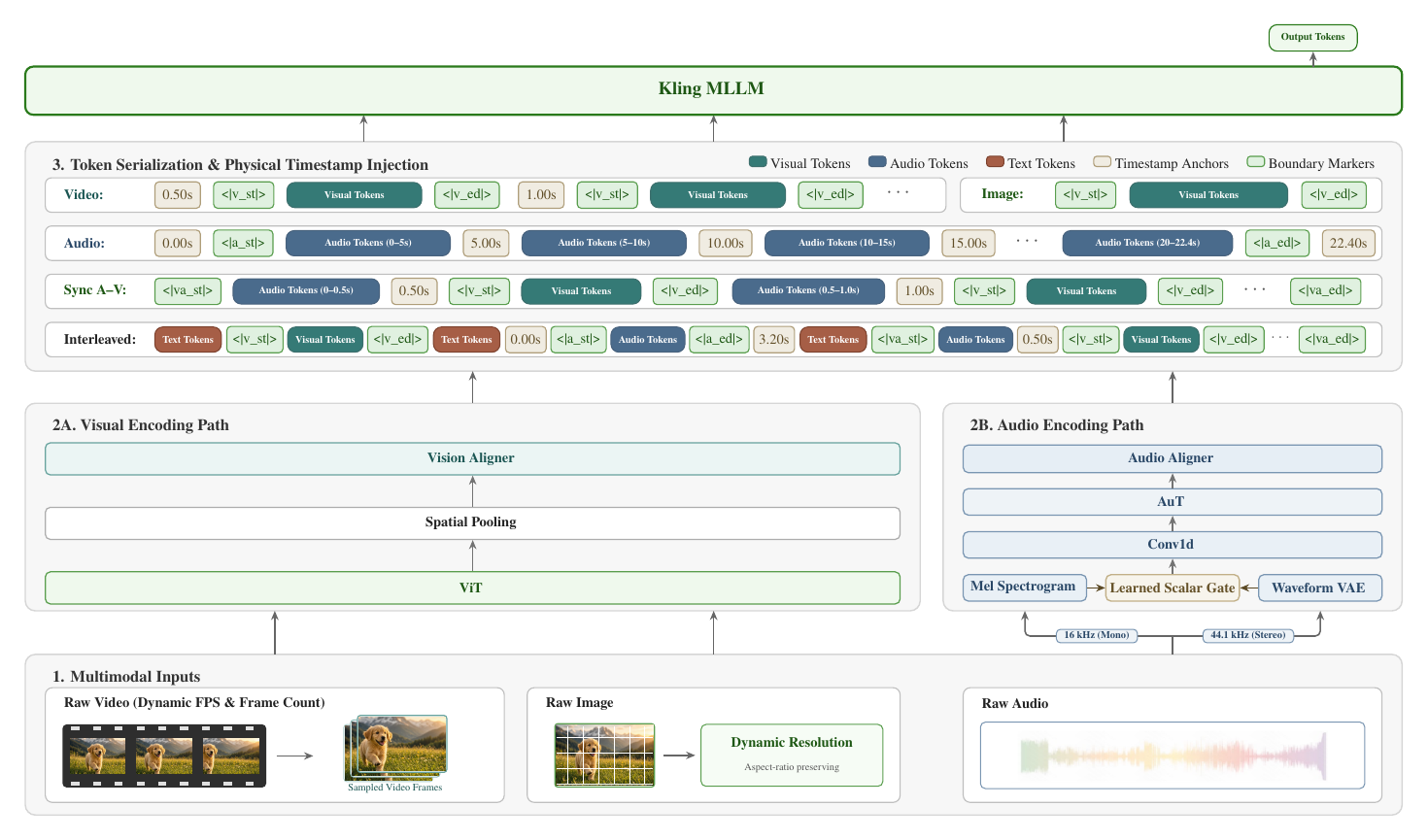}
\caption{Kling-MLLM architecture.  Multimodal inputs are encoded by visual and
audio paths into aligned continuous tokens, delimited by modality boundary
markers, anchored with physical timestamps, and interleaved with text before
entering the LLM backbone.  The audio path illustrates the dual-branch
Mel--VAE tower used in larger variants. `Sync A-V' denotes synchronized audio--video input.}
  \label{fig:kling-mllm-architecture}
\end{figure}

At the input interface, every supported stream is mapped to a common sequence
consumed by a single LLM (Figure~\ref{fig:kling-mllm-architecture}).  The visual path
consists of a ViT, spatial pooling, and a vision aligner.  The audio path uses
a scale-dependent audio tower and projects its output to the LLM hidden width.
Consequently, the LLM can process text, visual embeddings, and acoustic
embeddings in a single sequence without quantizing the input media into a
shared discrete codebook.

\paragraph{Text and interleaved records.}
Text is segmented by the native LLM tokenizer and mapped through its embedding
table.
Modality-specific boundary tokens make each span self-delimiting, so multiple
images, videos, and audio clips can appear at arbitrary positions in one context.

\paragraph{Images and videos.}
Images are processed at dynamic resolution while preserving their aspect
ratios.  The resolution is selected jointly with the available context budget,
allowing the model to retain useful visual detail without requiring a fixed
input shape.

Videos use both a dynamically selected frame rate and frame count.
The processor adapts these choices to the video's duration and the
available context, while sampling across the full temporal range so that short
and long videos can be represented efficiently.  The selected frames follow
the same dynamic-resolution path as images and are encoded into visual-token
blocks.  Each block is associated with its frame's physical timestamp, enabling the
LLM to reason about event order and temporal location without fixed sampling assumptions.

\paragraph{Audio.}
Kling-MLLM uses two scale-dependent audio towers.  For smaller model variants,
the tower follows the Qwen3-Omni design \cite{xu2025qwen3omni}: a 16-kHz
monaural waveform is converted to a Mel spectrogram and encoded as 13 audio
tokens per second.  Larger model variants instead use a dual-branch tower that
fuses Mel and waveform-VAE features.  For each clip, the data pipeline supplies
two time-aligned views: a 16-kHz, 128-bin Mel spectrogram at 100 frames per
second and a 44.1-kHz waveform for an in-house VAE.  A Conv1d front end maps the
Mel stream to the target 25-token-per-second grid.  In parallel, the VAE latent
sequence is linearly interpolated along the temporal axis to match the
Mel-branch token grid, then normalized and projected to the Mel hidden width.
A learned scalar gate adds these aligned VAE features residually to the Mel
features.  Positional
encoding and the audio Transformer operate on the fused sequence, after which
the output projection maps it to LLM-width embeddings
$Z_A\in\mathbb{R}^{L_A\times d}$ at 25 audio tokens per second.  The two tower
designs therefore expose the same embedding interface to the LLM.  The
processor instantiates audio-boundary, timestamp-injection, and synchronized
audio--video serialization at the selected tower's token rate.  Audio boundary
markers delimit each span, and absolute timestamps provide temporal anchors
for long audio sequences.

\noindent\emph{Audio timestamp injection.}
Within each audio branch, the acoustic-token rate is fixed, so elapsed time is
in principle recoverable from token position.  In practice, however, this
implicit clock provides insufficient temporal control: the language model does
not always convert token indices into reliable absolute-time estimates, which
weakens duration awareness and temporal grounding.  Prompting experiments with
Gemini APIs \cite{googledeepmind2026gemini31pro} support this
interpretation: explicitly providing the total audio duration in the prompt
produces a marked improvement in temporal grounding.

The adopted representation places explicit start and end timestamps around
each audio span and inserts absolute-time anchors every five seconds.  The
boundary timestamps expose the clip duration directly, while the periodic
anchors reduce the interval over which time must be inferred and support
fine-grained event localization.  They also align the temporal organization
of audio with synchronized audio--video sequences, where acoustic and visual
tokens share a timestamped timeline.

\paragraph{Synchronized audio--video.}
Audio and video are serialized jointly inside one synchronized multimedia
element, rather than represented as two separate media elements.  Each
processed video frame forms an atomic visual-token block whose tokens share
the frame timestamp, while individual audio tokens retain timestamps
determined by the audio-token rate.  The processor traverses the video frames
chronologically and, before each visual block, inserts all audio tokens that
occur after the preceding frame and no later than the current frame.  Thus,
the original order within each modality is preserved, and an audio token is
placed before a visual block when they have the same timestamp.  The concrete
token arrangement is illustrated in
Figure~\ref{fig:kling-mllm-architecture}.

\subsection{Four-Stage Training Curriculum}
\label{subsec:training-curriculum}

Training follows a curriculum that broadens multimodal supervision and
progressively increases temporal coverage and context capacity.

\paragraph{Stage 1: acoustic initialization.}
The vision tower and LLM are frozen, and only the audio tower is optimized.
Training uses a short sequence length, and the data mixture consists entirely
of audio data.  Its core speech component combines Mandarin, English, and
multilingual transcription data with spoken question answering,
speech-conditioned text continuation, and instruction following.  Non-speech
audio coverage includes environmental and acoustic events, music, non-verbal
vocalizations, speaker and accent attributes, and detailed captioning.  Data
from concatenated speech, multi-speaker dialogues, emotion-rich conversations,
audiobooks, and other long recordings provide longer acoustic contexts.  Thus,
this stage establishes a broad
acoustic representation rather than an ASR-only encoder, without perturbing
the other pretrained components.

\paragraph{Stage 2: joint multimodal training.}
All three modules are unfrozen, and training uses a short sequence length.
Audio and synchronized audio--video data constitute the largest part of the
mixture.  This is the first stage in which the audio tower and LLM are
optimized jointly, so the curriculum deliberately emphasizes audio and
synchronized audio--video data to establish robust acoustic--language alignment and teach the LLM
to integrate temporally synchronized auditory and visual evidence.  Audio
supervision covers multilingual ASR, spoken question answering and instruction
following, environmental sounds, music, timbre, non-verbal vocalizations,
and detailed captioning.  Synchronized audio--video examples add joint
captioning, transcription, and basic temporal-event alignment.

Image, video, text, and other mixed-modality data remain in the mixture at
smaller proportions.  Their role is to preserve the model's visual
perception, language understanding, reasoning, and general multimodal capabilities while
the acoustic pathway is being integrated, thereby mitigating catastrophic
forgetting.  These retained data include dense image and video captioning,
edit-instruction inference, multi-image composition, counting, OCR, document
and chart understanding, spatial reasoning, and educational and scientific
text.

\paragraph{Stage 3: temporal-capacity expansion.}
All modules remain trainable, and training transitions to a long sequence
length.  The mixture shifts toward audio, video, and audio--video data.  Its
objective is to improve fine-grained temporal perception.  The video sources
span movie and general video question answering, timestamped procedural steps,
instruction-guided temporal grounding, multi-shot narrative understanding,
spatial reasoning, unusual or anomalous action recognition, and
playback-speed-sensitive motion.  Audio training continues to include
multilingual ASR, spoken instructions, sound and music captioning, and audio
question answering;
synchronized examples add timestamped audiovisual event alignment.  Counting,
OCR, document and chart understanding remain in the mixture alongside
multilingual instruction, mathematical, and multi-domain reasoning data.
Additional supervision covers detailed media captioning, shot-boundary
understanding, visual artifact localization, and reference-conditioned
generation-prompt rewriting.

\paragraph{Stage 4: long-context consolidation.}
Both modality towers are frozen and only the LLM is updated.  This stage also
uses a long sequence length, allowing the model to focus on long-form
integration while retaining the frozen modality encoders.
The highest-quality data available for every supported modality are placed in
this stage, allowing the final LLM update to consolidate capabilities using
the cleanest and most reliable supervision.
Long-video data provide supervision for retrieval, question answering,
multi-span temporal grounding, reference-conditioned search,
fine-grained motion, and evidence aggregation over extended timelines.
Long-form audio receives parallel treatment through event localization,
interval-conditioned description, timestamped summarization, temporal-event
reasoning, spoken QA, and soundscape segmentation.  Joint audio--visual data
emphasize synchronized description, detail recovery, cross-stream reasoning,
and constraint-following, while interleaved audio--image examples require
evidence to be composed across modalities.  Visual logic, science, counting,
text recognition, document/chart understanding, and general text reasoning
remain in the mixture.  The capstone stage is designed to help the LLM
organize fixed perceptual features over longer evidence chains and return
grounded, structured responses.

The supervision mix is likewise adjusted by stage.  Captioning is combined
with recognition, retrieval, localization, structured extraction, question
answering, and both direct-response and reasoning-oriented examples.  Several
large sources are partitioned across successive stages, preserving selected
speech, OCR/document, chart, video, and reasoning coverage as new task
families are introduced.  The curriculum is therefore a progression in both
modality coupling and task diversity, with increasing temporal and contextual
demands, rather than a sequence of four disjoint corpora.  We also explore model merging and
find that merging the Stage~1 and Stage~4 checkpoints yields a better
checkpoint than either alone.

\subsection{Data Construction}
\label{subsec:data-construction}

The training corpus combines public data, Kling-specific task data, and
agent-oriented data.  At the system-design level, heterogeneous
sources are represented through a common conversation-oriented schema, with
media metadata and modality boundaries retained so that one loader
can construct image, video, audio, synchronized audio--video, and interleaved
examples, following common practice in
omni-modal systems~\cite{xu2025qwen3omni}.

\paragraph{Public Data Ingestion}

Public corpora enter a three-part workflow:
\emph{AutoResearch} $\rightarrow$ \emph{Auto Data Process} $\rightarrow$
\emph{Auto Revise}, with revised records looping back from Auto Revise to Auto
Data Process for revalidation.  In AutoResearch, one search agent monitors
Hugging Face Daily Papers, filters work on text,
image, video, audio,
audio--video, and multimodal foundation models, and extracts candidate
datasets from the selected papers.  A second agent monitors trending Hugging
Face datasets.  Their results are merged and deduplicated both against one
another and against the historical registry.  During dataset selection, we
also exclude datasets containing sensitive material, datasets whose licenses
are incompatible with the intended use, and datasets that overlap with any
evaluation set.  Papers associated with the remaining candidates are routed
for review, while the selected new datasets enter automated processing.

Auto Data Process downloads each selected dataset and invokes a
modality-specific conversion skill for text, image, video, audio,
audio--video, or interleaved-media records.  Table~\ref{tab:modality-data-pipelines} summarizes this interface.
\begin{table*}[htp]
  \centering
  \caption{High-level normalization and supervision pipelines by modality.}
  \label{tab:modality-data-pipelines}
  \footnotesize
  \renewcommand{\arraystretch}{1.20}
  \setlength{\tabcolsep}{6pt}
  \begin{tabularx}{\textwidth}{@{}L{0.16\textwidth} L{0.44\textwidth} Y@{}}
    \toprule
    \rowcolor{klingbg}
    \textbf{Modality} & \textbf{Conversion-time normalization and validation} &
      \textbf{Representative supervision} \\
    \midrule
    \textbf{Text} &
      Encoding \& language checks, document deduplication, conversational turn normalization &
      Next-token prediction, instruction following, reasoning \\
    \rowcolor{black!2}
    \textbf{Image} &
      Decode \& resolution checks, perceptual deduplication, aspect-ratio handling, OCR/metadata extraction &
      Short \& dense captions, VQA, OCR, grounding, counting, pose \\
    \textbf{Video} &
      Decode check, shot segmentation, timestamped frame selection, optional OCR; audio routed to sync branch &
      Temporal captioning, event \& motion QA, camera understanding, multi-shot reasoning \\
    \rowcolor{black!2}
    \textbf{Audio} &
      Waveform/channel normalization, segmentation, VAD \& signal-quality checks, transcript alignment &
      ASR, audio captioning \& QA, acoustic-event reasoning, temporal grounding \\
    \textbf{Audio--video} &
      Synchronization validation, shared timestamped token timeline construction &
      Joint transcription, cross-modal QA, grounding, captioning \\
    \rowcolor{black!2}
    \textbf{Interleaved media} &
      Document/dialogue order preservation, media-reference checking, cross-record deduplication &
      Contextual QA \& reasoning over multiple heterogeneous media items \\
    \bottomrule
  \end{tabularx}
\end{table*}

Auto Revise progressively narrows the audit set.  A coarse pass samples
questions to assess formatting and question diversity.  The first fine pass
computes perplexity with Qwen3-Omni, and the second samples multimodal content
and question--answer pairs for review by Gemini.  The resulting error report
undergoes human review.  Accepted data are assigned to an appropriate training
stage; low-quality data are removed or repaired, with repaired data returning
to Auto Data Process for revalidation.

\paragraph{Kling-Specific Task Data}

Internal task data target the perception and reasoning capabilities required
by Kling's image and video generation and editing systems.  They cover world
knowledge through interleaved image--text records and a controlled taxonomy of
actions, sounds, subjects, and scenes; image perception through bilingual OCR
across diverse text appearances and layouts, counting, grounding,
metadata-aware description, and pose understanding; dense image and video
captioning of scenes, subjects, motion, style, camera behavior,
and multi-shot structure; and editing and reference understanding through
paired source--target images or videos, instructions for addition, deletion,
and modification, and multi-reference conditioning.

\paragraph{Multimodal Agent Data}

The agent-data mixture spans both general-purpose tool use and multimodal
perception-to-action tasks.  General-agent examples cover multi-step API and
transactional workflows, search and evidence-grounded web research,
software-engineering and terminal interaction, structured reasoning and data
transformation, and safety-aware everyday assistance.  Their supervision
ranges from complete tool-use loops to partial trajectories for next-action
selection and direct responses, training the model to plan dependent actions,
interpret tool returns, verify intermediate results, and decide when tool use
is necessary.  Multimodal-agent examples include image-grounded retrieval and
visual reasoning, GUI interaction grounded in screen observations, agentic
video understanding that searches and reasons over temporally distributed
evidence, and Blender-based tasks involving controllable 3D scenes, objects,
motion, camera viewpoints, and rendering.  These data may be constructed or
augmented with model-generated instructions, trajectories, and responses
followed by consistency checks.

\subsection{Evaluation}
\label{subsec:evaluation}

\providecommand{\oursofficial}[2]{#1\,/\allowbreak\,#2}
\providecommand{\na}{--}

Evaluation covers five input settings: audio, synchronized audio--video,
multimodal interleaving, image, and video.  Tables~\ref{tab:audio}--\ref{tab:video}
combine the corresponding public and Kling-specific benchmarks.  We use the
LMMs-Eval framework
\cite{zhang2024lmmsevalrealitycheckevaluation}; when a benchmark
is already integrated into LMMs-Eval, we use its default evaluation prompt.
Arrows indicate metric direction, and ``--'' denotes an unavailable score.
For open-weight models, paired values list our evaluation result followed by the model
developer's reported result; a single value means that no directly comparable
official score is included.  For LATBench, values separated by ``\texttt{|}''
denote Chinese and English results, respectively; for VidCapBench, they denote
Accuracy and Coverage, respectively.
Boldface marks the best internally evaluated value in each metric; ties are
bolded jointly.

\begin{table}[hbp]
  \centering
  \caption{Audio benchmarks. Paired values
  normally list our evaluation result followed by the model developer's reported result;
  for LATBench, values separated by ``\texttt{|}'' denote Chinese and English
  results, respectively.}
  \label{tab:audio}
  \scriptsize
  \setlength{\tabcolsep}{2.5pt}
  \renewcommand{\arraystretch}{1.18}
  \resizebox{\linewidth}{!}{%
  \begin{tabular}{@{}
    >{\raggedright\arraybackslash}p{4.0cm}
    *{9}{>{\centering\arraybackslash}c}@{}}
    \toprule
    \multirow{2}{*}{\textbf{Model}} &
      \multicolumn{2}{c}{\textbf{LibriSpeech}} &
      \multirow{2}{*}{\textbf{WorldSense $\uparrow$}} &
      \multirow{2}{*}{\textbf{TREA $\uparrow$}} &
      \multicolumn{5}{c}{\textbf{LATBench}} \\
    \cmidrule(lr){2-3}\cmidrule(lr){6-10}
    & \textbf{clean $\downarrow$} & \textbf{other $\downarrow$}
    & & & \textbf{mIoU $\uparrow$} & \textbf{R@0.3 $\uparrow$}
    & \textbf{R@0.5 $\uparrow$} & \textbf{R@0.7 $\uparrow$} & \textbf{FENSE $\uparrow$} \\
    \midrule
    Qwen3-Omni-30B-A3B-Instruct &
      \oursofficial{1.32}{1.22} & \oursofficial{2.66}{2.48} &
      46.03 & 71.0 & 14.8\textbar15.8 & 21.4\textbar26.4 &
      12.4\textbar16.0 & 7.0\textbar7.0 & 28.4\textbar31.0 \\
    MiniCPM-o-4.5 &
      \oursofficial{1.54}{1.4} & \oursofficial{3.79}{2.8} &
      44.99 & 54.8 & 3.0\textbar5.2 & 4.0\textbar7.4 & 2.1\textbar3.7 &
      0.8\textbar2.0 & 41.1\textbar41.9 \\
    Qwen3.5-Omni-Plus (DashScope) &
      1.32 & 3.14 & 63.90 & 87.7 &
      60.2\textbar51.2 & 81.1\textbar71.1 & 63.2\textbar54.5 &
      40.5\textbar33.8 & 64.2\textbar53.5 \\
    Doubao2.0-Lite (non-thinking) &
      1.28 & \textbf{2.41} & 53.94 &
      71.5 & 53.1\textbar53.2 & 73.9\textbar73.2 &
      58.9\textbar57.9 & 38.2\textbar37.2 & 57.4\textbar48.7 \\
    Doubao2.0-Lite (thinking, high) &
      1.35 & 2.55 & \textbf{66.77} &
      79.2 & 52.6\textbar55.1 & 72.4\textbar77.5 &
      59.1\textbar61.9 & 39.0\textbar37.8 & 58.7\textbar42.1 \\
    \midrule
    Kling-MLLM-35B-A3B &
      1.68 & 3.30 & 61.92 &
      93.0 & 58.8\textbar51.4 & 81.8\textbar71.2 &
      61.3\textbar55.1 & 38.7\textbar35.4 &
      66.5\textbar\textbf{72.0} \\
    Kling-MLLM-27B &
      1.32 & 2.93 & 61.79 &
      \textbf{93.2} & 61.0\textbar\textbf{62.7} &
      82.0\textbar\textbf{87.7} & 66.2\textbar\textbf{71.9} &
      42.2\textbar\textbf{43.8} & 67.6\textbar70.5 \\
    Kling-MLLM-397B-A17B &
      \textbf{1.18} & 2.71 & 63.56 &
      91.2 & \textbf{65.0}\textbar61.7 &
      \textbf{86.4}\textbar85.4 & \textbf{68.1}\textbar68.2 &
      \textbf{47.4}\textbar43.4 & \textbf{70.8}\textbar68.9 \\
    \bottomrule
  \end{tabular}}
\end{table}

\paragraph{Audio.}
Table~\ref{tab:audio} combines LibriSpeech
\cite{panayotov2015librispeech}, WorldSense
\cite{hong2026worldsenseevaluatingrealworldomnimodal}, TREA \cite{bhattacharya2025benchmarking}, and LAT-Bench
\cite{shao2026listeningtimeprecisetemporal}.  Together they cover speech
recognition, audio understanding, and temporal grounding.
Kling-MLLM-397B-A17B gives the lowest LibriSpeech clean WER and the strongest
Chinese LATBench results, while Kling-MLLM-27B gives the strongest TREA and
English LATBench temporal-grounding results.  Kling-MLLM-35B-A3B gives the
strongest English LATBench FENSE result.  The comparison remains mixed on the other
columns: Doubao2.0-Lite gives the best LibriSpeech other and WorldSense scores.

\begin{table}[htbp]
  \centering
  \caption{Audio--video and multimodal-interleaving benchmarks.}
  \label{tab:audio-video}
  \scriptsize
  \setlength{\tabcolsep}{2.5pt}
  \renewcommand{\arraystretch}{1.18}
  \resizebox{\linewidth}{!}{%
  \begin{tabular}{@{}
    >{\raggedright\arraybackslash}p{4.0cm}
    *{9}{>{\centering\arraybackslash}c}@{}}
    \toprule
    \multirow{2}{*}{\textbf{Model}} &
      \multirow{2}{*}{\textbf{OmniBench $\uparrow$}} &
      \multicolumn{3}{c}{\textbf{video-SALMONN}} &
      \multirow{2}{*}{\textbf{UGC-VideoCap}} &
      \multicolumn{2}{c}{\textbf{OmniCap-IF}} &
      \multicolumn{2}{c}{\textbf{Multimodal interleaving}} \\
    \cmidrule(lr){3-5}\cmidrule(lr){7-8}\cmidrule(lr){9-10}
    & & \textbf{Miss $\downarrow$} & \textbf{Hallucination $\downarrow$} &
      \textbf{Total error $\downarrow$} & &
      \textbf{CSR $\uparrow$} & \textbf{ISR $\uparrow$} &
      \textbf{IV-Bench $\uparrow$} & \textbf{MVU-Eval $\uparrow$} \\
    \midrule
    Qwen3-Omni-30B-A3B-Instruct &
      60.35 & 36.32 & 16.87 & 53.20 & 80.01 & 65.87 & 9.06 &
      35.02 & 56.41 \\
    MiniCPM-o-4.5 &
      52.04 & 33.37 & \textbf{15.35} & 48.72 & 78.45 &
      70.84 & 11.69 & 24.70 & 47.81 \\
    Qwen3.5-Omni-Plus (DashScope) &
      67.61 & \textbf{17.47} & 21.67 & 39.13 &
      87.78 & 75.90 & 18.95 & 53.31 & 60.64 \\
    Doubao2.0-Lite (non-thinking) &
      57.79 & 28.20 & 17.26 & 45.46 & 84.22 & 76.56 & 20.29 &
      43.48 & 60.42 \\
    Doubao2.0-Lite (thinking, high) &
      \textbf{70.18} & 19.47 & 19.97 & 39.45 &
      \textbf{91.05} & 78.21 & 21.62 &
      \textbf{61.41} & \textbf{63.71} \\
    \midrule
    Kling-MLLM-35B-A3B &
      61.15 & 21.63 & 18.55 & 40.19 & 85.65 & 81.02 &
      25.39 & 54.23 & 61.18 \\
    Kling-MLLM-27B &
      62.57 & 22.47 & 18.02 & 40.49 & 84.61 & 81.55 & 26.06 &
      57.31 & 61.02 \\
    Kling-MLLM-397B-A17B &
      69.38 & 20.00 & 17.07 &
      \textbf{37.07} & 88.21 & \textbf{82.75} &
      \textbf{28.13} & 61.05 & 63.16 \\
    \bottomrule
  \end{tabular}}
\end{table}

\paragraph{Audio--video.}
Table~\ref{tab:audio-video} reports synchronized audio--video and multimodal
interleaving results on OmniBench \cite{li2025omnibenchfutureuniversalomnilanguage},
video-SALMONN \cite{sun2024videosalmonnspeechenhancedaudiovisuallarge},
UGC-VideoCap \cite{wu2025ugcvideocaptioneromniugcvideo}, OmniCap-IF
\cite{wang2026omnicapifbenchmarkingimprovinginstruction}, IV-Bench
\cite{ma2025ivbenchbenchmarkimagegroundedvideo}, and MVU-Eval
\cite{peng2025mvuevalmultivideounderstandingevaluation}.  On video-SALMONN,
Kling-MLLM-397B-A17B has the lowest total error, Qwen3.5-Omni-Plus has the lowest miss
rate, and MiniCPM-o-4.5 has the lowest hallucination rate.  Doubao2.0-Lite (thinking,
high) leads UGC-VideoCap, while Kling-MLLM-397B-A17B leads both OmniCap-IF
metrics.  Its OmniBench audio-transcript score is slightly below the strongest
reported result.

\paragraph{Multimodal interleaving.}
Kling-MLLM-397B-A17B achieves the strongest results among the Kling-MLLM
variants on both interleaving benchmarks.  It is particularly competitive
on IV-Bench and MVU-Eval, reaching 61.05 and 63.16, respectively---within 0.36
and 0.55 points of the best reported results.  This consistent performance
indicates strong cross-modal evidence integration across interleaved inputs.

\begin{table}[htbp]
  \centering
  \caption{Image benchmarks.}
  \label{tab:image}
  \scriptsize
  \setlength{\tabcolsep}{2.5pt}
  \renewcommand{\arraystretch}{1.18}
  \begin{tabular}{@{}
    >{\raggedright\arraybackslash}p{4.0cm}
    *{8}{>{\centering\arraybackslash}c}@{}}
    \toprule
    \multirow{2}{*}{\textbf{Model}} &
      \multicolumn{2}{c}{\textbf{MMBench}} &
      \multirow{2}{*}{\textbf{OCRBench $\uparrow$}} &
      \multicolumn{4}{c}{\textbf{TextPecker}} &
      \multirow{2}{*}{\textbf{PerceptionBench $\uparrow$}} \\
    \cmidrule(lr){2-3}\cmidrule(lr){5-8}
    & \textbf{EN (dev) $\uparrow$} & \textbf{CN (dev) $\uparrow$} & &
      \textbf{EN\textsuperscript{+} $\uparrow$} &
      \textbf{EN\textsuperscript{-} $\downarrow$} &
      \textbf{CN\textsuperscript{+} $\uparrow$} &
      \textbf{CN\textsuperscript{-} $\downarrow$} & \\
    \midrule
    Qwen3-Omni-30B-A3B-Instruct &
      87.29 & 83.68 & 86.80 & 0.032 & 0.078 & 0.047 & 0.041 & 32.17 \\
    MiniCPM-o-4.5 &
      85.22 & 82.99 & \oursofficial{61.00}{87.6} &
      0.094 & 0.099 & 0.024 & 0.085 & 29.37 \\
    Gemma-4-31B-IT &
      87.11 & 87.11 & 84.70 & 0.252 & \textbf{0.054} & 0.187 &
      0.072 & 47.23 \\
    Qwen3-VL-32B &
      87.63 & 86.17 & \oursofficial{84.80}{89.5} &
      0.159 & 0.166 & 0.171 & 0.069 & 33.97 \\
    Qwen3.5-Omni-Plus (DashScope) &
      91.24 & 88.75 & 85.20 & 0.106 & 0.060 & 0.072 & 0.038 & 37.50 \\
    Qwen3.5-397B-A17B &
      \oursofficial{90.03}{93.7} & 89.09 &
      \oursofficial{83.70}{93.1} &
      0.187 & 0.078 & 0.087 & 0.044 & 47.47 \\
    Doubao2.0-Lite (non-thinking) &
      89.60 & 87.54 & 82.90 & 0.430 & 0.071 & 0.347 & 0.045 & 44.13 \\
    Doubao2.0-Lite (thinking, high) &
      \textbf{91.84} & 89.26 & 84.50 &
      0.453 & 0.078 & 0.371 & 0.049 & \textbf{53.93} \\
    \midrule
    Kling-MLLM-35B-A3B &
      88.40 & 86.94 & 84.90 &
      0.649 & 0.092 & 0.774 & 0.053 & 39.00 \\
    Kling-MLLM-27B &
      87.29 & 87.54 & 83.10 & 0.536 & 0.103 & 0.702 & 0.049 & 40.90 \\
    Kling-MLLM-397B-A17B &
      90.55 & \textbf{89.52} & \textbf{87.20} &
      \textbf{0.660} & 0.090 &
      \textbf{0.782} & \textbf{0.036} & 48.63 \\
    \bottomrule
  \end{tabular}
\end{table}

\paragraph{Image.}
Table~\ref{tab:image} combines MMBench
\cite{liu2024mmbenchmultimodalmodelallaround}, OCRBench
\cite{liu2024ocrbenchhiddenmysteryocr}, TextPecker
\cite{zhu2026textpeckerrewardingstructuralanomaly}, and PerceptionBench
\cite{lin2026perceptionbenchevaluatingatomicvisual}, covering image
perception, OCR, and text-rendering assessment.  Kling-MLLM-397B-A17B leads
MMBench-CN, OCRBench, and three of the four TextPecker measures, while
Doubao2.0-Lite (thinking, high) leads MMBench-EN and PerceptionBench.
Gemma-4-31B-IT leads TextPecker EN$^{-}$.

\begin{table}[htp]
  \centering
  \caption{Video benchmarks. For VidCapBench, values separated by
  ``\texttt{|}'' denote Accuracy and Coverage.}\vspace{-2pt}
  \label{tab:video}
  \scriptsize
  \setlength{\tabcolsep}{1.1pt}
  \renewcommand{\arraystretch}{1.18}
  \resizebox{\linewidth}{!}{%
  \begin{tabular}{@{}
    >{\raggedright\arraybackslash}p{4.0cm}
    *{14}{>{\centering\arraybackslash}c}@{}}
    \toprule
    \multirow{2}{*}{\textbf{Model}} &
      \multirow{2}{*}{\makecell{\textbf{LongVideo}\\\textbf{Bench $\uparrow$}}} &
      \multirow{2}{*}{\textbf{MVBench $\uparrow$}} &
      \multirow{2}{*}{\makecell{\textbf{MME-}\\\textbf{VideoOCR $\uparrow$}}} &
      \multirow{2}{*}{\makecell{\textbf{VSI-}\\\textbf{Bench $\uparrow$}}} &
      \multirow{2}{*}{\makecell{\textbf{ShortVid-}\\\textbf{Bench $\uparrow$}}} &
      \multirow{2}{*}{\makecell{\textbf{DSI-}\\\textbf{Bench $\uparrow$}}} &
      \multirow{2}{*}{\makecell{\textbf{Event-}\\\textbf{Bench $\uparrow$}}} &
      \multirow{2}{*}{\makecell{\textbf{Motion-}\\\textbf{Bench $\uparrow$}}} &
      \multicolumn{4}{c}{\textbf{TimeLens}} &
      \multirow{2}{*}{\makecell{\textbf{CameraMotion}\\\textbf{VQA $\uparrow$}}} &
      \multirow{2}{*}{\textbf{VidCapBench $\uparrow$}} \\
    \cmidrule(lr){10-13}
    & & & & & & & & & \textbf{Overall $\uparrow$} &
      \textbf{Charades $\uparrow$} & \textbf{QVHighlights $\uparrow$} &
      \textbf{ActivityNet $\uparrow$} & & \\
    \midrule
    Qwen3-Omni-30B-A3B-Instruct &
      64.70 & 70.00 & 62.35 & 61.12 & 63.90 & 51.95 & 69.91 &
      61.57 & 12.64 & 14.90 & 8.77 & 14.26 & 31.68 &
      17.78\textbar84.24 \\
    MiniCPM-o-4.5 &
      \oursofficial{64.85}{66.0} & 64.00 & 62.20 & 50.38 & 64.60 &
      26.23 & 66.48 & \oursofficial{60.75}{61.4} & 10.01 &
      9.21 & 9.62 & 11.19 & 23.53 & 15.22\textbar79.16 \\
    Gemma-4-31B-IT &
      61.78 & 64.40 & 68.25 & 48.10 & 68.50 & 55.23 & 65.11 &
      61.05 & 51.98 & 46.66 & 61.80 & 47.49 & 28.09 &
      16.64\textbar90.17 \\
    Qwen3-VL-32B &
      67.61 & \oursofficial{73.65}{72.8} & 75.25 &
      \oursofficial{64.16}{61.5} & 67.30 & 53.36 & 74.89 & 67.32 &
      52.16 & 52.79 & 55.71 & 47.99 & 26.71 & 20.01\textbar87.49 \\
    Qwen3.5-Omni-Plus (DashScope) &
      69.26 & \textbf{80.37} & 79.20 & 60.67 & 70.10 & 59.36 & 76.62 & 71.55 &
      33.31 & 43.14 & 25.92 & 30.88 & 37.70 &
      18.29\textbar 92.60 \\
    Qwen3.5-397B-A17B &
      68.44 & \oursofficial{73.90}{77.6} &
      76.70 & 58.68 & 69.80 & 58.68 &
      76.16 & 68.62 & 38.74 & 44.77 & 42.87 &
      28.59 & 32.90 & 16.44\textbar90.34 \\
    Doubao2.0-Lite (non-thinking) &
      71.20 & 73.30 & 76.70 & 48.83 & 69.40 & 49.97 & 71.42 &
      68.96 & 58.72 & 52.66 & 67.74 & 55.77 & 31.84 &
      15.83\textbar75.85 \\
    Doubao2.0-Lite (thinking, high) &
      \textbf{75.17} & 78.25 & \textbf{84.65} & 53.03 &
      \textbf{70.70} & 58.62 & 72.69 & 71.98 &
      \textbf{62.35} & \textbf{58.50} & \textbf{69.98} &
      \textbf{58.56} & 35.42 &
      \textbf{20.08}\textbar91.36 \\
    \midrule
    Kling-MLLM-35B-A3B &
      73.37 & 75.58 & 76.00 & \textbf{75.07} & 68.60 &
      57.94 & 77.31 & 70.98 &
      60.64 & 57.72 & 67.15 & 57.04 & 42.02 & 19.87\textbar85.59 \\
    Kling-MLLM-27B &
      73.22 & 75.35 & 78.80 & 73.20 & 70.60 & 62.86 & 76.35 &
      71.83 & 58.25 & 56.45 & 63.53 & 54.75 & 47.23 &
      19.35\textbar88.20 \\
    Kling-MLLM-397B-A17B &
      72.48 & 77.88 & 80.40 & 74.77 & 70.40 &
      \textbf{65.52} & \textbf{78.58} & \textbf{73.94} &
      60.55 & 58.31 & 65.53 & 57.82 &
      \textbf{54.72} & 19.51\textbar\textbf{93.81} \\
    \bottomrule
  \end{tabular}}
\end{table}

\paragraph{Video.}
Table~\ref{tab:video} covers both general and fine-grained video
understanding.  It includes LongVideoBench
\cite{wu2024longvideobenchbenchmarklongcontextinterleaved}, MVBench
\cite{li2024mvbench}, MME-VideoOCR
\cite{shi2026mmevideoocrevaluatingocrbasedcapabilities}, VSI-Bench
\cite{yang2025thinkingspacemultimodallarge}, ShortVid-Bench
\cite{ge2025archunyuanvideo7bstructuredvideocomprehension}, DSI-Bench
\cite{zhang2025dsibenchbenchmarkdynamicspatial}, Event-Bench
\cite{du2024eventorientedlongvideounderstanding}, MotionBench
\cite{hong2026motionbenchbenchmarkingimprovingfinegrained}, TimeLens
\cite{zhang2026timelensrethinkingvideotemporal}, Camera Motion VQA
\cite{feng2026geometryguidedcameramotionunderstanding}, and VidCapBench
\cite{chen2025vidcapbench}.  The TimeLens results additionally report subsets
derived from Charades \cite{sigurdsson2016hollywoodhomescrowdsourcingdata},
QVHighlights \cite{lei2021qvhighlightsdetectingmomentshighlights}, and
ActivityNet Captions \cite{krishna2017densecaptioningeventsvideos}.
Kling-MLLM-397B-A17B leads five columns: DSI-Bench, Event-Bench,
MotionBench, Camera Motion VQA, and VidCapBench coverage, while Kling-MLLM-35B-A3B leads VSI-Bench.
The high-thinking Doubao2.0-Lite setting leads LongVideoBench, MME-VideoOCR,
ShortVid-Bench, the overall TimeLens score and all three of its subset scores,
and VidCapBench accuracy.

Overall, the combined results show the clearest gains for
Kling-MLLM-397B-A17B on video perception and temporal understanding, audio
temporal grounding, and audio--video captioning.  Image, speech recognition,
video captioning, and multimodal interleaving results remain mixed across
systems.

\section{Caption: From Pixels to Trainable Semantics}
\label{section:caption}

When training a video generator, a caption is more than a human-readable summary: it is the semantic interface that converts visual evidence into conditional supervision. Omitted details leave controllable factors underspecified, forcing the generator to fall back on dataset priors, whereas inaccurate statements inject conflicting training signals. Building a production-grade captioning system therefore requires rethinking both what information captions preserve and how they organize visual and acoustic evidence over time.

\begin{figure}[t]
  \centering
\includegraphics[width=\textwidth]{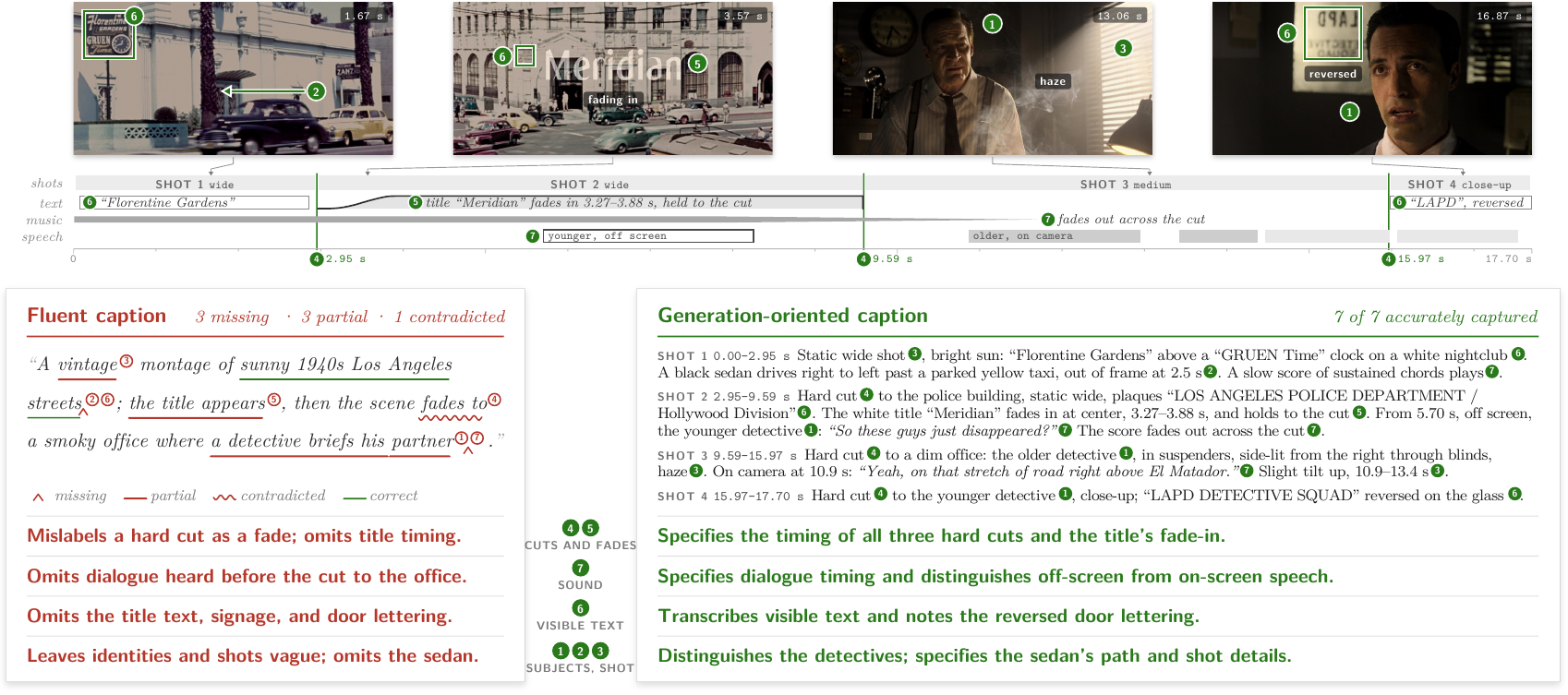}
\caption{Two captions of a 17.7\,s clip from \emph{Meridian}. Circled numbers link seven facts across the frames, timeline, and captions. The fluent caption reads well but omits or misrepresents key details. The generation-oriented caption preserves shot timing, audiovisual alignment, and visible text, including mirror-reversed lettering.}
\label{fig:ch3_teaser}
\end{figure}

A fluent sentence may correctly identify a scene while reversing spatial directions, collapsing sequential actions into a single state, hallucinating sounds, or losing subject identity across shots. Figure~\ref{fig:ch3_teaser} shows how linguistic fluency conceals supervision errors: the caption recognizes the setting, yet mislabels a hard cut as a fade, omits the title's timing, misses pre-appearance dialogue, and leaves visible text untranscribed. Such captions may score well on lexical-overlap metrics despite errors that compromise their value as supervision for generation.

\begin{observationpanel}
\textbf{Traditional captioning leaderboards and production captioning emphasize different objectives.} Standard metrics reward linguistic similarity to acceptable references, while training generative models requires faithful, complete, temporally organized, and consistently formatted representations of the source video.
\end{observationpanel}

This gap is increasingly recognized in the research literature. Standard sentence-level metrics fail to capture hallucinations even in short image captions~\cite{rohrbach2018object}. As captions become denser, evaluation has shifted from direct sentence matching toward decomposed multidimensional question-answer verification~\cite{chen2025vidcapbench}. These efforts suggest that once captions become supervision for generation, evaluation should move from sentence-level resemblance toward verifiable semantic facts.

However, evaluation using verifiable questions risks conflating captioning capability with VQA capability. VQA tests whether a model can verify a specified detail on demand, bypassing the harder problem of deciding what to state on its own. As CAPability shows, models exhibit a ``know but cannot tell'' gap: strong visual recognition does not guarantee comprehensive autonomous description~\cite{liu2025capability}. Consequently, strong VQA performance alone is insufficient to establish that a model can produce effective captions for training.

This section formalizes captioning as a tightly coupled engineering system comprising four core components:
(1) a semantic contract specifying what to describe and how to express it;
(2) a multi-expert production pipeline that distills complex signals into a deployable student model;
(3) conjunctive acceptance criteria for assessing caption usability; and
(4) downstream validation protocols that determine whether improved captions translate into better generation.

\subsection{The Semantic Contract: Content and Conventions}
\label{subsection:caption_contract}

The content specification of a caption should be dictated by the controllable factors and failure modes of the downstream generator. Table~\ref{tab:caption_contract} lists six types of information relevant to generation, including details that existing captioning datasets do not consistently capture.

\begin{table}[h]
\centering
\footnotesize
\caption{Generation-oriented caption specification.}
\renewcommand{\arraystretch}{1.20}
\setlength{\tabcolsep}{5pt}
\begin{tabularx}{\linewidth}{@{}L{0.19\linewidth} L{0.43\linewidth} Y@{}}
\toprule
\rowcolor{klingbg}
\textbf{Information family} & \textbf{Required caption elements} & \textbf{Downstream failure if omitted or corrupted} \\
\midrule
\textbf{Subject \& action} & Distinct entities, action phases, interaction objects, resulting states & Static posing, role swaps, unresolved actions \\
\rowcolor{black!2}
\textbf{Space \& composition} & Left--right/front--back relations, entry/exit directions, framing, depth & Reversed motion, misplaced subjects, uncontrolled composition \\
\textbf{Cinematography \& light} & Shot scale, viewpoint, camera motion, focal shift, light direction/quality, contrast, color temp. & Generic camera behavior, appealing but irreproducible style \\
\rowcolor{black!2}
\textbf{Time \& shots} & Event order, state changes, shot boundaries, timestamps, recurring subjects & Frame-list descriptions, discontinuity, unintended cuts or freezes \\
\textbf{Visible \& audible text} & Faithful OCR, dialogue/narration, env.\ \& action sounds, music, time anchors & Misspelled text, hallucinated audio, misaligned A-V events \\
\rowcolor{black!2}
\textbf{Aesthetics} & Concrete texture, palette, medium, style, and finish (beyond composition/light) & Vague non-executable descriptors (e.g., ``cinematic'') \\
\bottomrule
\end{tabularx}
\label{tab:caption_contract}
\end{table}

However, rigidly enforcing this specification can be counterproductive. Subject identity, action, and temporal order are generally mandatory, while OCR, speech, sound effects, and multi-shot structure should be included only when supported by the input. Furthermore, caption granularity also depends on the downstream use case. Short captions suit retrieval, whereas dense captions provide richer supervision for generation. Mixing the two without a clear policy creates an unmanageable training distribution.

Even when generating fluent descriptions, models frequently fail on critical dimensions, including spatial relations, camera movement, composition, lighting, and shot boundaries. While often attributed to limited capacity, these failures stem primarily from supervision gaps. Generic datasets prioritize vague aesthetic descriptors such as ``cinematic lighting'' and ``beautiful composition'' over explicit spatial and technical labels, limiting the benefits of scaling. Long videos further expose temporal failures, including reordered events, merged shots, and inconsistent references. Therefore, a video caption should not be viewed as an extended image description but rather as a compact temporal data structure.

While content dictates which facts enter the supervision signal, writing conventions determine whether those facts form a stable, bindable interface for the text encoder and generator. To reliably transmit these semantic details, four formatting practices yield disproportionate engineering leverage:

\begin{enumerate}[leftmargin=*]
    \item \textbf{Name entities explicitly and consistently.} Use discriminative noun phrases and retain them across sentences and shots. Avoid ambiguous expressions such as ``he'', ``it'', or ``the other person'' when the referent may be unclear to the generator. Stable naming improves both textual coreference and visual identity continuity.

    \item \textbf{Treat shot timestamps as alignment signals.} Anchor each shot boundary to its corresponding content using monotonic, validated timestamps. Accurate timestamp-content pairs explicitly teach the model when semantic conditions shift.

    \item \textbf{Transcribe visible text faithfully.} Preserve original characters and salient styles instead of paraphrasing. Omitting visible text teaches the model that lettering is an uncontrolled texture, while incorrect transcription introduces contradictory supervision.

    \item \textbf{Integrate audio with visual elements temporally.} Describe dialogue, salient sound effects, and music at the points where they occur within the visual narrative. Crucially, audio-related fields should be activated only when an audio track has been inspected.
\end{enumerate}

\subsection{The Supervision Pipeline: Separating Complex Discovery from Efficient Deployment}
\label{subsection:caption_pipeline}

Satisfying this semantic contract at production scale is beyond what a single prompt to a proprietary model can deliver. Constructing reliable supervision requires a multi-stage pipeline:

\begin{enumerate}[leftmargin=*]
    \item \textbf{Inspection and Routing:} Videos are classified by content type, duration, motion dynamics, shot density, and audio availability to determine the necessary expert pathways.

    \item \textbf{Perceptual Evidence Preservation:} Preprocessing steps such as downsampling, black-frame removal, audio decoding, and OCR/ASR ensure that critical signals remain intact before reaching expert models.

    \item \textbf{Routed Expert Claims:} A general visual expert identifies subjects and scene content, while specialized models extract specific dimensions, such as camera motion, spatial relations, lighting, and shot boundaries.

    \item \textbf{Integration and Conflict Resolution:} The integration layer resolves conflicting expert claims against source evidence, normalizes terminology, assigns stable entity names, and removes redundancy.

    \item \textbf{Conjunctive Quality Gating:} Candidates should strictly satisfy thresholds across accuracy, coverage, and diversity, alongside cross-modal and cross-shot consistency checks. Rejected cases flow into a difficulty-indexed error pool for diagnosis and targeted remediation.
\end{enumerate}

\begin{insightpanel}
\textbf{Robust supervision originates from orchestrated pipelines.} Explicitly mining previously unobserved dimensions and resolving inter-expert conflicts expand semantic coverage and accuracy far more effectively than simply scaling up a general-purpose annotator.
\end{insightpanel}

Although this multi-expert pipeline produces comprehensive and reliable supervision, deploying it directly incurs prohibitive latency because of its serial execution. To address this limitation, the pipeline's output should be distilled into a single-pass student model. This reduces the deployment footprint while preserving globally consistent entity naming and the semantic contract synthesized by the offline factory.
\subsection{Acceptance Criteria: Conjunctive Quality and Atomic Benchmarking}
\label{subsection:caption_criteria}

A production-grade caption should satisfy four criteria simultaneously: factual accuracy, semantic coverage, beneficial diversity, and internal consistency. These four criteria are strictly conjunctive: optimizing any single dimension in isolation induces predictable shortcuts and specific generation failures, as illustrated in Table~\ref{tab:caption_criteria}.

\begin{table}[h]
\centering
\footnotesize
\caption{Shortcuts from optimizing each caption criterion in isolation.}\vspace{-2pt}
\renewcommand{\arraystretch}{1.22}
\setlength{\tabcolsep}{5pt}
\begin{tabularx}{\linewidth}{@{}L{0.19\linewidth} L{0.36\linewidth} Y@{}}
\toprule
\rowcolor{klingbg}
\textbf{Criterion} & \textbf{Characteristic shortcut} & \textbf{Downstream consequence} \\
\midrule
\textbf{Factual accuracy} & Omit uncertain or complex attributes & Underspecified condition $\rightarrow$ dataset-prior fallback \\
\rowcolor{black!2}
\textbf{Semantic coverage} & Add plausible but unverified details & Hallucinated entities, actions, relations, or text \\
\textbf{Beneficial diversity} & Use uncontrolled synonyms or poetic terms & Fragmented TE distribution $\rightarrow$ unstable control \\
\rowcolor{black!2}
\textbf{Internal consistency} & Reuse rigid templates; ignore local nuances & Lost dynamic state changes $\rightarrow$ robotic generation \\
\bottomrule
\end{tabularx}
\label{tab:caption_criteria}
\end{table}

To prevent these shortcuts, evaluation should move beyond lexical metrics and test these criteria directly. As CHAIR demonstrates, higher sentence-level scores do not imply fewer hallucinations~\cite{rohrbach2018object}. This vulnerability is especially pronounced in dense captions, where local errors are easily diluted by surrounding fluent text.

Therefore, the appropriate evaluation unit is an atomic, source-grounded information item. Instead of matching surface text, evaluators should verify whether a candidate caption correctly expresses predefined essential facts and their temporal scopes. This atomic formulation is largely agnostic to surface phrasing, effectively disentangling the four acceptance criteria that reference-based metrics tend to conflate.

A comprehensive benchmark should expose at least four complementary views:
\begin{itemize}[leftmargin=*]
    \item \textbf{Content validity:} factual precision, semantic coverage, beneficial diversity, and internal consistency;
    \item \textbf{Generation-specific dimensions:} action progression, spatial direction, cinematography, composition, lighting, temporal ordering, shot transitions, OCR, speech, and sound;
    \item \textbf{Interface compliance:} parse rate, timestamp validity, terminology compliance, and marker validity;
    \item \textbf{Difficulty slices:} duration, shot count, entity count, event count, motion speed, OCR density, audio complexity, and content domain.
\end{itemize}

Finally, automatic judges require rigorous stability validation prior to deployment. This entails measuring judgment variance on identical captions, calibrating against human annotations on stratified subsets, and checking ranking robustness across different model families. Crucially, evaluation should control for caption length to prevent judges from conflating verbosity with true semantic coverage. More broadly, the judge model, evaluation prompt, decoding parameters, and preprocessing steps together define the benchmark configuration and should therefore be versioned as part of the evaluation protocol.

\begin{insightpanel}
\textbf{True superiority withstands granular decomposition.} A credible improvement specifies which facts are more accurate, which dimensions are more complete, and under which difficulty slices the improvements persist. Gains that vanish when controlling for caption length or appear only in aggregate scores provide insufficient evidence of better supervisory quality.
\end{insightpanel}

\subsection{Downstream Validation: The Generator as the Final Consumer}
\label{subsection:caption_downstream}

Intrinsic caption evaluations function as proxies until validated by a downstream consumer. Recent benchmarks such as VidCapBench~\cite{chen2025vidcapbench} and CapRiCorn-1K~\cite{chen2026capricorn} demonstrate strong correlations between their evaluation scores and downstream performance, including text-to-video generation quality and video reconstruction fidelity, thereby establishing a core experimental principle: caption quality should ultimately be judged by the generative capabilities it enables in the downstream model.

\begin{observationpanel}
\textbf{Descriptive quality does not automatically translate into supervisory quality.} A caption may become more readable while exceeding the text encoder's effective context, adding details that the generator cannot localize in time, or inflating evaluation scores via generic aesthetic terms. Downstream validation is essential to distinguish superficial linguistic improvement from usable conditional information.
\end{observationpanel}

Controlled downstream experiments help distinguish usable conditional information from superficial linguistic gains. Matching caption lengths and verified fact counts can reduce confounding effects, while targeted interventions, such as modifying camera-motion descriptions alone, can reveal whether improvements transfer to the corresponding generation capability. Combining human preferences with fine-grained metrics---spanning subject identity, action, space, camera, temporal structure, text rendering, audio alignment, and visual quality---exposes specific control failures that a global text--video similarity score obscures.

Ultimately, the value of a captioning strategy lies in its measurable effect on generation. Small-scale fine-tuning or evaluation with a frozen generator provides a practical way to assess whether improved captions yield more faithful instruction following before changes are scaled to larger training pipelines.

\subsection{Experiments}
\begin{table}[htbp]
  \centering
  \caption{Captioning benchmarks.}
  \label{tab:caption_bench}
  \scriptsize
  \setlength{\tabcolsep}{2.5pt}
  \renewcommand{\arraystretch}{1.18}
  \resizebox{\linewidth}{!}{%
  \begin{tabular}{@{}
    >{\raggedright\arraybackslash}p{4.0cm}
    *{14}{>{\centering\arraybackslash}c}@{}}
    \toprule
    \multirow{2}{*}{\textbf{Model}} & \multicolumn{4}{c}{\textbf{VidCapBench}} & \multicolumn{4}{c}{\textbf{UGC-VideoCap}} & \multicolumn{2}{c}{\textbf{Motion (Internal)}} & \multicolumn{2}{c}{\textbf{Camera (Internal)}} & \multicolumn{2}{c}{\textbf{Sound (Internal)}} \\
    \cmidrule(lr){2-5} \cmidrule(lr){6-9} \cmidrule(lr){10-11} \cmidrule(lr){12-13} \cmidrule(lr){14-15}
    & \textbf{Acc.} & \textbf{Pre.} & \textbf{Cov.} & \textbf{Con.} & \textbf{Audio} & \textbf{Visual} & \textbf{Detail} & \textbf{Avg.} & \textbf{Pre.} & \textbf{Rec.} & \textbf{Pre.} & \textbf{Rec.} & \textbf{Pre.} & \textbf{Rec.} \\
    \midrule
    Gemini-4-Argon & 16.20 & 51.66 & 85.63 & \textbf{10.69} & 79.80 & 87.69 & 85.67 & 84.38 & 64.62 & 45.78 & 40.83 & 76.22 & 79.74 & 29.38 \\
    Doubao2.1-Lite (non-thinking) & 17.20 & 51.78 & 87.47 & 8.32 & 82.88 & 90.67 & 88.81 & 87.45 & 62.87 & 56.31 & 28.37 & 61.46 & 73.18 & 26.79 \\
    Doubao2.1-Lite (thinking, high) & 20.23 & 56.44 & 94.15 & 3.36 & 90.37 & 96.10 & 94.97 & 93.81 & 64.15 & 59.02 & 32.89 & 64.61 & 77.56 & 31.69 \\
    \midrule
    Kling-Captioner-35B-A3B & \textbf{21.84} & 59.28 & 94.81 & 3.56 & 92.83 & 96.46 & 95.90 & 95.06 & 67.28 & 64.75 & 52.18 & 80.93 & 78.52 & 55.83 \\
    Kling-Captioner-397B-A17B & 21.70 & \textbf{60.51} & \textbf{96.05} & 3.19 & \textbf{93.41} & \textbf{96.70} & \textbf{96.38} & \textbf{95.50} & \textbf{68.11} & \textbf{67.42} & \textbf{54.82} & \textbf{83.43} & \textbf{80.21} & \textbf{58.11} \\
    \bottomrule
  \end{tabular}}
\end{table}

Building on Kling-MLLM-35B-A3B and Kling-MLLM-397B-A17B, we construct targeted post-training data guided by the insights in this section to obtain Kling-Captioner. Table~\ref{tab:caption_bench} compares Kling-Captioner with frontier models on VidCapBench~\cite{chen2025vidcapbench}, UGC-VideoCap~\cite{wu2025ugcvideocaptioneromniugcvideo}, and our internal benchmark. The results show that Kling-Captioner consistently outperforms the baselines across all three benchmarks, achieving the best overall accuracy, precision, and coverage on VidCapBench while leading in every dimension on UGC-VideoCap and across our internal evaluation targets for motion, camera, and sound.

\subsection{Discussion}

Scaling the captioning system described in this section leaves several open challenges for future research. First, accurately capturing highly dynamic details necessitates balancing temporal and spatial fidelity against limited context capacity. Second, long-form videos demand sophisticated strategies to navigate the trade-off between overall caption length and descriptive density while maintaining robust referential consistency for subjects over extended time horizons. Finally, stronger audiovisual integration is needed to accurately align acoustic and visual events and explicitly associate speakers with their utterances.

The framework developed in this section focuses on absolute video descriptions, detailing the objective content of a target video. Reference-conditioned generation, however, introduces new operational constraints: the caption should additionally specify reference bindings, isolate preserved attributes, and describe relative transformations while maintaining compatibility with the existing textual conditioning interface. The engineering principles governing this relative semantic contract are discussed in the next section.

\section[Reference-Grounded Captioning: Orchestrating Coherent Multimodal Anchors]{{\fontsize{14.1pt}{18pt}\selectfont\hspace{-6pt} Reference-Grounded Captioning: Orchestrating Coherent Multimodal Anchors}}
\label{section:ref_caption}

\begin{figure}[t]
  \centering
\includegraphics[width=\textwidth]{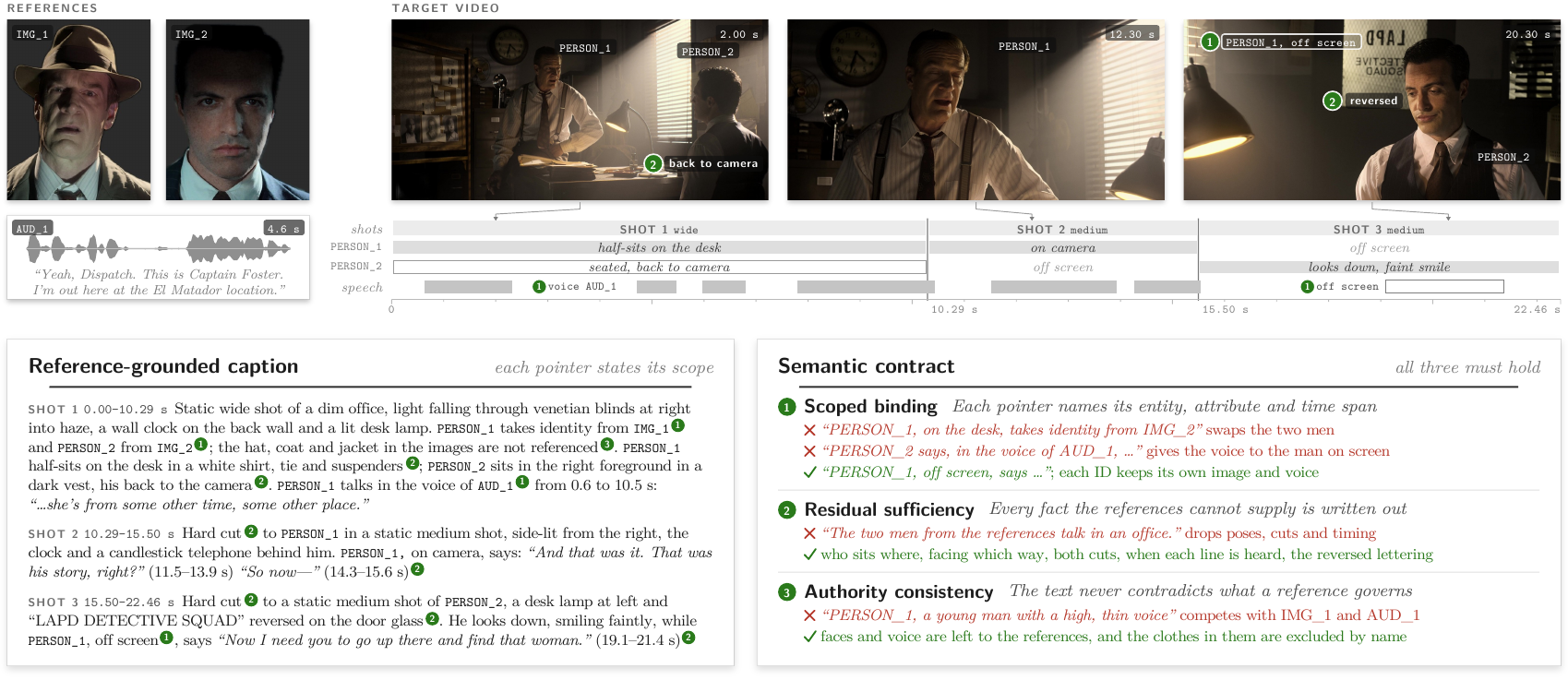}
\caption{Reference-grounded captioning on a 22.5\,s clip from \emph{Meridian}, given two portraits (IMG\_1, IMG\_2) and a voice track (AUD\_1). The caption states each binding with its scope and writes out everything the references cannot supply. The timeline shows who is on screen and when PERSON\_1 speaks. Circled numbers mark where each property of the semantic contract applies, contrasting typical shortcuts with correct formulations.}
\label{fig:ch4_teaser}
\end{figure}

Reference-conditioned generation enables controllable video creation through multimodal references, such as character portraits, source clips, and voice tracks. However, introducing reference conditions creates a demanding correspondence problem. Each reference requires explicit binding to the correct target entity, property, modality, and temporal interval. Without precise alignment, a generated video may appear perceptually flawless while suffering from critical systemic failures, such as exchanged identities, leaked attributes, edited protected content, or broken state continuity.

Reference-grounded captioning (RGC) describes target videos conditioned on one or more multimodal references. Extending the semantic interface from Section~\ref{section:caption}, RGC explicitly specifies reference-target bindings, modified properties, and supplemental target facts. At training scale, simply pairing references with conventional captions leaves these relationships implicit, causing the generator to learn global appearance correlations without reliable entity-level control. RGC resolves this problem by making the intended correspondences explicit, thereby providing robust supervision for both reference fidelity and controlled modification. Figure~\ref{fig:ch4_teaser} illustrates RGC on a multi-shot clip conditioned on two character portraits and a voice track.

\subsection{Task Definition: Reference Semantics and Systemic Challenges}
\label{subsection:rgc_definition}

Rather than generating an unconstrained summary from a target video, reference assets, and task instructions, RGC defines the resulting caption as an integration of three distinct components: explicit pointers linking reference assets to target entities, supplementary descriptions capturing unreferenced facts, and scoped preservation or change records preventing text-reference conflicts.

Crucially, RGC requires determining the scope of each reference's authority, which is asymmetric across properties. For instance, a portrait may dictate facial identity while leaving body pose free; a source video may lock background geometry while licensing a requested localized edit; and an audio clip may define voice or rhythm without dictating visual action. Table~\ref{tab:rgc_tasks} outlines how superficially similar inputs carry divergent semantic authority based on the product task.

\begin{table}[h]
\centering
\footnotesize
\caption{Product tasks covered by RGC.}
\renewcommand{\arraystretch}{1.20}
\setlength{\tabcolsep}{5pt}
\begin{tabularx}{\linewidth}{@{}L{0.16\linewidth} L{0.23\linewidth} L{0.28\linewidth} Y@{}}
\toprule
\rowcolor{klingbg}
\textbf{Product task} & \textbf{Typical references} & \textbf{Binding question} & \textbf{Residual target facts} \\
\midrule
\textbf{Video editing} & Source video, instruction, optional local ref. & Edited entity/region and protected properties & Requested state change, time span, secondary effects \\
\rowcolor{black!2}
\textbf{Multi-reference generation} & Character, object, scene, style, voice, or motion refs. & Reference governing each role, attribute, and modality & Target interaction, action, camera, timing, composition \\
\textbf{Video continuation} & Prefix video + optional entity or style refs. & Subjects, states, and scene geometry persisting across cut & Post-boundary events and causal link to prefix \\
\rowcolor{black!2}
\textbf{Performance transfer} & Driving video, pose, trajectory, or audio & Identity source and performance source & Target retargeting, camera behavior, synchronization \\
\bottomrule
\end{tabularx}
\label{tab:rgc_tasks}
\end{table}

RGC inherits all requirements of standard captioning while introducing four additional challenges:
\begin{itemize}[leftmargin=*]
  \item \textbf{Reference-target confusion:} Populating a scene with multiple entities introduces severe combinatorial assignment ambiguity, especially when the references are semantically similar.
  \item \textbf{Asymmetric modality authority:} Different modalities naturally govern distinct properties (e.g., pixels for appearance, audio for timbre, video for motion), yet explicit instructions can selectively override these priors, requiring dynamic conflict resolution at the granular level of entities, attributes, and temporal intervals.
  \item \textbf{Temporal binding persistence:} Maintaining consistent identity associations across time is inherently fragile. Models should track and bind entities through severe visual and temporal disruptions, such as occlusions, abrupt cuts, and drastic viewpoint changes.
  \item \textbf{Degenerate shortcuts:} Models can artificially inflate precision by omitting entity IDs, using vague phrasing, or skipping complex edits entirely. Although factually safe, these shortcuts starve the generator of essential supervision, necessitating strict evaluation of required coverage alongside correctness.
\end{itemize}

\begin{observationpanel}
\textbf{Difficulty is governed by reference separability and complex attribute binding, not by reference count alone.} Similar subjects increase assignment ambiguity, while capturing localized state changes requires isolating relevant properties from a vast set of contextual invariants.
\end{observationpanel}

\subsection{The Semantic Contract: Scoped Binding, Residual Sufficiency, and Authority Consistency}
\label{subsection:rgc_contract}

Reliable RGC supervision relies on three conjunctive properties: scoped binding, residual sufficiency, and authority consistency. Table~\ref{tab:rgc_contract} summarizes their roles and illustrates how violating any single axis compromises the reliability of the resulting supervision signal.

\begin{table}[h]
\centering
\footnotesize
\caption{The semantic contract of RGC.}
\renewcommand{\arraystretch}{1.22}
\setlength{\tabcolsep}{5pt}
\begin{tabularx}{\linewidth}{@{}L{0.19\linewidth} L{0.36\linewidth} Y@{}}
\toprule
\rowcolor{klingbg}
\textbf{Requirement} & \textbf{What it guarantees} & \textbf{Characteristic shortcut and downstream failure} \\
\midrule
\textbf{Scoped binding} & Pointer binds to target entity, attribute, event, \& interval & Property bound to wrong target $\rightarrow$ spurious correlations \\
\rowcolor{black!2}
\textbf{Residual sufficiency} & All unreferenced target facts stated explicitly & Vague ``same as ref'' omits action/camera/time/audio $\rightarrow$ underspecified \\
\textbf{Authority consistency} & Reference-governed traits kept unless scoped override & Verbose text contradicts ref.\ $\rightarrow$ competing conditions \\
\bottomrule
\end{tabularx}
\label{tab:rgc_contract}
\end{table}

These requirements extend the four caption criteria introduced in Section~\ref{subsection:caption_criteria}. Factual accuracy now mandates rigorous binding verification alongside factual correctness. Semantic coverage expands to encompass standard target facts, required references, and residual target specifications. Beneficial diversity primarily applies to the target description, while pointer syntax and entity naming should remain deliberately stable. Internal consistency requires strict alignment among the caption, references, and the target.

\textbf{Scoped Binding.} A valid binding specifies the source evidence, target entity, authorized attribute, and temporal scope. The instruction ``Use image 2'' is incomplete when the image contains a person, a specific garment, a room, and a photographic style. A sufficient binding should instead specify, for example, that ``\texttt{PERSON\_1} takes its identity and outfit from \texttt{IMG\_2}; the background is not referenced.'' Persistent identifiers are thus essential, as bare noun phrases cannot reliably anchor identities when multiple entities recur.

\textbf{Residual Sufficiency.} While an RGC caption need not reproduce the entire target description, it should specify every generation-relevant fact that cannot be recovered from authorized references. For character portraits, the caption usually includes action, interaction, camera behavior, timing, scene placement, and sound. For continuation, it details the post-prefix event and the state transition at the temporal boundary. For editing, it includes the edited attribute, region, and temporal interval, along with any intended cross-modal consequence.

\textbf{Authority Consistency.} Once an attribute is delegated to a reference, the caption should not independently assign it an incompatible value. Re-describing intricate facial geometry, voice timbre, or complex motion in text is inherently lossy and may also introduce competing conditions. To enforce authority consistency, the caption should defer to the authorized reference for governed properties, except where an override is explicitly declared together with its scope. By otherwise restricting itself to the necessary residual facts, the caption avoids silent cross-modal overrides.

\begin{insightpanel}
\vspace{-2pt}
\textbf{The semantic contract transforms the caption into an active control schematic.} Scoped binding delegates authority to the reference evidence, residual sufficiency fills the contextual gaps, and authority consistency prevents cross-modal interference, delivering a coherent and unambiguous instruction set to the generator.
\end{insightpanel}

\vspace{-6pt}
\subsection{Relational Supervision: Learning from Arbitrary Video Pairs}\vspace{-2pt}
\label{subsection:rgc_relational_supervision}

RGC lets us turn any two videos with sufficient visual or acoustic similarity
into training data for the DiT. We take one as the reference and annotate
the other as the target, describing what they share and what the generator
needs to change. The videos need not come from an editing process, and no
original edit instruction is required. The reference and the resulting
caption serve as conditioning inputs, with the target video providing the supervision.

The similarity may cover only part of the content. Two videos can show a
similar action performed by different people in different surroundings.
The caption uses the reference for the shared movement and writes out the
target's appearance, setting, and differences in the performance. Two
recordings of the same speaker can likewise share a voice while differing in
words and delivery. The target dialogue and its timing remain in the caption.
In either case, the annotation specifies which part of the target the
reference supplies, in accordance with the scope established above.

This expands the supply of training data beyond purpose-built editing pairs.
A target clip can be paired with several references: it may share a subject
with one clip and a motion pattern with another. Each pairing asks the DiT
to use its reference differently. The target remains fixed, but the caption and
the content drawn from the reference both change. One video can therefore
contribute several reference-conditioned examples alongside its standard
caption. The number and variety of useful training pairs can grow with the
video collection, without producing a separate edited result for every pair.

Training on this wider range of pairs strengthens the Omni
DiT's ability to follow visual and acoustic conditions. Each example requires
it to use the relevant content from the reference and produce the differences
described by the caption, including changes that unfold over time. This
improvement is reflected in reference fidelity and in edits that change the
intended content while preserving the rest. The captions retain the target
coverage and conventions of standard captioning, allowing standard and
reference-conditioned data to be used together.

\begin{insightpanel}
\vspace{-2pt}
\textbf{Any sufficiently similar video pair is a potential training example
for the DiT.} Relational annotation makes the pair usable by stating what
to retain and what to change.
\end{insightpanel}

\vspace{-6pt}
\subsection{Unified Representation: Integrating Captions with Reference Pointers}\vspace{-2pt}
\label{subsection:rgc_format}

Reference-grounded tasks inherit the structure, terminology, and descriptive granularity of standard captions and augment them with stable reference pointers and explicit change scopes. This unified representation allows standard and reference-grounded data to share the same captioner, text encoder, PE policy, and validators, with standard captioning treated as a zero-reference case. The generator therefore preserves a consistent target organization regardless of whether details are expressed literally or supplied through reference pointers. As a result, large-scale standard-caption data can provide broad semantic supervision, while scarcer multi-reference data focuses on learning reference binding and controlled modification.

\begin{insightpanel}
\textbf{Representation consistency enables compositional supervision.} A unified representation allows the model to combine broad semantic coverage learned from large-scale standard captions with the reference-binding logic derived from scarcer multi-reference data.
\end{insightpanel}

Different reference modalities contribute distinct information and introduce unique ambiguities. Table~\ref{tab:rgc_modalities} summarizes the typical authority, main ambiguity, and corresponding caption-side response for each modality.

\begin{table}[h]
\centering
\footnotesize
\caption{Authority and ambiguity of each reference modality.}
\renewcommand{\arraystretch}{1.20}
\setlength{\tabcolsep}{5pt}
\begin{tabularx}{\linewidth}{@{}L{0.09\linewidth} L{0.22\linewidth} L{0.27\linewidth} Y@{}}
\toprule
\rowcolor{klingbg}
\textbf{Reference} & \textbf{Typical authority} & \textbf{Main ambiguity} & \textbf{Caption-side response} \\
\midrule
\textbf{Image} & Identity, appearance, object detail, scene/style & Which entity and image subset should transfer & Bind explicit ID \& scope; state target action, placement, camera, changes \\
\rowcolor{black!2}
\textbf{Video} & Motion, state, performance, continuity, camera & Time alignment; transfer vs.\ keep; cut boundary & Declare source interval \& role; timestamp new events \& preserved content \\
\textbf{Audio} & Voice, dialogue, rhythm, sound effects & Speaker/source assignment \& A-V sync & Bind speaker/event IDs to visible actions; no unseen visual hallucination \\
\rowcolor{black!2}
\textbf{Text} & Edit instruction, script, names, constraints & Conflicts with sources or over-specifies details & Treat as typed change spec.; resolve authority at attribute level \\
\textbf{Mixed set} & Joint composition of the above & Overlapping authority, duplication, contradiction & Build attribute-level authority map before composing final caption \\
\bottomrule
\end{tabularx}
\label{tab:rgc_modalities}
\end{table}

Specifically, image references risk identity swaps and attribute leakage, demanding bindings that distinguish identity, pose, and style while defining the precise transfer scope for each attribute. Video references introduce role and temporal ambiguity. Since a source video may dictate content, motion, editing, or camera behavior, captions require explicit role bindings and temporal scopes that withstand shifts in speed, viewpoint, or duration. Audio references trigger source-assignment ambiguity, necessitating independent speaker and event scopes. Text instructions encode intended transformations rather than the actual state of the target, meaning captions must verify execution outcomes to differentiate intent from observed reality.

Together, these modality-specific requirements converge on a hybrid representation featuring (1) a compact reference registry with stable IDs, modalities, roles, and permitted scopes; (2) a foundational target-video description utilizing registered IDs; (3) explicit preservation and modification records for editing and continuation; and (4) shot or event timestamps that connect bindings to applicable intervals.

\subsection{The Supervision Pipeline: Extending for Cross-Modal Correspondences}
\label{subsection:rgc_pipeline}

Generating RGC data at scale requires extending the multi-expert design in
Section~\ref{subsection:caption_pipeline} to include dedicated stages for cross-modal reference parsing, rigorous correspondence checking, and systematic conflict resolution.

\begin{enumerate}[leftmargin=*]
    \item \textbf{Type the task and references.} Identify whether each input supplies identity, appearance, motion, camera, voice, style, or instruction. Record its modality, quality, temporal extent, and permissible transfer scope.

    \item \textbf{Describe target and references independently.} Preserve a standard target description to maintain target-side coverage. In parallel, extract entities, attributes, actions, shots, audio events, and specialist evidence from each reference without prematurely assuming correspondences.

    \item \textbf{Propose and verify bindings.} Use detection, segmentation, tracking, OCR or ASR, motion features, and MLLM reasoning to generate and jointly verify candidate matches across viewpoints and over time.

    \item \textbf{Determine preservation and residual sufficiency.} Compare target and source under the declared task. Explicitly distinguish requested changes, preserved properties, incidental changes, and unresolved evidence.

    \item \textbf{Resolve authority and conflicts.} Construct an entity-attribute-time map in which each cell identifies the authoritative input. Adjudicate mutually inconsistent expert claims before prose generation.

    \item \textbf{Write once in the common format.} Insert scoped pointers, retain standard target facts not supplied by references, and remove only verified redundancy. A single integrated target description is preferable to serial rewrites that repeatedly rename or rebind entities.

    \item \textbf{Apply deterministic and semantic gates.} Check pointer validity, binding coverage, forbidden cross-scope transfer, text-reference conflicts, temporal alignment, and the four standard captioning criteria.
\end{enumerate}

\subsection{Experiments}
\begin{table}[htbp]
  \centering
  \caption{RGC benchmarks.}
  \label{tab:rgc_bench}
  \scriptsize
  \setlength{\tabcolsep}{2.5pt}
  \renewcommand{\arraystretch}{1.18}
  \resizebox{0.9\linewidth}{!}{%
  \begin{tabular}{@{}
    >{\raggedright\arraybackslash}p{4.0cm}
    *{10}{>{\centering\arraybackslash}c}@{}}
    \toprule
    \multirow{2}{*}{\textbf{Model}} & \multicolumn{3}{c}{\textbf{ViDiC}} & \multicolumn{6}{c}{\textbf{MRVBench}} \\
    \cmidrule(lr){2-4} \cmidrule(lr){5-10}
    & \textbf{Avg.} & \textbf{Diff.} & \textbf{Sim.} & \textbf{KP-Cov} & \textbf{VQA} & \textbf{Ref-Tag-R} & \textbf{Ref-Bind} & \textbf{Subj-F1} & \textbf{MRVScore} \\
    \midrule
    Gemini-4-Argon & 77.18 & 75.51 & 81.86 & 93.31 & 74.41 & \textbf{97.23} & 93.56 & \textbf{90.37} & \textbf{91.22} \\
    Doubao2.1-Lite (non-thinking) & 69.24 & 65.27 & 80.33 & 89.45 & 68.09 & 75.43 & 93.35 & 49.45 & 78.10 \\
    Doubao2.1-Lite (thinking, high) & 75.63 & 74.96 & 77.49 & 90.75 & 71.55 & 81.41 & 93.98 & 56.86 & 81.32 \\
    \midrule
    Kling-RGC-35B-A3B & 77.57 & 75.59 & 83.06 & 94.12 & 76.01 & 94.80 & 92.88 & 84.51 & 88.57 \\
    Kling-RGC-397B-A17B & \textbf{78.89} & \textbf{76.29} & \textbf{86.12} & \textbf{95.18} & \textbf{76.34} & 95.28 & \textbf{94.85} & 85.24 & 90.31 \\
    \bottomrule
  \end{tabular}}
\end{table}

Following the relational supervision pipeline and unified pointer representation introduced in this section, we further post-train Kling-Captioner on curated cross-modal correspondence data to obtain Kling-RGC. Table~\ref{tab:rgc_bench} compares Kling-RGC with frontier models on ViDiC~\cite{wu2025vidic} and MRVBench~\cite{liu2026refcaptioner}. Kling-RGC leads across all ViDiC metrics in both difference and similarity captioning; on MRVBench, it achieves the highest keypoint coverage, VQA accuracy, and reference-binding precision while remaining highly competitive with Gemini-4-Argon and substantially outperforming Doubao2.1-Lite across the remaining metrics.

\subsection{Discussion}

Future work could increase the flexibility of reference content and composition. Key directions include supporting diverse input forms (e.g., selected regions, sketches, and trajectories), enabling complex multimodal combinations to jointly specify a single target, and developing finer-grained temporal binding for precise synchronization and transitions. Advancing these areas requires targeted supervision and evaluation.

In real-world inference, user requests typically offer references and brief instructions without specifying the complete binding structure or scope of each reference. The remaining challenge is therefore to translate these underspecified requests into the same scoped, residual-sufficient, and authority-consistent caption distribution used during training. We address this challenge through PE in the following section.

\section{Prompt Enhancement: From User Intent to the Caption Distribution}
\label{section:pe}

Captioning determines the language that the generator learns
(Sections~\ref{section:caption} and~\ref{section:ref_caption}); PE
determines whether inference-time requests arrive in that
language.  Production requests are often short and colloquial or, increasingly,
long passages written by other AI models, whereas the generator is trained on dense captions that
describe subjects, action phases, camera movements, lighting, shots, and sound with a consistent
vocabulary and sentence style.  PE rewrites a request into a condition drawn
from this caption distribution while preserving the user's intent~\cite{zhu2026wanpe}.
HunyuanVideo established learned prompt rewriting for video
generation~\cite{kong2024hunyuanvideo}, and RAPO showed that
retrieval can move prompts toward the training distribution~\cite{gao2025rapo}.

\begin{figure}[t]
\centering
\begin{tikzpicture}[
  font=\sffamily\scriptsize,
  >={Stealth[length=1.8mm]},
  stage/.style={draw=black!45, rounded corners=3pt, fill=white, align=left,
    text width=4.3cm, inner sep=4pt, minimum height=2.2cm},
  infer/.style={draw=black!45, rounded corners=3pt, fill=white, align=center,
    text width=2.3cm, inner sep=3pt, minimum height=1.25cm},
  key/.style={draw=klinggreen, thick, fill=klingbg},
  lane/.style={draw=black!14, fill=black!3, rounded corners=5pt, inner sep=6pt},
  lanelabel/.style={font=\sffamily\bfseries\footnotesize, text=black!75,
    anchor=south west, inner sep=1pt},
  flow/.style={->, thick, black!60},
]
  \node[stage] (s1) at (2.3,0) {%
    \textbf{Stage 1: Caption-aligned SFT}\\[2pt]
    $\bullet$~Captions of compelling videos\\
    $\bullet$~Backward-built user prompts\\
    \hphantom{$\bullet$~}in real-traffic formats\\
    $\bullet$~F/R/V-verified pairs\\
    $\bullet$~RAG pairs via reject sampling};
  \node[stage] (s2) at (7.8,0) {%
    \textbf{Stage 2: Text-level RL}\\[2pt]
    $\bullet$~Faithfulness: checklist\\
    $\bullet$~Reasonableness: spatial and\\
    \hphantom{$\bullet$~}interaction graphs\\
    $\bullet$~Vividness: technique-library\\
    \hphantom{$\bullet$~}hit count};
  \node[stage] (s3) at (13.3,0) {%
    \textbf{Stage 3: DiT-in-the-loop RL}\\[2pt]
    $\bullet$~Visual quality\\
    $\bullet$~Motion quality (artifacts)\\
    $\bullet$~Text alignment (checklist)\\[1pt]
    {\color{black!60}PE \(\rightarrow\) text encoder \(\rightarrow\) DiT
    rollouts}};
  \draw[flow] (s1) -- (s2);
  \draw[flow] (s2) -- (s3);
  \node[lanelabel] (tl) at ($(s1.north west)+(0,0.12)$) {Training stages
    (basic PE)};

  \node[infer] (up) at (1.25,-3.5) {\textbf{User prompt}\\[1pt]
    {\color{black!60}often AI-written}};
  \node[infer] (sr) at (4.525,-3.5) {\textbf{Story refinement}\\[1pt]
    {\color{black!60}resolve plot contradictions}};
  \node[infer] (lib) at (4.525,-5.3) {\textbf{Retrieval}\\[1pt]
    {\color{black!60}Top-5 captions of compelling videos}};
  \node[infer, key] (pe) at (7.8,-3.5) {\textbf{Basic PE}\\[1pt]
    {\color{black!60}select from Top-5; write in the caption dialect}};
  \node[infer] (te) at (11.075,-3.5) {\textbf{Text encoder}\\
    \textbf{+ DiT}};
  \node[infer] (vid) at (14.35,-3.5) {\textbf{Video}};
  \draw[flow] (up) -- (sr);
  \draw[flow] (sr) -- (lib);
  \draw[flow] (sr) -- (pe);
  \draw[flow] (lib.east) -| node[pos=0.25, below, text=black!70] {Top-5}
    (pe.south);
  \draw[flow] (pe) -- (te);
  \draw[flow] (te) -- (vid);
  \node[lanelabel] (il) at ($(up.north west)+(0,0.12)$) {Inference stages};
  \node[font=\sffamily\bfseries\scriptsize, text=klinggreen, anchor=south,
    inner sep=1pt] (al) at ($(sr.north)+(0,0.12)$) {Agentic PE};

  \draw[flow, dashed, klinggreen] (s3.south) -- ++(0,-0.65)
    -| node[pos=0.25, above, text=klinggreen] {deployed basic PE} (pe.north);

  \begin{scope}[on background layer]
    \node[lane, fit=(tl)(s1)(s2)(s3)] {};
    \node[lane, fit=(il)(up)(vid)(lib)] {};
    \node[draw=klinggreen, dashed, rounded corners=4pt, fill=klingbg!60,
      inner sep=4pt, fit=(al)(sr)(lib)] {};
  \end{scope}
\end{tikzpicture}
\caption{Overview of PE.  \emph{Top:} three training stages of basic PE.
Stage~1 learns the mapping from real-format user prompts to captions of
compelling videos, using only pairs that pass faithfulness, reasonableness, and
vividness (F/R/V) verification; Stage~2 optimizes the same three properties
with text-only rewards; Stage~3 optimizes basic PE through the frozen text
encoder and DiT with video-level rewards.  \emph{Bottom:} at inference, agentic
PE makes the plot coherent and retrieves five captions of compelling videos;
basic PE writes the final condition, using the retrieved captions as optional
evidence.}
\label{fig:pe_overview}
\end{figure}
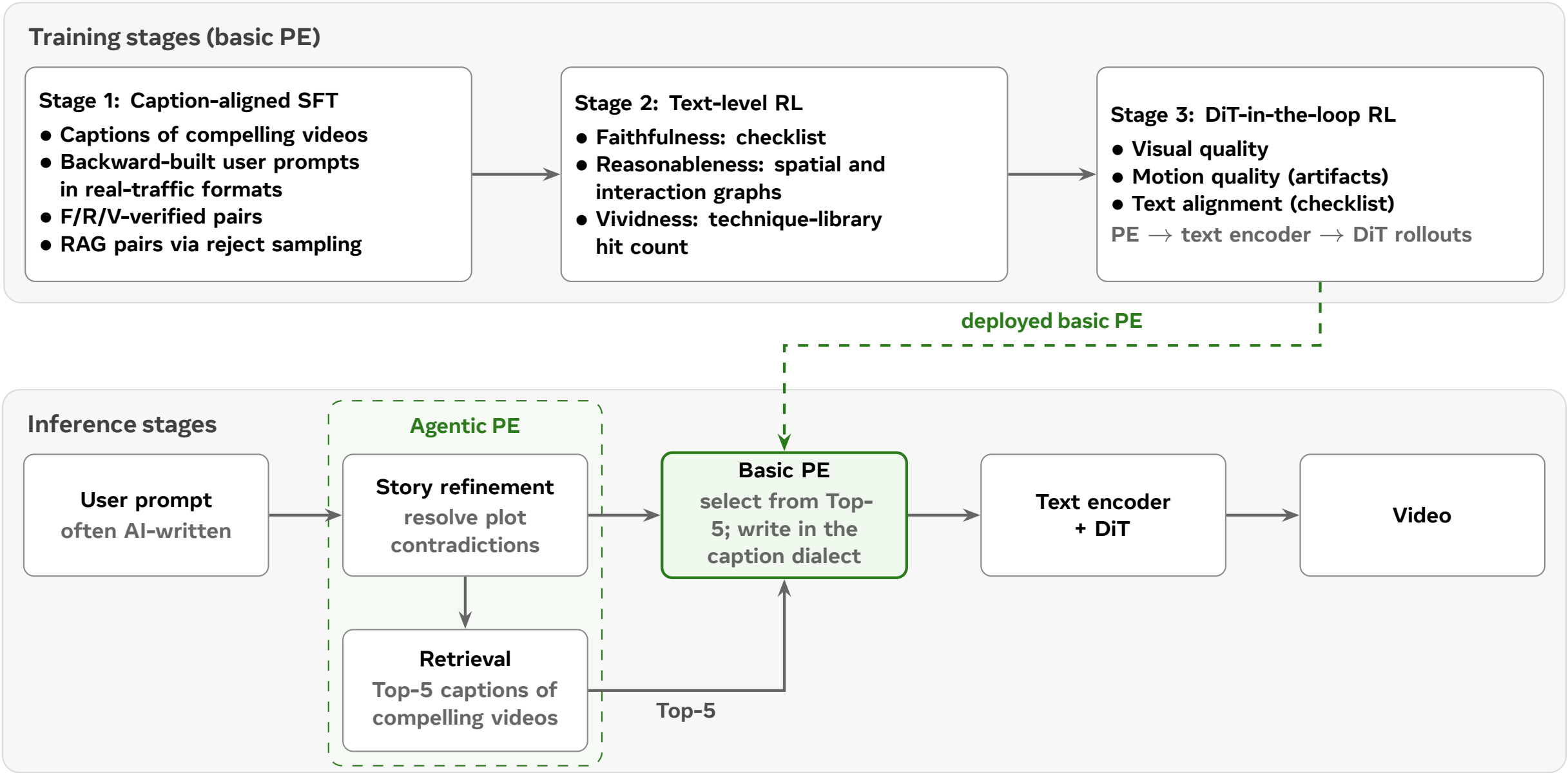

We split PE into two modules (Figure~\ref{fig:pe_overview}).  \emph{Basic PE}
is the trained rewriter that produces the final condition in the caption dialect.
\emph{Agentic PE} runs before it at inference and decides what that condition
should contain: it makes the story coherent and retrieves compelling examples.
Our experience across Kling releases can be distilled into five lessons, the first
three concerning basic PE and the last two concerning agentic PE.
(1)~PE output must match captions in wording and sentence patterns as well as in
format; we achieve this by constructing data backward from captions of compelling
videos.
(2)~Much of the gain comes from making the text faithful, reasonable, and vivid,
which can be verified and optimized without running the generator.
(3)~Reinforcement learning (RL) with the DiT in the loop then aligns PE with
product objectives and stabilizes quality, at a measurable cost in peak vividness.
(4)~Contradictions in AI-written prompts are best resolved by a story-refinement
step before basic PE.
(5)~Retrieving compelling captions raises vividness, provided that training uses
strict rejection sampling.

\subsection{Basic PE: Caption Alignment through Backward Construction}
\label{subsection:pe_alignment}

A user request always contains text and may also include images,
videos, or audio clips as references.  Basic PE rewrites only the text.  It keeps
every reference pointer in the instruction intact but does not modify the
attached media, which reach the generator unchanged
(Section~\ref{section:ref_caption}).  The alignment problem discussed below
therefore concerns the text condition alone.

\begin{figure}[h]
\centering
\newcommand{\lvltag}[1]{\tcbox[on line, tcbox raise=-1pt, arc=2pt, boxsep=0pt, left=4pt, right=4pt, top=1.5pt, bottom=1.5pt, boxrule=0.4pt, colframe=black!25, colback=white, fontupper=\sffamily\bfseries\scriptsize\color{black!75}]{#1}\hspace{3pt}}
\begin{tikzpicture}[
  font=\footnotesize,
  badbox/.style={draw=badborder!45, fill=badbg, rounded corners=3.5pt, inner sep=7pt, text width=7.38cm, align=left},
  goodbox/.style={draw=klinggreen!55, fill=klingbg, rounded corners=3.5pt, inner sep=7pt, text width=7.38cm, align=left}
]
  \node[badbox] (bad) at (0,0) {
    {\sffamily\bfseries\color{badtxt}\xmark\;\;Generic LLM rewrite (Out-of-Distribution)}\\[5pt]
    \lvltag{Format} Free paragraphs opening with a summary and closing with a mood statement; length varies with verbosity\\[4pt]
    \lvltag{Lexicon} \emph{``A breathtaking cinematic shot slowly reveals the scene''}\\[4pt]
    \lvltag{Sentence patterns} \emph{``As if time itself had paused, she gazes outward, lost in distant memories''}
  };

  \node[goodbox, right=0.38cm of bad] (good) {
    {\sffamily\bfseries\color{klinggreen}\cmark\;\;Caption-aligned PE output (In-Distribution)}\\[5pt]
    \lvltag{Format} Subject, action, scene, camera, lighting, and audio in caption field order; shot timestamps; task-regime length\\[4pt]
    \lvltag{Lexicon} \emph{``The camera pushes in slowly from a medium shot to a close-up of the woman's face''}\\[4pt]
    \lvltag{Sentence patterns} \emph{``The woman in the red coat stands still and looks out of the window; her expression changes from calm to sad''}
  };
\end{tikzpicture}
\caption{Three levels of caption alignment contrasting generic LLM rewrites with caption-aligned PE.}
\label{fig:pe_alignment_levels}
\end{figure}
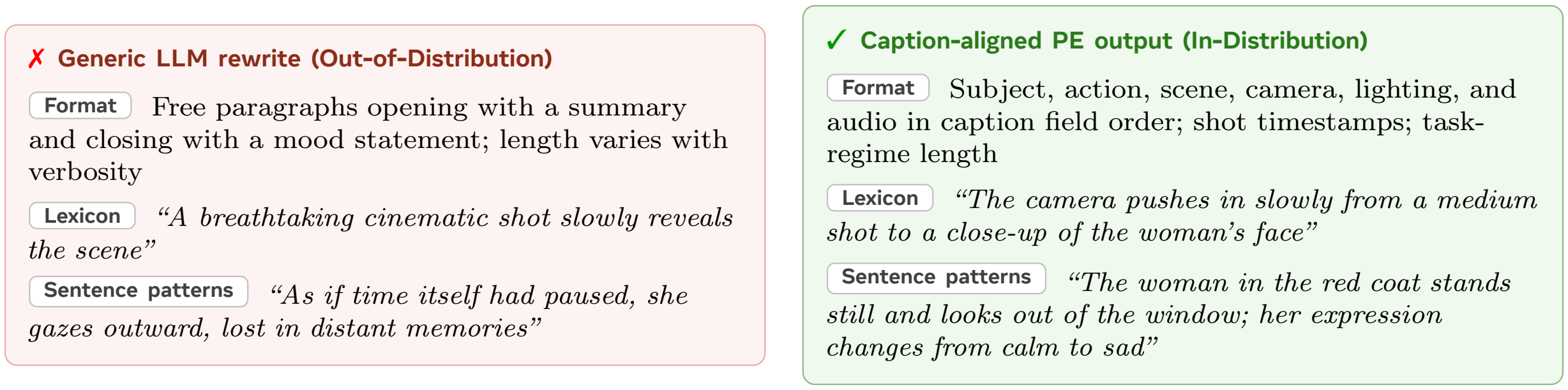

Two conditions with similar semantics can produce visibly different videos when
their length, slot order, terminology, or sentence structure differ.  A generic
LLM rewriter tends to write metaphors, evaluative adjectives, motivational
clauses, and long subordinate sentences.  Such text reads well but lies outside
the caption distribution: the generator spends conditioning capacity on
language patterns it has rarely seen and may ignore controls phrased in unfamiliar
terms.  We therefore require alignment at three levels: \emph{format}, meaning
field order, length regime, and shot and timestamp markers; \emph{lexicon},
meaning the camera, lighting, and action terms of the caption pipeline; and
\emph{sentence patterns}, meaning the declarative, present-tense,
entity-explicit sentences of production captions.  Matching format alone leaves
the other two levels free to drift.  Figure~\ref{fig:pe_alignment_levels}
contrasts typical generic rewrites with caption-aligned output at each level.

\vspace{4pt}

The most reliable way to achieve alignment at all three levels is to use real captions as training
targets.  We construct the data backward: starting from the caption of a real
video, a teacher model writes the user prompt that could have produced it, and
PE learns the mapping from prompt to caption.  Synthetic prompts follow the
statistics of real traffic rather than the teacher's own style.  For each pair
we sample a target length, language mixture, register, typo rate, and set of
omitted slots from logged queries, and include references, multi-shot requests,
and explicit exclusions at their observed frequencies.  Prompts that leak the
caption's skeleton, for example by listing every action phase in caption order,
are rejected, since PE would otherwise learn copying instead of completion.
Exclusions follow the same principle.  Captions describe what is visible, so PE
rewrites ``no people'' as ``an empty street with unobstructed pavement'' and
routes exclusions without a positive equivalent either to a negative-conditioning
field or to sampling-time rejection.

Backward construction also fixes the creative ceiling of PE, because the model
learns to write like its target captions.  If source captions are a random
sample of training videos, most targets describe ordinary footage with static
framing and uneventful action, and PE learns to expand every request into a
competent but flat description.  We therefore draw source captions mainly from
videos that are compelling in themselves, such as professionally shot scenes,
performances with clear dramatic beats, and clips with deliberate camera
design, selected using aesthetic and motion scores and the vividness metric
of Section~\ref{subsection:pe_text}.  Domain and length quotas prevent this
selection from collapsing onto a few genres.

\begin{observationpanel}
\textbf{PE inherits the temperament of its target captions.}  When source
captions are drawn uniformly from the training corpus, PE outputs are well
formed and faithful but dull: different requests converge to similar generic
framing, even lighting, and uneventful action.  Choosing compelling source
videos addresses this in the data, before any change to the model or the
objective.
\end{observationpanel}

\subsection{Text-Level Quality: Faithfulness, Reasonableness, and Vividness}
\label{subsection:pe_text}

Optimizing PE through the generator is expensive: every rollout requires video
sampling and judging, and video-level rewards are noisy.  A large fraction of
the end-to-end gain is available earlier, from the text alone.  Reasonableness
deserves particular attention.  The DiT is trained almost exclusively on real
or carefully produced videos, so it has rarely seen a condition that describes
an impossible scene.  When the text contains a spatial contradiction, an object
that changes identity, or an interaction without a physical cause, the
generator has no supervised behavior to fall back on and improvises freely.
The result appears as identity drift, inconsistent layouts, objects passing
through one another, or hands that never make contact.  Unreasonable text is
thus one of the most direct causes of consistency and interaction failures.

We measure three properties, each scored on a scale from 1 to 5, and use them in two
places: every supervised fine-tuning (SFT) pair must pass all three verifiers, and the same scores serve
as rewards for text-level RL.

\textbf{Faithfulness} measures whether the output preserves user intent.  A
judge decomposes the user prompt into a checklist of atomic requirements
covering entities, attributes, counts, actions, spatial relations, camera
instructions, quoted text, audio, reference bindings, and exclusions.  It then
marks each item in the PE output as satisfied, violated, or unaddressed.
Violations weigh more than omissions, and an addition that contradicts an item
counts as a violation.  The checklist makes the score attributable, because a
regression can be traced to the requirement type that failed.

\textbf{Reasonableness} measures whether the described video could exist.  The
judge reads the output sentence by sentence and incrementally builds two
graphs.  The \emph{spatial graph} links entities by relative position, contact,
containment, and viewpoint over time; the \emph{interaction graph} links
agents, actions, and patients together with their preconditions and resulting
states.  Each new sentence is checked against both graphs and against world
knowledge: whether positions remain consistent, whether an object used in an
action is present, whether a result follows from its cause, and whether the
number of events fits the duration.  Each conflict lowers the score and is
reported together with the sentence that introduced it.  For example, in ``the
man sets the cup on the table and walks to the door; he takes a sip as he opens
it,'' the spatial graph places the cup on the table and the man at the door, so
the final interaction lacks its object.  A generator conditioned on this text
must either duplicate the cup or teleport it, and both outcomes appear as
artifacts.

\textbf{Vividness} measures expressive richness.  The judge extracts expressive
elements along several axes, including performance and facial expression,
camera movement and shot design, lighting and color, pacing, and sound design.
It matches them against a curated library of professional techniques compiled
from compelling videos and cinematographic references, and the score increases
with the number of distinct, contextually appropriate hits.  Hits are
deduplicated per axis and capped, and only elements that pass the
reasonableness check are counted, so stacking technical terms does not raise
the score.

Text-level RL uses GRPO to optimize a reward that combines the three
scores~\cite{shao2024deepseekmath}.  Faithfulness should gate the other terms: an
output that violates an explicit requirement earns no credit for vividness,
which prevents the policy from trading intent for spectacle.  A distribution
term can additionally penalize outputs that fall outside the caption length and
vocabulary regime, protecting the alignment of
Section~\ref{subsection:pe_alignment}.  Because no video is rendered, this stage
is cheap enough to cover many prompts and iterations, and it precedes any
DiT-in-the-loop training.

\begin{insightpanel}
\textbf{Solve text problems in text.}  Faithfulness, reasonableness, and
vividness can be verified without rendering a single frame.  Removing these
defects before involving the generator makes end-to-end RL cheaper and its
rewards easier to interpret, because the remaining failures are more likely to
be genuine generation limits.
\end{insightpanel}

\subsection{DiT-in-the-Loop RL}
\label{subsection:pe_dit_rl}

Text-level judges approximate what the generator needs, but they cannot observe
how a specific checkpoint responds to a condition.  In the final training stage,
we connect PE to the frozen text encoder and DiT, sample several rewrites per
prompt, render a video for each, and update PE with a group-relative policy
gradient on video-level rewards.  All three rewards come from the decision
reward model of Section~\ref{section:reward_model}: visual quality and motion
quality, the latter covering structural, physical, and temporal artifacts, are
read from its taxonomy fields, and text alignment between the rendered video
and the user prompt is scored by its checklist fields
(Section~\ref{subsection:rm_rubric}).
This stage aligns PE with product
objectives and checkpoint-specific behavior, including which camera
moves the current model executes reliably, while reducing quality variance
across prompts and seeds.

The order of the stages matters.  DiT-in-the-loop RL starts from the text-level RL
checkpoint, so its rollouts are already faithful, reasonable, and vivid, and
the video rewards spend their limited sample budget on checkpoint-specific
behavior instead of on defects that a text judge could have caught.  Because
each update requires rendering several videos per prompt, this stage uses far
fewer prompts than text-level RL, and those prompts are chosen to cover product
domains and the failure slices found in production.

\begin{observationpanel}
\textbf{DiT-in-the-loop RL raises the floor and lowers the ceiling.}  After
this stage, severe failures become rarer and worst-case quality improves.  At
the same time, the most impressive outputs become less impressive: PE learns to
avoid complex choreography, large camera moves, and dense interactions, which
are precisely the descriptions with the highest artifact risk.
\end{observationpanel}

A plausible explanation is that all three rewards penalize risk.  None of them
measures whether a video is exciting, and a conservative condition is the
easiest way to avoid a low score.  Candidate remedies include keeping the
vividness reward in the end-to-end objective, regularizing the policy toward
the text-level RL checkpoint, and selecting checkpoints based on ceiling as well as floor
metrics.  Rewarding ambition directly without reintroducing artifacts remains
an open problem.

\subsection{Agentic PE: Story Refinement and Retrieval}
\label{subsection:pe_agentic}

Basic PE is trained primarily for faithfulness.  Reasonableness and vividness
act on it as floors: its output must be no less reasonable and no less vivid
than the user's input, and every detail it adds must itself be reasonable.  It
is not, however, the place where the input itself is rethought: repairing a
contradiction that the user wrote would require departing from that input.
Agentic PE takes that role at inference.
It runs before basic PE and performs two steps: story refinement makes the
requested plot coherent, and retrieval supplies captions of compelling videos
as creative evidence.  Its output is still a user-level specification plus
optional context; basic PE always writes the final condition, so the caption
dialect learned in Sections~\ref{subsection:pe_alignment}--\ref{subsection:pe_dit_rl}
is preserved regardless of what agentic PE adds.  This division keeps creative
decisions inspectable and lets each part be evaluated against its own
objective.

\paragraph{Story Refinement}
\label{subsection:pe_story}

In production, many user prompts are themselves written by AI models through
scripting assistants or upstream agent pipelines.  These prompts are long and
fluent but often internally inconsistent: a character's clothing changes
between paragraphs, a scene is both indoors and outdoors, more events are
planned than the duration allows, or an action depends on an object that was
never introduced.  Asking basic PE alone to repair them places two conflicting
goals on one model, since faithfulness asks it to preserve what the user wrote
while reasonableness asks it to change it.

Story refinement resolves this conflict upstream.  It works at the level of
plot and leaves caption language to basic PE.  It detects contradictions with
the same spatial and interaction graphs used for reasonableness scoring,
applies minimal edits that restore coherence while keeping the core intent, and
passes the revised prompt on.  Basic PE treats the refined prompt as
authoritative and concentrates on translation into the caption distribution,
without pressure to silently rewrite the story.

Story refinement follows three rules.  First, explicit hard constraints such
as quoted dialogue, counts, reference bindings, and requested camera moves are
never edited unless two of them conflict, in which case the refiner makes the
smallest change that resolves the conflict and records the decision.  Second, its edits are minimal and
local: the refiner reorders, merges, or drops events and fixes attributes, but
does not introduce new characters or change the premise.  Third, its output is
checked by the reasonableness judge and by a faithfulness checklist computed
on the non-conflicting part of the original prompt, before it reaches basic PE.
When no coherent edit exists, the refiner can pass the prompt through unchanged
with a conflict flag, so that the product can ask the user instead of guessing.

\paragraph{Retrieval of Compelling Captions}
\label{subsection:pe_rag}

Retrieval raises vividness further by showing basic PE how compelling videos
with similar content are actually described~\cite{gao2025rapo}.  The index
contains captions from the same pool of compelling videos used for backward
construction.  For each refined request, agentic PE retrieves the top five
captions and passes them as context.  Basic PE decides which camera patterns,
performance details, lighting setups, or sound designs suit the request.  Letting the model choose among several candidates avoids dependence on a
single top hit, which often matches the topic but not the intended action or
tone.

Training basic PE to use this context requires strict rejection sampling.
Retrieved captions contain specific entities, scene layouts, brands, and
dialogue.  If training targets copy such details, the model learns to import
content the user never requested, a form of hallucination that the faithfulness checklist detects only
partially.  For each training prompt we sample multiple candidates with
retrieval context and reject any candidate that introduces an entity,
attribute, or event that cannot be traced to either the user prompt or a
deliberate and compatible borrowing from a retrieved caption.  Surviving
candidates must also pass all three text verifiers.  The training set further
includes cases in which all five retrieved captions are irrelevant and the
accepted target ignores them, so that the model treats retrieval as optional
evidence.

\subsection{Experiments}
\label{subsection:pe_experiments}

Table~\ref{tab:pe_results} reports the contribution of each component.  Text
metrics are computed on held-out real user prompts.  For faithfulness,
reasonableness, and vividness, we report the average score on a 1--5 scale, and the share of
outputs scoring at least 4, which reflects how often an output is directly
usable.  Video metrics use a fixed generator, sampler, and seed policy and
report the overall scores of StoryEval~\cite{wang2024storyeval},
T2AV-Compass~\cite{cao2026t2avcompass}, AVE-Compass~\cite{wen2026avecompass},
and MultiRef-Compass~\cite{zhang2026multirefcompass}.  StoryEval (averaged across seven scores) tests whether
consecutive events are completed in order; T2AV-Compass (averaged across IF and Realism scores) covers joint audio-video
generation; AVE-Compass (averaged across Editing Intent, Fidelity Preserving, and Realism scores) covers audiovisual editing; and MultiRef-Compass (averaged across IF and RC scores)
covers multi-reference generation.  For editing and multi-reference prompts, PE
rewrites the instruction while preserving reference pointers, following
Section~\ref{section:ref_caption}.  Rows are cumulative: each adds one
component to the row above.

\begin{table}[t]
\centering
\footnotesize
\caption{Cumulative contribution of PE components.}
\setlength{\tabcolsep}{4pt}
\begin{tabular}{l cc cc cc cccc}
\toprule
\multirow{3}{*}{\textbf{Setting}} & \multicolumn{6}{c}{\textbf{Text metrics}} & \multicolumn{4}{c}{\textbf{Video metrics}} \\
\cmidrule(lr){2-7} \cmidrule(lr){8-11}
& \multicolumn{2}{c}{\textbf{Faithfulness}} & \multicolumn{2}{c}{\textbf{Reasonableness}}
& \multicolumn{2}{c}{\textbf{Vividness}} & \multirow{2}{*}{\makecell{\textbf{StoryEval}}} & \multirow{2}{*}{\makecell{\textbf{T2AV-}\\\textbf{Compass}}} & \multirow{2}{*}{\makecell{\textbf{AVE-}\\\textbf{Compass}}} & \multirow{2}{*}{\makecell{\textbf{MultiRef-}\\\textbf{Compass}}} \\
\cmidrule(lr){2-3} \cmidrule(lr){4-5} \cmidrule(lr){6-7}
& \textbf{Avg.} & \textbf{\(\geq\)4 (\%)} & \textbf{Avg.} & \textbf{\(\geq\)4 (\%)} & \textbf{Avg.} & \textbf{\(\geq\)4 (\%)}
& ~ & ~ & ~ & ~ \\
\midrule
\multicolumn{11}{l}{\emph{Basic PE}} \\
SFT, random-source captions      & 4.02 & 75.42 & 3.95 & 69.17 & 3.72 & 62.33 & 65.42 & 70.15 & 56.2 & 4.6137 \\
\quad + compelling-video sources & 4.00 & 74.92 & 3.92 & 68.08 & 4.03 & 72.58 & 66.87 & 70.60 & 56.7 & 4.6224 \\
\quad + F/R/V data verification  & 4.27 & 80.83 & 4.18 & 74.33 & 4.11 & 75.08 & 71.93 & 73.07 & 57.9 & 4.6591 \\
\quad + text-level RL            & 4.45 & 86.67 & 4.31 & 79.25 & 4.25 & 79.83 & 75.58 & 75.34 & 58.4 & 4.7012 \\
\quad + DiT-in-the-loop RL       & 4.38 & 83.50 & 4.42 & 83.67 & 4.23 & 81.42 & 76.46 & 75.91 & 59.3 & 4.7163 \\
\midrule
\multicolumn{11}{l}{\emph{Agentic PE + Basic PE}} \\
Hailuo IR (over Basic PE)        & 4.19 & 77.17 & 4.51 & 85.83 & 4.21 & 80.08 & 73.38 & 74.62 & 58.1 & 4.6987 \\
Gemini 4 Argon (over Basic PE)   & 4.34 & 81.75 & 4.57 & 88.58 & 4.30 & 83.17 & 75.73 & 75.37 & 58.8 & 4.7098 \\
Agentic, random few-shot + reflection & 4.43 & 85.25 & 4.48 & 85.50 & 4.22 & 81.33 & 76.79 & 76.12 & 59.5 & 4.7142 \\
\quad + story refinement         & 4.41 & 84.58 & 4.67 & 91.92 & 4.26 & 82.25 & 77.53 & 76.03 & -- & 4.7249 \\
\quad + retrieval (Top-5)        & 4.36 & 82.17 & 4.65 & 91.42 & 4.44 & 87.33 & 77.72 & 75.68 & -- & 4.7158 \\
\bottomrule
\end{tabular}
\label{tab:pe_results}
\end{table}

\subsection{Discussion}

PE translates user intent into the language the generator was trained on, and
it does so through two modules.  Basic PE owns the caption dialect.  Backward
construction from compelling videos aligns its output with captions in format,
wording, and sentence patterns, while also setting its creative ceiling.
Faithfulness, reasonableness, and vividness verification, applied first as a
data filter and then as text-level rewards, removes the defects most likely to
make the generator improvise.  DiT-in-the-loop RL aligns basic PE with product
objectives and raises the quality floor, at a current cost in peak vividness.
Agentic PE owns the content: story refinement resolves contradictions in
AI-written prompts, and retrieval supplies compelling examples, so that no
single module carries every responsibility.  Two directions remain open:
rewards that credit ambitious but executable content, and joint training of
agentic PE and basic PE against shared verifiers.

Several extensions of agentic PE are ongoing work. In
experimental versions it calls tools to edit the user's image, video, or audio
references, or to supply information that the request leaves out.  We are also
exploring low-resolution previews that expose composition and motion before
full-resolution generation, 3D previsualization that fixes layout and camera
paths in advance, and post-generation verification that checks the output
against the request and triggers revision when it fails.  The main bottlenecks
are latency and cost, since every tool call, preview, or verification pass adds
inference time and compute to a request.  We expect these costs to fall as
inference efficiency improves.

\section{Text Encoder: Perceptual Grounding for Conditioning}
\label{section:text_encoder}
Captioning, RGC, and PE establish the explicit semantic contract of our
pipeline, translating visual evidence and user intent into dense, structured
text (Sections~\ref{section:caption}--\ref{section:pe}).  Yet a critical
last-mile question remains: \emph{does the model that reads this text actually
understand what it describes?}  The text encoder (TE) plays this pivotal role by
converting the text prompt into the continuous conditioning representations
consumed by the DiT.  At the level of the generator, this conditioning interface
can be written abstractly as
\begin{equation}
    \mathbf{H}_{\mathrm{cond}}=\Phi_{\theta}(c),
    \qquad
    \widehat{v}=G_{\phi}\!\left(\mathbf{H}_{\mathrm{cond}},\boldsymbol{\epsilon}\right),
    \label{eq:te_interface}
\end{equation}
where $c$ denotes the text prompt, $\Phi_{\theta}$ the TE, $G_{\phi}$ the video generator, and $\boldsymbol{\epsilon}$ the sampling
randomness.  Because $G_{\phi}$ observes the prompt solely through the sequence
representation $\mathbf{H}_{\mathrm{cond}}$, any visual or temporal semantics
left unresolved by $\Phi_{\theta}$ cannot reach the generator, no matter how
faithfully $c$ is written upstream.

To make $\mathbf{H}_{\mathrm{cond}}$ a reliable conditioning signal for video generation,
the TE must internalize visual, spatial, and temporal knowledge and express
semantics across multiple granularities---spanning fine-grained lexical details,
mid-level relational layouts, and global narrative structure.  Large text
encoders offer substantial representational potential: their parameter
counts provide ample capacity, while extensive pre-training exposes them to broad
lexical, relational, and conceptual structures.  However, this potential does not automatically translate
into effective video conditioning.  Existing approaches
typically rely directly on off-the-shelf TE weights and expose the DiT to
features from only a single encoder layer, even when inserting small-parameter
auxiliary modules such as lightweight connectors or token refiners between the
TE and the DiT~\cite{kong2024hunyuanvideo}.
Using unadapted backbone weights while collapsing the encoder output to a single
layer leaves much of its representational capacity unused and discards
distinctions needed for video generation.

We therefore unlock this representational potential at both the backbone and
interface levels by adapting the encoder for perceptual grounding and
extracting a multi-granular, training-stable conditioning sequence.  Across
Kling iterations, our experience can be distilled into four core lessons:
\begin{itemize}[leftmargin=*,nosep]
    \item \textbf{Sequence-level evaluation rather than general proxies.} General QA
    benchmarks and single-vector retrieval fail to reflect how a DiT consumes
    full sequence representations.
    \item \textbf{Perceptual adaptation rather than reliance on off-the-shelf backbones.}
    Current foundation models prioritize agentic skills over perception, and even MLLMs
    are trained on broad task mixtures.  These trends motivate adaptation for spatiotemporal grounding.
    \item \textbf{Multi-granular semantic conditioning.} Video generation
    requires preserving literal details, relations, temporal structure, and
    global narrative context in one usable condition.  The multi-layer
    representations of the TE provide complementary levels of abstraction that
    support these granularities.
    \item \textbf{A stable conditioning interface.} The multi-layer conditioning
    representation must remain stable as the diffusion generator progresses
    across noise regimes and training stages.
\end{itemize}

\subsection{Representation Evaluation Beyond General Proxies}
\label{subsection:te_selection}

Selecting a TE requires measuring whether its frozen hidden states resolve a
prompt as a fine-grained scene specification.  However, common lightweight
proxies---general comprehension benchmarks and retrieval metrics---are
structurally mismatched with how a DiT uses the encoder~\cite{li2025granted}.
When serving as a TE, the encoder performs only a prefill forward pass and
supplies a sequence of token-level hidden states $\mathbf{H}_{\mathrm{cond}}$ to
the DiT.  General QA and comprehension benchmarks instead evaluate
\emph{autoregressive decoding}, measuring generated token outputs rather than
the underlying hidden states used for conditioning.  Cross-modal retrieval,
meanwhile, evaluates \emph{single-vector matching}, pooling the entire prompt
into one vector rather than preserving the sequence dimension that the DiT
attends to throughout generation.  Evaluating a TE for video conditioning
should therefore operate directly on sequence-level hidden
states~\cite{li2025granted}.

\begin{observationpanel}
\textbf{General evaluation proxies are structurally mismatched with DiT conditioning.}
A DiT conditions on the full sequence of prefill hidden states, whereas QA benchmarks evaluate autoregressively decoded outputs and retrieval metrics evaluate pooled single-vector representations; neither proxy directly measures the representation interface consumed by the generator.
\end{observationpanel}

Effective TE screening should therefore probe the sequence of prefill hidden
states on minimally contrasting descriptions~\cite{li2025granted}.  Such
fine-grained semantic contrasts enable rapid text-only filtering of candidate
backbones and post-training checkpoints before costly end-to-end DiT training.

\subsection{Adapting Foundation Backbones for Video Conditioning}
\label{subsection:te_adaptation}

Current foundation models increasingly prioritize agentic capabilities over
multimodal perception.  Even among multimodal models, training spans a wide
variety of task types with a heavy emphasis on visually grounded agent workflows.  While
such broad supervision produces versatile general-purpose assistants, it does
not specialize the backbone's hidden states for the dense spatiotemporal
grounding required by video conditioning.  Empirical evidence supports
this distinction: although multimodal pre-training improves generative
conditioning over pure-text LLMs, broad multi-task instruction tuning offers no
reliable gain, and reasoning-oriented post-training can even
degrade prefill hidden states when it shifts capacity from descriptive
grounding toward step-by-step deduction~\cite{li2025granted}.

We therefore strengthen the TE through targeted multimodal adaptation rather than
relying on off-the-shelf weights.  We build on Kling-MLLM
(Section~\ref{sec:mm-understanding-foundation}), which already provides stronger
spatiotemporal perception than general Qwen backbones, and further adapt it using
dense video descriptions and perceptual understanding tasks
(Section~\ref{section:caption}).  This adaptation enhances the perceptual
representation capacity of $\Phi_{\theta}$ in Eq.~\ref{eq:te_interface},
allowing its hidden states to more faithfully encode the visual and temporal
details specified in the prompt.

\begin{insightpanel}
\textbf{Perceptual adaptation strengthens the TE for video conditioning.}
Training the backbone on dense descriptions and spatiotemporal perception tasks enhances its ability to resolve visual and temporal details in its hidden states, providing stronger conditioning representations.
\end{insightpanel}

\subsection{Multi-Granular Semantic Conditioning}
\label{subsection:te_fusion}

Beyond adapting the backbone weights, effective conditioning also depends on how
features are extracted from the TE.  Video-generation prompts are inherently
multi-granular and hierarchical: at the local level, they specify visible text,
appearance, and instantaneous actions; at the relational level, they bind
attributes to entities and encode spatial or temporal interactions; at the
global level, they organize scene context, shot progression, and narrative
intent.  The TE exhibits a
related hierarchy across its depth, with different layers capturing
complementary levels of lexical, relational, and conceptual
abstraction~\cite{gurnee2023language,li2026semanticrouting}.  Exposing only a single layer to the
DiT therefore creates a structural mismatch between the multi-granular prompt
and the representation supplied to the generator.  Instead, the conditioning
interface should aggregate information across encoder layers so that local
details remain anchored to relational structure and global scene context.
\begin{insightpanel}
\textbf{A multi-granular prompt requires a multi-layer condition.}
Aggregating complementary encoder layers preserves lexical details, entity
relations, temporal structure, and global context in one sequence representation
rather than forcing a single layer to carry every level of abstraction.
\end{insightpanel}

\subsection{Stable Conditioning During Diffusion Training}
\label{subsection:te_stability}
Combining features across multiple TE layers typically relies on a learnable
fusion module optimized alongside the DiT.  However, diffusion training is
inherently non-stationary: the generator primarily optimizes coarse-grained
structure in early training and gradually shifts toward fine-grained details in
later stages, while training configurations such as timestep sampling and video
resolution also change across stages.  Because different TE layers capture
different semantic granularities, these shifting optimization targets
continuously pull the fusion module toward different layer combinations.  Rather
than converging to a stationary weighting, the fusion module exhibits
\emph{fusion instability}---its layer weights continue to drift or fluctuate
throughout training~\cite{li2025granted,li2026semanticrouting}.

Even when the TE backbone itself is frozen, an unstable fusion module makes the
fused representation $\mathbf{H}_{\mathrm{cond}}$ non-stationary.  As the fusion
weights drift across training stages, the conditioning distribution supplied to
the DiT shifts accordingly, forcing the generator to repeatedly re-adapt to the
same text prompt.  Stabilizing the fusion module is therefore essential to
prevent conditioning drift and maintain a consistent semantic interface across
all stages of diffusion training~\cite{li2025granted,li2026semanticrouting}.

\begin{insightpanel}
\textbf{Unstable multi-layer fusion creates a non-stationary condition.}
As diffusion training shifts from coarse structure to fine-grained details across stages, learnable fusion weights tend to drift; stabilizing the fusion module keeps the condition from becoming a moving target for the DiT.
\end{insightpanel}

\subsection{Empirical Evaluation}
\label{subsection:te_experiments}

We evaluate candidate TEs by training the video generator on dense captions
(Section~\ref{section:caption}) under strictly controlled settings---using the
same DiT backbone, training data stream, and number of training steps---and
comparing the resulting generated videos directly instead of relying on
training loss.  On a stratified internal benchmark, raters compare videos
produced by paired TE configurations side by side and record Good/Same/Bad
(GSB) judgments along three dimensions: overall preference, instruction
following, and aesthetics.  The three Kling-MLLM configurations cumulatively
introduce multimodal adaptation (Section~\ref{subsection:te_adaptation}),
multi-granular conditioning (Section~\ref{subsection:te_fusion}), and
conditioning stability (Section~\ref{subsection:te_stability}).  Each
configuration is compared against the same Qwen text-encoder baseline rather
than only against the preceding row, using the preference ratio
\((G+S)/(S+B)\), where values above \(1.00\) indicate net gains over the
baseline.

\begin{table}[H]
\centering
\small
\caption{Human preference evaluation (\((G+S)/(S+B)\)) relative to the Qwen baseline.}
\label{tab:te_gsb}
\renewcommand{\arraystretch}{1.08}
\setlength{\tabcolsep}{11pt}
\begin{tabular}{@{}l ccc@{}}
\toprule
\textbf{Kling-MLLM TE configuration} &
\textbf{Overall} &
\textbf{Instruction Following} &
\textbf{Aesthetics} \\
\midrule
Multimodal adaptation              & $1.10$ & $1.04$ & $1.11$ \\
\quad + Multi-granular conditioning & $1.22$ & $1.11$ & $1.22$ \\
\quad + Conditioning stability     & $\mathbf{1.27}$ & $\mathbf{1.14}$ & $\mathbf{1.25}$ \\
\bottomrule
\end{tabular}
\end{table}

Table~\ref{tab:te_gsb} shows that all three refinements yield consistent gains
over the Qwen baseline across every dimension, with performance improving
monotonically as multimodal adaptation, multi-granular conditioning, and
conditioning stability are added.  Notably, the gains in overall preference and
aesthetics are even larger than those in instruction following, indicating that
stronger perceptual encoding and stable multi-layer conditioning improve not
only prompt adherence but also overall visual quality.

\subsection{Discussion}

The TE is more than a passive linguistic lookup module attached to the DiT: it
is the component that maps multimodal understanding into the continuous
conditioning space of the generator.  While captioning, RGC, and PE express
multimodal understanding explicitly as structured text, the TE converts that
text into an implicit, multi-granular scene representation---screened at the
sequence level, adapted for spatiotemporal perception, aggregated across
encoder layers, and stabilized throughout diffusion training.

\section{Reward Models: Turning Understanding into an Optimization Signal}
\label{section:reward_model}

The preceding sections construct a semantic contract across the video generation
pipeline: captioning specifies what training videos contain, RGC states what
should be preserved or modified, PE translates user requests into that
contract, and the TE determines which distinctions reach the generator.
Post-training and inference-time selection require an explicit optimization
direction to enforce this contract: among candidate rollouts, the system must
determine which video should be assigned a higher probability and identify
which specific regions or time intervals failed and why.  A reward model (RM)
supplies this executable utility signal.

\subsection{Objectives, Challenges, and Application Scenarios}
\label{subsection:rm_objectives}

Building MLLM-based video RMs across successive Kling generations exposed three
challenges:
\begin{enumerate}[leftmargin=*]
    \item \textbf{Perceptual bandwidth before reasoning depth.}  For
    generated-video defects, visual perception is a tighter bottleneck than
    language reasoning~\cite{lin2025forensicchat}.  Scaling a general-purpose
    MLLM or eliciting a long chain of thought (CoT) does not improve artifact
    detection if the vision tower is insensitive to low-level synthesis
    distortions, or if transient anatomical and physical anomalies are erased
    by sparse frame sampling and token compression~\cite{artifactbench2026}.
    \item \textbf{Dimensional entanglement and loser-centric attribution bias.}
    Video quality decomposes into at least four axes---visual quality (VQ),
    motion quality (MQ), consistency, and text alignment (TA)---each with finer
    sub-dimensions
    (Table~\ref{tab:rm_taxonomy})~\cite{videoreward2025human}.  In
    human pairwise annotations, however, dimensions are strongly coupled with overall
    preference, and defect attributions are \emph{loser-centric}: annotators
    reliably record the decisive flaw of the losing video but frequently leave
    the winning video's defects unannotated.  Treating unannotated winners as clean negatives
    or deriving criteria from candidate contrasts with metadata
    shortcuts~\cite{liu2026rever}, corrupts the decision boundary.
    \item \textbf{The tension between interpretability and fast decision-making.}
    Diagnostic evaluation requires structured rationales with spatial bounding
    boxes and temporal intervals~\cite{he2025videoscore2}, whereas
    online RL for DiTs and PE
    models requires scoring thousands of rollouts across dozens of dimensions
    with minimal latency.  Free-form CoT generation is computationally
    prohibitive for online RL and prone to tie-abstention on close pairs,
    whereas linear scalar heads discard semantic grounding and suffer from
    variance collapse.
\end{enumerate}

\begin{table}[!t]
\centering
\scriptsize
\caption{Application scenarios of our RMs.}
\setlength{\tabcolsep}{4pt}
\renewcommand{\arraystretch}{1.0}
\begin{tabularx}{\linewidth}{>{\raggedright\arraybackslash}p{0.15\linewidth} l l
  >{\raggedright\arraybackslash\hsize=0.9\hsize}X
  >{\raggedright\arraybackslash\hsize=1.2\hsize}X
  >{\raggedright\arraybackslash\hsize=0.9\hsize}X}
\toprule
\rowcolor{klingbg}
\textbf{Scenario} & \textbf{Model} & \textbf{Regime} & \textbf{Input} & \textbf{Output} & \textbf{Key constraint} \\
\midrule
DiT RL & DRM & Online
  & Rollouts of one prompt
  & Per-dimension rewards (GRPO); preference pairs (DPO)
  & Latency \\
\addlinespace[2pt]
PE RL & DRM & Online
  & Videos rendered from candidate rewrites
  & Per-rewrite reward on VQ, MQ, and TA
  & TA judged against the user prompt \\
\addlinespace[2pt]
Auxiliary machine evaluation & GRM & Offline
  & Single video with prompt or reference
  & Defect tags, intervals, boxes, and rationale
  & Verifiable evidence \\
\addlinespace[2pt]
Checkpoint selection & DRM \& GRM & Offline
  & Videos from competing checkpoints
  & Per-dimension win rates; defect profiles
  & Low variance \\
\addlinespace[2pt]
Data filtering & DRM & Offline
  & Single clip
  & Calibrated defect probabilities for keep/drop
  & Throughput; low false positives \\
\bottomrule
\end{tabularx}
\vspace{-2pt}
\label{tab:rm_applications}
\end{table}

To trade off latency against explainability, we split reward modeling between two
complementary models: a \textbf{General Reward Model (GRM)} that produces
explainable, spatiotemporally localized critiques, and a \textbf{Decision
Reward Model (DRM)} that scores many dimensions in a single prefill.  Together
they serve five production scenarios (Table~\ref{tab:rm_applications}):
\textbf{(1)~DiT RL} with GRPO and
DPO~\cite{shao2024deepseekmath,rafailov2023direct,videoreward2025human};
\textbf{(2)~PE RL} with the generator in the loop
(Section~\ref{subsection:pe_dit_rl}); \textbf{(3)~auxiliary machine
evaluation} that supplies human raters with localized evidence;
\textbf{(4)~checkpoint selection} across capability slices; and
\textbf{(5)~data filtering} of pre-training and SFT corpora.  The scenarios
differ in what is queried, how the output is consumed, and whether they run
online or offline; Table~\ref{tab:rm_applications} summarizes these
differences and the model used for each scenario.

\vspace{-3pt}
\subsection{Model Design}
\label{subsection:rm_architecture}
\label{subsection:rm_rubric}

\begin{figure}[!t]
\centering
\includegraphics[width=\linewidth]{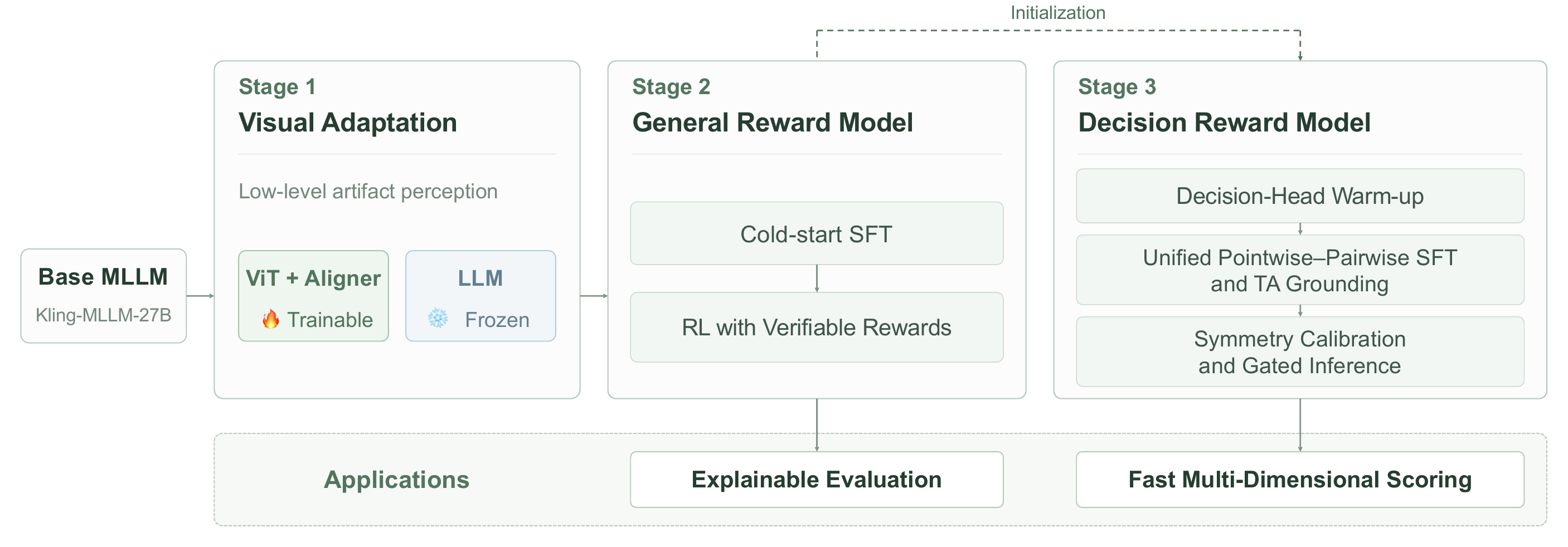}
\caption{Three-stage training pipeline for our RMs, starting from Kling-MLLM-27B. \textbf{Stage~1} adapts the vision encoder and aligner with the LLM frozen for low-level artifact perception; \textbf{Stage~2} trains the General Reward Model (GRM) with cold-start SFT and RL with verifiable rewards for explainable evaluation; \textbf{Stage~3} initializes the Decision Reward Model (DRM) from the GRM and performs decision-head warm-up, unified pointwise--pairwise SFT with TA grounding, and symmetry calibration with gated inference for fast multi-dimensional scoring.}
\label{fig:rm_overview}
\end{figure}

Built on Kling-MLLM-27B
(Section~\ref{sec:mm-understanding-foundation}), our pipeline unifies all
supervision sources under a hierarchical evaluation taxonomy
(Table~\ref{tab:rm_taxonomy}) spanning four Level-1 categories, ten Level-2
dimensions, and fine-grained leaf defect tags.  Each L2 dimension is scored on
its own rather than only through an overall preference, so that a reward can
be traced to the dimension that produced it and a dimension can be added when
optimization exposes a blind spot.  Training progresses through three stages
(Figure~\ref{fig:rm_overview}).

\begin{observationpanel}
\vspace{-3pt}
\textbf{Penalizing artifacts alone rewards stillness.}
A generator optimized against defect penalties can lower its risk by moving
less, and a slow or frozen video contains few anatomical or physical errors.
Motion magnitude is therefore scored as a dimension of its own and acts as a
counterweight in DiT RL.
\end{observationpanel}

\vspace{-4pt}
\paragraph{Stage 1: Visual Adaptation for Low-Level Artifact Perception.}
General-purpose vision encoders are pre-trained for semantic recognition, which
encourages invariance to high-frequency texture noise, frame-to-frame flicker,
and local geometric distortions~\cite{lin2025forensicchat}.  In Stage~1, we
freeze the LLM and optimize only the vision encoder and
aligner on two tasks: (1)~\textbf{VAE reconstruction discrimination}, pairing
natural clips with VAE-decoded counterparts to sensitize the encoder to subtle
blur, codec, and smoothing artifacts; and (2)~\textbf{templated defect probes}
derived from RL-pair and single-video defect logs.
Frame sampling is tailored to the dimension being judged: densely sampled
frames are used for motion and artifact dimensions, whereas sparsely sampled
frames at native high resolution are used for VQ dimensions.  Retaining a
$20\%$ Stage~1 replay in
later stages prevents catastrophic forgetting of low-level visual discrimination.

\begin{observationpanel}
\vspace{-3pt}
\textbf{A judge cannot reason about a defect it does not see.}
Longer chains of thought do not recover an artifact that the vision encoder
smooths away or that sparse frame sampling skips, and many anatomical and
physical anomalies last under two seconds.  Perception has to be trained
before judgment.
\end{observationpanel}

\paragraph{Stage 2: GRM for Explainable Evaluation with Localized Evidence.}
Stage~2 unfreezes the backbone and trains the GRM on three sources: the training
splits of open-source datasets (their validation and test splits are never
used); our human-annotated forensic corpus, which specifies a defect category,
a temporal interval $[t_{\mathrm{start}}, t_{\mathrm{end}}]$, per-frame
bounding boxes, and a rationale; and rubric-derived checklist VQA instances
written without access to preference labels~\cite{liu2026rever}.
The GRM is trained in two steps:
\begin{itemize}[leftmargin=*]
    \item \textbf{Step 1: Cold-start SFT.}
    We fine-tune the GRM to produce structured critiques, defect
    evidence tubes (each consisting of a temporal interval with per-frame boxes), and ordinal
    ratings in a fixed output format; the probability-weighted expectation
    of the rating serves as its scalar score.  Training uses a $1:10$
    vision-to-LLM learning-rate ratio and a $20\%$ Stage~1 replay so that
    the low-level perception acquired in Stage~1 is preserved.
    \item \textbf{Step 2: RL with verifiable rewards.}
    We optimize $G=8$ rollouts per video via GRPO~\cite{shao2024deepseekmath}
    with a verifiable reward that is gated by output-format validity and
    combines four checkable terms: ordinal rating calibration against human
    scores, defect-presence correctness, tag-set $\mathrm{F1}$ against the
    annotated defect categories, and spatiotemporal $\mathrm{mIoU}$ of the
    predicted intervals and bounding boxes.  Unmatched
    ground-truth defects are penalized, so the policy cannot raise its
    reward by reporting fewer defects.
\end{itemize}

\begin{insightpanel}
\textbf{Charge a judge for what it misses.}
Scoring only reported defects rewards omission. Penalizing unmatched ground-truth defects makes missed and incorrect reports equally costly.
\end{insightpanel}

\begin{table}[!tb]
\centering
\scriptsize
\caption{Unified evaluation taxonomy of our RMs.}
\setlength{\tabcolsep}{3.5pt}
\renewcommand{\arraystretch}{0.95}
\begin{tabularx}{\linewidth}{l l >{\raggedright\arraybackslash}p{0.12\linewidth} >{\raggedright\arraybackslash}X}
\toprule
\rowcolor{klingbg}
\textbf{L1 Category} & \textbf{L2 Evaluation Dimension} & \textbf{Input} & \textbf{Representative L3/L4 Leaf Defects} \\
\midrule
\multirow[t]{3}{*}{\textbf{1. Visual Quality}}
  & 1.1 Image Clarity \& Compression & Video & Low clarity/defocus blur, codec blockiness \& mosquito noise \\
  & 1.2 Visual Content Integrity & Video & AI plastic/greasy skin, texture smearing, edge halo, text distortion, watermark, exposure \\
  & 1.3 Visual Aesthetics & Video & Color/lighting harmony, spatial composition, material texture, subject styling, set design \\
\midrule
\multirow[t]{3}{*}{\textbf{2. Motion Quality}}
  & 2.1 Basic Motion Quality & Video & Anatomy, deformation/liquefaction, temporal drift, collision, physics, flicker \\
  & 2.2 Motion Aesthetics \& Narrative & Video & Weak pacing, acting, editing, framing, camera path, or creativity \\
  & 2.3 Motion Magnitude & Video & Insufficient motion amplitude, abnormal freezing \\
\midrule
\multirow[t]{2}{*}{\textbf{3. Consistency}}
  & 3.1 Input Image Consistency & Video + input/ref image & First-frame color/style shift, subject ID drift, text preservation \\
  & 3.2 Multi-Shot Consistency \& Coherence & Multi-shot video & Cross-shot action, spatial axis, narrative, scene, appearance, color \\
\midrule
\multirow[t]{2}{*}{\textbf{4. Text Alignment}}
  & 4.1 Basic Prompt Following & Video + prompt & Core action, temporal order, details, speed, subject count, camera, spatial logic, OCR \\
  & 4.2 Multi-Shot Prompt Following & Multi-shot video + prompt & Shot count omission/addition, shot order swap, cut timing/transition \\
\bottomrule
\end{tabularx}
\label{tab:rm_taxonomy}
\end{table}

\paragraph{Stage 3: DRM for Fast Multi-Dimensional Scoring.}
Initialized from the GRM, Stage~3 trains the DRM, a unified evaluator that
answers $K$ fine-grained questions---taxonomy defect checks, atomic
prompt-verification items, and pairwise dimension comparisons---in a single
prefill, with a readout adapted from
Intern-Decision~\cite{internlm2026interndecision}.  Its \emph{decision head}
is the option-letter rows of the language-model head, read at the placeholder
positions; no parameters are added.  A prefill holds one or two videos.
Scoring $N$ rollouts pointwise costs $N$ prefills; in the two-video form,
pointwise fields name the video they inspect (Video~1 or Video~2) and pairwise
fields compare the two, so per-video defect checks, prompt-verification items,
and dimension comparisons are answered in one pass, and the shared context is
markedly more reliable on subtle motion defects than isolated single-video
scoring.  The template below shows the two-video form.
\begin{tcolorbox}[colback=gray!8, colframe=gray!40, boxrule=0.4pt, arc=2pt,
                  left=5pt, right=5pt, top=4pt, bottom=4pt, boxsep=0pt]
\ttfamily\scriptsize
\setlength{\parindent}{0pt}
{\rmfamily\bfseries [user]}\\
\#\# State: Video 1: <video> | Video 2: <video> | Prompt: ...\\
\#\# Guidelines: ...\\
\#\# Decision schema:\\
\hspace*{1.5em}1. Human Anatomy (Video 1): ... A=no | B=yes\\
\hspace*{1.5em}2. Human Anatomy (Video 2): ... A=yes | B=no\\
\hspace*{1.5em}3. Subject Action (Video 1, "a cat leaps onto the chair"): ... A=yes | B=partial | C=no\\
\hspace*{1.5em}4. Subject Action (Video 2, "a cat leaps onto the chair"): ... A=no | B=partial | C=yes\\
\hspace*{1.5em}5. Motion Quality - Worse: ... A=Video 1 | B=Video 2 | C=Tie\\
\hspace*{1.5em}6. Motion Quality - Better: ... A=Video 1 | B=Video 2 | C=Tie\\[3pt]
{\rmfamily\bfseries [assistant]}\\
\{"Human Anatomy (Video 1)": <|fim\_pad|>, "Human Anatomy (Video 2)": <|fim\_pad|>, ...,\\
\hspace*{0.5em}"Motion Quality - Worse": <|fim\_pad|>, "Motion Quality - Better": <|fim\_pad|>\}
\end{tcolorbox}
At each placeholder, normalizing the option-letter logits over that field's
options yields a continuous score and probability in one forward pass.
Because placeholders are fixed in the prefix rather than decoded
autoregressively, earlier answers cannot leak into later fields, reinforced by
option-letter randomization, field permutation, and placeholder-only loss
masking.  The DRM is trained in three steps:
\begin{itemize}[leftmargin=*]
    \item \textbf{Step 1: Decision-Head Warm-up.}
    With the backbone frozen, we train only the decision head on generic
    multiple-choice corpora (VQA, preference, and tool-use samples), then
    unfreeze the full model on video critiques with vote-aggregated soft
    labels and on temporal localization.  Each field is supervised with soft
    cross-entropy over its options, plus a ranked probability score (RPS) term
    for ordinal fields.
    \item \textbf{Step 2: Unified Pointwise--Pairwise SFT and TA Grounding.}
    We fine-tune jointly on single-video defect and checklist items and on
    two-video packs that carry per-video and pairwise fields, so that defect
    sensitivity and pairwise discrimination are learned in the same prefill.
    Text alignment is strengthened with instruction-following pairs and with
    audiovisual checklists comprising several atomic questions per video.  As an
    auxiliary task, a randomly sampled subset of checklist items in each
    training sample is paired with temporal fields, one binary field per
    one-second bin, to mark when the prompted event occurs; this anchors
    alignment decisions to the frames that support them and improves the
    judgment of event timing and temporal order.
    \item \textbf{Step 3: Symmetry Calibration and Gated Inference.}
    Causal/linear attention and M-RoPE~\cite{wang2024qwen2} induce position
    bias across video order~\cite{zheng2023judging}, and
    the \emph{worse} and \emph{better} fields of one dimension can disagree;
    averaging over swapped orders and both polarities costs two to four passes
    per pair.  We instead place both polarities in one prefill and regularize
    training with (i)~a \emph{symmetric KL} between the \emph{worse} field and
    the option-swapped \emph{better} field of each dimension and (ii)~a
    \emph{cross-order Bradley--Terry loss}~\cite{bradley1952rank} between
    original and order-swapped packs.  At inference, a confidence gate invokes
    the swapped pass only on low-margin cases, matching multi-pass accuracy at
    near-single-pass latency; when invoked, the second pass uses complementary
    dense/sparse frame sampling to catch transient collapses within the
    context limit.
\end{itemize}

\begin{insightpanel}
\textbf{Decide in the prefix, not in the decode.}
Fixing every answer slot in the prompt allows $K$ questions to be answered in
one prefill and keeps each answer independent of the others; the same mechanism
that reduces latency also removes the ordering shortcuts a decoded answer
sequence would learn.
\end{insightpanel}

\subsection{Experimental Evaluation}
\label{subsection:rm_experiments}

We evaluate our models on an internal pairwise benchmark covering nine of the ten L2
dimensions of Table~\ref{tab:rm_taxonomy} (all except 1.1 Image Clarity \&
Compression) and on three public benchmarks, VideoScore-Bench-v2,
VideoPhy2-test, and VideoGen-RewardBench.
The baselines include prompted MLLMs (Gemini-3.8-Flash and Gemini-4-Argon) and
open-source video reward models~\cite{videoreward2025human,wang2025unifiedreward,bansal2025videophy2,he2024videoscore,he2025videoscore2,wu2023dover}.

\begin{table}[t]
\centering
\scriptsize
\caption{Pairwise preference accuracy (\%) on the internal benchmark.}
\setlength{\tabcolsep}{2.5pt}
\renewcommand{\arraystretch}{0.9}
\resizebox{\linewidth}{!}{%
\begin{tabular}{l cc cc ccc cc cc}
\toprule
& \multicolumn{2}{c}{\textbf{Avg}}
& \multicolumn{2}{c}{\textbf{VQ}}
& \multicolumn{3}{c}{\textbf{MQ}}
& \multicolumn{2}{c}{\textbf{Consistency}}
& \multicolumn{2}{c}{\textbf{TA}} \\
\cmidrule(lr){2-3} \cmidrule(lr){4-5} \cmidrule(lr){6-8} \cmidrule(lr){9-10} \cmidrule(lr){11-12}
\textbf{Model}
& \multirow{2}{*}{\textbf{Macro}} & \multirow{2}{*}{\textbf{Micro}}
& \textbf{Visual Content} & \textbf{Visual}
& \textbf{Basic Motion} & \textbf{Motion Aesth.} & \textbf{Motion}
& \textbf{Input Image} & \textbf{Multi-Shot}
& \textbf{Basic Prompt} & \textbf{Multi-Shot} \\
& & & \textbf{Integrity} & \textbf{Aesthetics}
& \textbf{Quality} & \textbf{\& Narrative} & \textbf{Magnitude}
& \textbf{Consistency} & \textbf{Consistency}
& \textbf{Following} & \textbf{Prompt Fol.} \\
\midrule
\multicolumn{12}{l}{\emph{Prompting MLLM}} \\
Gemini-3.8-Flash$^{\dagger}$ & $67.59$ & $67.13$ & $66.27$ & $67.03$ & $61.99$ & $69.81$ & $88.89$ & $\underline{75.00}$ & $58.93$ & $47.10$ & $\underline{73.33}$ \\
Gemini-4-Argon$^{\dagger}$   & $73.54$ & $72.59$ & $72.78$ & $76.92$ & $65.42$ & $72.96$ & $85.71$ & $\underline{75.00}$ & $66.07$ & $60.30$ & $\mathbf{86.67}$ \\
\midrule
\multicolumn{12}{l}{\emph{Ours}} \\
GRM              & $\underline{74.72}$ & $\underline{73.32}$ & $\underline{75.15}$ & $\underline{79.67}$ & $\underline{70.40}$ & $\underline{75.47}$ & $\underline{93.65}$ & $72.50$ & $\underline{71.43}$ & $\underline{65.30}$ & $68.89$ \\
DRM$^{\ddagger}$ & $\mathbf{77.86}$ & $\mathbf{75.53}$ & $\mathbf{78.11}$ & $\mathbf{81.32}$ & $\mathbf{72.27}$ & $\mathbf{78.62}$ & $\mathbf{96.83}$ & $\mathbf{77.50}$ & $\mathbf{75.89}$ & $\mathbf{66.90}$ & $\underline{73.33}$ \\
\bottomrule
\end{tabular}%
}
\par\vspace{2pt}
{\raggedright\scriptsize $^{\dagger}$ Accessed through an internal API gateway; the macro accuracy of the same Gemini model varies by up to $1.8$ points across gateways.
$^{\ddagger}$ Motion dimensions are judged without the prompt; all others are judged with the prompt.\par}
\label{tab:rm_internal_bench}
\end{table}

\begin{table}[!t]
\centering
\scriptsize
\caption{Accuracy (\%) on open-source video reward and scoring benchmarks.}
\setlength{\tabcolsep}{8pt}
\renewcommand{\arraystretch}{0.88}
\begin{tabular}{l ccc}
\toprule
\textbf{Model} & \textbf{VideoScore-Bench-v2} & \textbf{VideoPhy2-test} & \textbf{VideoGen-RewardBench} \\
\midrule
\multicolumn{4}{l}{\emph{Prompting MLLM}} \\
Gemini-3.8-Flash & $36.90$ & $\mathbf{47.80}$ & $54.37$ \\
Gemini-4-Argon   & $33.60$ & $43.70$ & $56.00$ \\
\midrule
\multicolumn{4}{l}{\emph{Reward/Scoring Models for Video}} \\
VideoReward~\cite{videoreward2025human}     & --      & $31.33$ & $\mathbf{59.69}$ \\
UnifiedReward~\cite{wang2025unifiedreward}  & $25.07$ & $22.02$ & $53.31$ \\
VideoPhy2~\cite{bansal2025videophy2}        & $22.35$ & $37.64$ & $30.75$ \\
VideoScore1.1~\cite{he2024videoscore}       & $38.41$ & $27.95$ & $16.79$ \\
VideoScore2~\cite{he2025videoscore2}        & $44.35$ & $33.58$ & $51.53$ \\
Dover~\cite{wu2023dover}                    & $34.00$ & $28.21$ & $54.27$ \\
\midrule
\multicolumn{4}{l}{\emph{Ours}} \\
GRM & $\underline{51.60}$ & $38.53$ & $55.24$ \\
DRM & $\mathbf{52.00}$ & $\underline{44.23}$ & $\underline{57.33}$ \\
\bottomrule
\end{tabular}
\label{tab:rm_opensource_bench}
\end{table}

On the internal benchmark (Table~\ref{tab:rm_internal_bench}), the DRM
achieves the highest macro and micro accuracies and leads on eight of the nine
dimensions; Gemini-4-Argon remains stronger only on multi-shot prompt
following.  The GRM already surpasses Gemini-4-Argon in macro accuracy, and
Stage~3 adds a further $3.1$ points only at a single-prefill inference cost.  On public
benchmarks (Table~\ref{tab:rm_opensource_bench}), the DRM achieves leading performance on VideoScore-Bench-v2, VideoPhy2-test, and VideoGen-RewardBench.

\subsection{Discussion}
\label{subsection:rm_discussion}

The RMs close the chain that the preceding sections open.  The taxonomy that
organizes human annotation also names the fields the DRM scores, the defects
the GRM localizes, and the slices on which checkpoints and training data are
screened, so a failure surfaced by any consumer of the reward can be attributed
to a specific dimension and traced to its upstream source.  A reward signal,
however, is also the quantity a generator is optimized against, and it cannot
by itself reveal the shortcuts that optimization finds;
Section~\ref{section:benchmark} turns to evaluation protocols that remain
independent of it.

\section{Automatic Evaluation: Measuring What Each Task Requires}
\label{section:benchmark}

Automatic evaluation serves as the measurement layer for video generation. 
While training loss can indicate optimization failures, it cannot directly assess whether generated videos satisfy task-specific requirements, such as correct actions, consistent identity preservation, or audiovisual alignment. 
These properties must instead be evaluated on the generated outputs. 
Accordingly, the objective is not to design a single universal video quality score, but to determine four fundamental aspects of evaluation: the target task, the failure modes that matter, the tools capable of detecting them, and the reliability of the resulting measurements.
Table~\ref{tab:benchmark_task_taxonomy} summarizes the first three aspects for each major task; measurement reliability is discussed below.

\subsection{Task Taxonomy: What Should Be Evaluated?}

The evaluation protocol should be defined by the underlying generation task. 
Different tasks have distinct success criteria and correspondingly different failure modes. 
More importantly, they differ in the ways in which models can achieve high evaluation scores without genuinely satisfying the intended objective.

\paragraph{\textbf{Captioning.}}
Lexical-overlap metrics such as BLEU and CIDEr penalize valid paraphrases and are poorly suited to dense video descriptions. 
Recent benchmarks therefore shift from wording similarity to semantic coverage, decomposing video content into atomic keypoints and verifying whether each fact is correctly expressed. 
VidCapBench~\cite{chen2025vidcapbench} exemplifies this paradigm with 643 videos and 10,644 QA pairs spanning aesthetics, content, motion, and physical laws, while distinguishing items suitable for automatic evaluation from those requiring human judgment.
CapRiCorn-1K~\cite{chen2026capricorn} extends this paradigm to long videos by also evaluating referential consistency, testing whether entities remain consistently identified across time and shots.
For comparative captioning, factual coverage alone is insufficient: similarities and differences must be evaluated separately, as in ViDiC~\cite{wu2025vidic}.
Caption evaluation should also demonstrate downstream validity. 
VidCapBench reports correlations of 0.89--0.92 between caption accuracy and downstream text-to-video generation quality, while CapRiCorn-1K shows that caption quality correlates strongly with long-video reasoning and video reconstruction performance.
These results highlight a key criterion for caption evaluation: a metric should guide data or captioner design only if improvements in the metric translate into measurable downstream gains.

\begin{table}[t]
\centering
\footnotesize
\caption{Evaluation taxonomy by task.}
\renewcommand{\arraystretch}{1.20}
\setlength{\tabcolsep}{5pt}
\begin{tabularx}{\linewidth}{@{}L{0.17\linewidth} L{0.22\linewidth} L{0.28\linewidth} Y@{}}
\toprule
\rowcolor{klingbg}
\textbf{Task} &
\textbf{Core Objective} &
\textbf{Main Failure Modes} &
\textbf{Representative Tools} \\
\midrule
\textbf{Video captioning} &
Accurate \& consistent coverage of salient facts &
Omissions, hallucinations, vague references, lost temporal structure &
Keypoint QA, entailment judges, coreference metrics, downstream generation tests \\
\rowcolor{black!2}
\textbf{T2V / T2AV} &
Prompt adherence \& perceptual/physical plausibility &
Missing attributes/events, implausible motion, visual artifacts, wrong/desynced audio &
Checklist judges, DOVER++, aesthetic/audio quality models, CLIP/CLAP, Synchformer, lip-sync \\
\textbf{Reference-conditioned} &
Identity/detail preservation, binding, \& natural integration &
Identity drift, entity swaps, attribute leakage, source mismatch, copy--paste &
Detectors/segmenters, ID embeddings, face/speaker models, binding checks, anti-paste judges \\
\rowcolor{black!2}
\textbf{Keyframe / storyboard} &
Prescribed frames in correct order, timing, \& duration &
Missed/reordered frames, hard cuts, flashes, freezes, repeats, static storyboard &
DINO/SSIM/PSNR matching, Kendall rank correlation, window matching, camera/tracking \\
\textbf{Video / A-V editing} &
Execute requested changes while preserving unedited content &
Non-response, incomplete edit, total regeneration, collateral changes, A-V mismatch &
Atomic edit checklists, response detection, source--target tracking/seg., SSIM/DINO \\
\bottomrule
\end{tabularx}
\label{tab:benchmark_task_taxonomy}
\end{table}

\paragraph{\textbf{Text-driven Generation.}}
Text-to-video evaluation should distinguish instruction following, perceptual quality, and physical or commonsense plausibility. 
VBench~\cite{huang2024vbench} decomposes video quality into 16 dimensions rather than collapsing them into a single score, while VBench-2.0~\cite{zheng2025vbench2} further extends evaluation to human fidelity, controllability, creativity, physics, and commonsense.
For compositional prompts, attributes, spatial relations, interactions, and motion are evaluated with specialized tools, as no single similarity metric reliably captures all these aspects~\cite{sun2024t2vcompbench}.
Audio introduces additional dimensions that must be evaluated independently. 
T2AV-Compass~\cite{cao2026t2avcompass} and LongAV-Compass~\cite{liu2026longav} jointly assess video and audio quality, text--video and text--audio alignment, audiovisual synchronization, instruction following, and realism. 
Their results reveal substantial asymmetry across these dimensions: strong video quality does not necessarily imply realistic audio or consistent material--timbre correspondence. 
This distinction is fundamental: instruction-following checklists measure whether the requested content is present, whereas realism criteria assess whether that content is perceptually and physically plausible.

\paragraph{\textbf{Reference and Keyframe Control.}}
Reference-conditioned generation requires more than embedding similarity. 
Evaluation should account for entity identity, fine-grained details, role binding, and natural integration. 
MultiRef-Compass~\cite{zhang2026multirefcompass} further introduces an anti-paste coefficient,
\begin{equation}
    \mathrm{EF}_{\mathrm{final}}
    = \mathrm{EF}_{\mathrm{base}} P_{\mathrm{paste}},
\end{equation}
which penalizes direct copying of reference content and prevents copy--paste behavior from inflating fidelity scores.
Keyframe-conditioned generation introduces an additional temporal dimension: detecting whether a keyframe appears is necessary but insufficient. 
KeyFrame-Compass~\cite{tang2026keyframecompass} therefore decomposes keyframe response into presence, visual fidelity, temporal order, temporal position, persistence, and response uniqueness. 
This decomposition distinguishes successful temporal integration from failure modes such as slideshow-like transitions, transient flashes, prolonged freezes, and reversed sequences. 
The same principle generalizes to other temporally controlled settings, including first- and last-frame-conditioned generation, shot planning, and camera trajectory control.

\paragraph{\textbf{Editing.}}
Video editing must satisfy two competing objectives: the requested content should change, while unrelated content should remain intact. 
CoVEBench~\cite{wu2026covebench} decomposes complex editing instructions into atomic checks and separately evaluates instruction compliance, output quality, and source fidelity. 
AVE-Compass~\cite{wen2026avecompass} extends this evaluation to joint audio--video editing. 
Its Editing Intent score is defined for each case and each modality as
\begin{equation}
    \mathrm{EI}^{(c)}_M
    = \mathrm{IF}^{(c)}_M \mathrm{FP}^{(c)}_M,
\end{equation}
where IF measures instruction following and FP measures fidelity preservation. 
The multiplicative formulation ensures that neither leaving the source unchanged nor regenerating it indiscriminately can achieve a high score. 
Response gating provides an additional safeguard: preservation and realism are evaluated only when the target modality responds to the edit, while non-responses are retained as instruction-following failures.

\subsection{Evaluation Tools and Reliability: How Can Each Property Be Measured Accurately?}
No single tool observes every property of a generated video. 
Signal-based and embedding-based metrics such as DOVER++, CLIP, CLAP, and Synchformer are inexpensive and stable, yet they capture only global quality and similarity.
Specialist perception models, including detectors, trackers, and identity or speech models, add localized evidence about where and when an entity or change appears. 
MLLM checklist judges~\cite{he2025videoscore2} cover the remaining semantic requirements that resist simple measurement. 
In practice, each property is routed to the simplest tool able to observe it reliably, with specialist evidence grounding the semantic judgments.

Accurate evaluation, however, depends as much on protocol design as on the tools themselves. 
Holistic ratings should be replaced by atomic, directly observable criteria, and competing requirements should be scored jointly so that trivial shortcuts---such as leaving a video unedited or pasting a reference---cannot inflate the result. 
Frame rate, resolution, and token budget belong to the protocol rather than to implementation details; a genuine temporal metric must change when temporal evidence is removed or shuffled~\cite{ok2025tempcore}.
Automatic judges also require calibration against human annotations and explicit control of known biases~\cite{zheng2023judging}, together with a fixed configuration for reproducibility.
Finally, reporting results per dimension and difficulty slice exposes weaknesses that a single aggregate score would hide, while hidden prompts and periodic human review guard against benchmark overfitting.

\subsection{Operational Evaluation: How Should Evaluation Be Used in Practice?}

The same evaluation taxonomy can be instantiated at three cost levels. 
An hourly smoke test relies on deterministic signal-based metrics and a small set of high-agreement checklist items.
A daily regression board adds specialist evidence, stable MLLM-based checks, a fixed anchor model, and performance deltas across difficulty levels. 
A release-level evaluation runs the full routed protocol on challenging public benchmarks and hidden prompts, with repeated judging and human calibration.
These levels differ in evaluator configuration and rigor rather than merely in the number of evaluated samples.

Evaluation outputs should ultimately support diagnosis. 
For each failure, the system should retain the evaluation criterion, sampled evidence, tool outputs, judge rationale, and likely upstream source. 
Caption failures should be traced back to annotation or captioning, reference-binding failures to conditioning, realism failures to generation or reward modeling, and evaluator disagreements to protocol calibration. 
A useful benchmark should therefore do more than rank models: it should identify which capability fails, under what conditions, and where in the generation pipeline the failure originates.

\section{Conclusion}
\label{section:conclusion}

This report summarizes the observations and insights into multimodal
understanding that we gained from successive generations of Kling,
viewing generation quality as the projection of understanding through the
entire data--conditioning--alignment--evaluation chain.  Kling-MLLM, a shared
multimodal foundation model, provides the perceptual foundation for every
downstream module.  Captioning serves as a dense semantic contract for training,
and RGC extends this contract with explicit reference bindings.  PE translates user intent
into this caption distribution, with agentic steps that refine the story and
retrieve compelling examples.  A reliable TE comes from perceptual training and
multi-layer fusion, and candidate TEs can be screened from text alone.  Reward
modeling pairs an explainable GRM with a fast DRM, treating perceptual bandwidth
as a primary design parameter.  For automatic
evaluation, our review suggests that protocols should follow the task, expose
scoring shortcuts, and help trace failures to their upstream source.  We hope these lessons offer practical guidance for
building the understanding stack that supports controllable video generation.

\bibliographystyle{kling/plainnat}
\bibliography{paper}

@article{wang2024qwen2,
  title={Qwen2-vl: Enhancing vision-language model's perception of the world at any resolution},
  author={Wang, Peng and Bai, Shuai and Tan, Sinan and Wang, Shijie and Fan, Zhihao and Bai, Jinze and Chen, Keqin and Liu, Xuejing and Wang, Jialin and Ge, Wenbin and Fan, Yang and Dang, Kai and Du, Mengfei and Ren, Xuancheng and Men, Rui and Liu, Dayiheng and Zhou, Chang and Zhou, Jingren and Lin, Junyang},
  journal={arXiv preprint arXiv:2409.12191},
  year={2024}
}

@inproceedings{li2024mvbench,
  title={Mvbench: A comprehensive multi-modal video understanding benchmark},
  author={Li, Kunchang and Wang, Yali and He, Yinan and Li, Yizhuo and Wang, Yi and Liu, Yi and Wang, Zun and Xu, Jilan and Chen, Guo and Luo, Ping and Wang, Limin and Qiao, Yu},
  booktitle={Proceedings of the IEEE/CVF Conference on Computer Vision and Pattern Recognition},
  pages={22195--22206},
  year={2024}
}

@article{zhang2024lmmsevalrealitycheckevaluation,
      title={LMMs-Eval: Reality Check on the Evaluation of Large Multimodal Models}, 
      author={Kaichen Zhang and Bo Li and Peiyuan Zhang and Fanyi Pu and Joshua Adrian Cahyono and Kairui Hu and Shuai Liu and Yuanhan Zhang and Jingkang Yang and Chunyuan Li and Ziwei Liu},
      year={2024},
      journal={arXiv preprint arXiv:2407.12772},
}

@article{rafailov2023direct,
  title={Direct preference optimization: Your language model is secretly a reward model},
  author={Rafailov, Rafael and Sharma, Archit and Mitchell, Eric and Manning, Christopher D and Ermon, Stefano and Finn, Chelsea},
  journal={Advances in neural information processing systems},
  volume={36},
  pages={53728--53741},
  year={2023}
}

@article{shao2024deepseekmath,
  title={Deepseekmath: Pushing the limits of mathematical reasoning in open language models, 2024},
  author={Shao, Zhihong and Wang, Peiyi and Zhu, Qihao and Xu, Runxin and Song, Junxiao and Bi, Xiao and Zhang, Haowei and Zhang, Mingchuan and Li, YK and Wu, Y and others},
  journal={URL https://arxiv. org/abs/2402.03300},
  volume={2},
  number={3},
  pages={5},
  year={2024}
}

@article{zheng2025vbench2,
  title={{VBench-2.0}: Advancing Video Generation Benchmark Suite for Intrinsic Faithfulness},
  author={Zheng, Dian and Huang, Ziqi and Liu, Hongbo and Zou, Kai and He, Yinan and Zhang, Fan and Gu, Lulu and Zhang, Yuanhan and He, Jingwen and Zheng, Wei-Shi and Qiao, Yu and Liu, Ziwei},
  journal={arXiv preprint arXiv:2503.21755},
  year={2025}
}

@article{bansal2025videophy2,
  title={{VideoPhy-2}: A Challenging Action-Centric Physical Commonsense Evaluation in Video Generation},
  author={Bansal, Hritik and Peng, Clark and Bitton, Yonatan and Goldenberg, Roman and Grover, Aditya and Chang, Kai-Wei},
  journal={arXiv preprint arXiv:2503.06800},
  year={2025}
}

@inproceedings{chen2025vidcapbench,
  title={{VidCapBench}: A Comprehensive Benchmark of Video Captioning for Controllable Text-to-Video Generation},
  author={Chen, Xinlong and Zhang, Yuanxing and Rao, Chongling and Guan, Yushuo and Liu, Jiaheng and Zhang, Fuzheng and Song, Chengru and Liu, Qiang and Zhang, Di and Tan, Tieniu},
  booktitle={Findings of the Association for Computational Linguistics: ACL 2025},
  pages={8543--8563},
  year={2025},
  doi={10.18653/v1/2025.findings-acl.449}
}

@article{chen2026capricorn,
  title={{CapRiCorn-1K}: A Comprehensive Benchmark for Video Captioning and Subject Referential Consistency Across Temporal Scales},
  author={Chen, Xinlong and Tang, Jiafu and Ding, Yue and Jia, Yizhuo and Li, Bozhou and Zeng, Bohan and Shi, Yang and Li, Shihao and Ji, Yiyan and Liu, Qiang and others},
  journal={arXiv preprint arXiv:2606.21949},
  year={2026}
}

@article{wu2025vidic,
  title={{ViDiC}: Video Difference Captioning},
  author={Wu, Jiangtao and Li, Shihao and Bian, Zhaozhou and Chen, Jialu and Wen, Runzhe and Ping, An and He, Yiwen and Wang, Jiakai and Zhang, Yuanxing and Liu, Jiaheng},
  journal={arXiv preprint arXiv:2512.03405},
  year={2025}
}

@inproceedings{huang2024vbench,
  title={{VBench}: Comprehensive Benchmark Suite for Video Generative Models},
  author={Huang, Ziqi and He, Yinan and Yu, Jiashuo and Zhang, Fan and Si, Chenyang and Jiang, Yuming and Zhang, Yuanhan and Wu, Tianxing and Jin, Qingyang and Chanpaisit, Nattapol and Wang, Yaohui and Chen, Xinyuan and Wang, Limin and Lin, Dahua and Qiao, Yu and Liu, Ziwei},
  booktitle={Proceedings of the IEEE/CVF Conference on Computer Vision and Pattern Recognition},
  pages={21807--21818},
  year={2024}
}

@inproceedings{sun2024t2vcompbench,
  title={{T2V-CompBench}: A Comprehensive Benchmark for Compositional Text-to-Video Generation},
  author={Sun, Kaiyue and Huang, Kaiyi and Liu, Xian and Wu, Yue and Xu, Zihan and Li, Zhenguo and Liu, Xihui},
  booktitle={Proceedings of the IEEE/CVF Conference on Computer Vision and Pattern Recognition},
  pages={8406--8416},
  year={2025}
}

@article{he2025videoscore2,
  title={{VideoScore2}: Think before You Score in Generative Video Evaluation},
  author={He, Xuan and Jiang, Dongfu and Nie, Ping and Liu, Minghao and Jiang, Zhengxuan and Su, Mingyi and Ma, Wentao and Lin, Junru and Ye, Chun and Lu, Yi and Wu, Keming and others},
  journal={arXiv preprint arXiv:2509.22799},
  year={2025}
}

@inproceedings{wang2024storyeval,
  title={Is Your World Simulator a Good Story Presenter? {A} Consecutive Events-Based Benchmark for Future Long Video Generation},
  author={Wang, Yiping and He, Xuehai and Wang, Kuan and Ma, Luyao and Yang, Jianwei and Wang, Shuohang and Du, Simon Shaolei and Shen, Yelong},
  booktitle={Proceedings of the IEEE/CVF Conference on Computer Vision and Pattern Recognition},
  year={2025},
  eprint={2412.16211},
  archivePrefix={arXiv}
}

@inproceedings{cao2026t2avcompass,
  title={{T2AV-Compass}: Towards Unified Evaluation for Text-to-Audio-Video Generation},
  author={Cao, Zhe and Wang, Tao and Wang, Jiaming and Wang, Yanghai and Zhang, Yuanxing and Chen, Jialu and Deng, Miao and Wang, Jiahao and Guo, Yubin and Liao, Chenxi and Zhang, Yize and Zhang, Zhaoxiang and Liu, Jiaheng},
  booktitle={International Conference on Machine Learning},
  year={2026},
  eprint={2512.21094},
  archivePrefix={arXiv}
}

@article{tang2026keyframecompass,
  title={{KeyFrame-Compass}: Towards Comprehensive Evaluation of Keyframe-Conditioned Video Generation},
  author={Tang, Yuqi and Liu, Tengfei and Lai, Yizheng and Wang, Yuran and Shi, Yang and Su, Wanshun and Zhang, Zhuoran and Wang, Qixun and Zhang, Xiaohan and Yu, Xinlei and Bai, Xuehai and Zhu, Xuanyu and Zeng, Bohan and Li, Bozhou and Li, Shujie and Dai, Yifan and Wei, Yujie and Liu, Shixuan and Wang, Haotian and Chen, Jialu and Zhang, Yuanxing},
  journal={arXiv preprint arXiv:2607.14202},
  year={2026}
}

@article{zhang2026multirefcompass,
  title={{MultiRef-Compass}: Towards Comprehensive Evaluation of Multi-Reference-to-Audio-Video Generation},
  author={Zhang, Xiaohan and Wen, Yuqing and Chen, Junlin and Tang, Yuqi and He, Yiting and Shao, Lizhuo and Zhu, Weiming and Liu, Tengfei and Shi, Yang and Chen, Jialu and Zhang, Yuanxing and Li, Huaxiong},
  journal={arXiv preprint arXiv:2607.14189},
  year={2026}
}

@article{wu2026covebench,
  title={{CoVEBench}: Can Video Editing Models Handle Complex Instructions?},
  author={Wu, Jiangtao and Wang, Jiaming and He, Yiwen and Zhang, Yuanxing and Li, Shihao and Liu, Dunyuan and Zhao, Xuedong and Chen, Jialu and Wang, Zekun Moore and Liu, Jiaheng},
  journal={arXiv preprint arXiv:2606.08415},
  year={2026}
}

@article{wen2026avecompass,
  title={{AVE-Compass}: Towards Holistic Evaluation for Audio-Video Editing Abilities},
  author={Wen, Yuqing and Huang, Yukai and Xie, Qianqian and Wu, Jiangtao and Lin, Yibin and Gu, Yikai and Chen, Jialu and Zhang, Yuanxing and Liu, Jiaheng},
  journal={arXiv preprint arXiv:2607.24821},
  year={2026}
}

@article{ok2025tempcore,
  title={{TempCore}: Are Video {QA} Benchmarks Temporally Grounded? A Frame Selection Sensitivity Analysis and Benchmark},
  author={Ok, Hyunjong and Lee, Jaeho},
  journal={arXiv preprint arXiv:2509.01167},
  year={2025}
}

@inproceedings{zheng2023judging,
  title={Judging {LLM}-as-a-Judge with {MT-Bench} and Chatbot Arena},
  author={Zheng, Lianmin and Chiang, Wei-Lin and Sheng, Ying and Zhuang, Tianle and Wu, Zhanghao and Zhuang, Yonghao and Lin, Zi and Li, Zhuo and Li, Dacheng and Xing, Eric P. and others},
  booktitle={Advances in Neural Information Processing Systems},
  volume={36},
  pages={46595--46623},
  year={2023}
}

@inproceedings{rohrbach2018object,
  title={Object Hallucination in Image Captioning},
  author={Rohrbach, Anna and Hendricks, Lisa Anne and Burns, Kaylee and Darrell, Trevor and Saenko, Kate},
  booktitle={Proceedings of the 2018 Conference on Empirical Methods in Natural Language Processing},
  pages={4035--4045},
  year={2018},
  doi={10.18653/v1/D18-1437}
}

@inproceedings{liu2025capability,
  title={{CAPability}: A Comprehensive Visual Caption Benchmark for Evaluating Both Correctness and Thoroughness},
  author={Liu, Zhihang and Xie, Chen-Wei and Wen, Bin and Yu, Feiwu and Chen, Jixuan and Li, Pandeng and Zhang, Boqiang and Yang, Nianzu and Li, Yinglu and Gao, Zuan and Zheng, Yun and Xie, Hongtao},
  booktitle={Advances in Neural Information Processing Systems, Datasets and Benchmarks Track},
  year={2025},
  eprint={2502.14914},
  archivePrefix={arXiv}
}

@article{kong2024hunyuanvideo,
  title={{HunyuanVideo}: A Systematic Framework for Large Video Generative Models},
  author={Kong, Weijie and Tian, Qi and Zhang, Zijian and Min, Rox and Dai, Zuozhuo and Zhou, Jin and Xiong, Jiangfeng and Li, Xin and Wu, Bo and Zhang, Jianwei and others},
  journal={arXiv preprint arXiv:2412.03603},
  year={2024}
}

@inproceedings{gao2025rapo,
  title={The Devil Is in the Prompts: Retrieval-Augmented Prompt Optimization for Text-to-Video Generation},
  author={Gao, Bingjie and Gao, Xinyu and Wu, Xiaoxue and Zhou, Yujie and Qiao, Yu and Niu, Li and Chen, Xinyuan and Wang, Yaohui},
  booktitle={Proceedings of the IEEE/CVF Conference on Computer Vision and Pattern Recognition},
  pages={3173--3183},
  year={2025}
}

@inproceedings{videoreward2025human,
  title={Improving Video Generation with Human Feedback},
  author={Liu, Jie and Liu, Gongye and Liang, Jiajun and Yuan, Ziyang and Liu, Xiaokun and Zheng, Mingwu and Wu, Xiele and Wang, Qiulin and Xia, Menghan and Wang, Xintao and Liu, Xiaohong and Yang, Fei and Wan, Pengfei and Zhang, Di and Gai, Kun and Yang, Yujiu and Ouyang, Wanli},
  booktitle={Advances in Neural Information Processing Systems},
  year={2025},
  eprint={2501.13918},
  archivePrefix={arXiv}
}

@article{artifactbench2026,
  title={{Artifact-Bench}: Evaluating {MLLMs} on Detecting and Assessing the Artifacts of {AI}-Generated Videos},
  author={Tang, Yuqi and Shi, Yang and Zhang, Zhuoran and Wang, Qixun and Bai, Xuehai and Ding, Yue and Chen, Ruizhe and Zeng, Bohan and Chen, Xinlong and Zhu, Xuanyu and Li, Bozhou and Wang, Yuran and Dai, Yifan and Tong, Chengzhuo and Liu, Xinyu and Ji, Yiyan and Wei, Yujie and Dong, Yuhao and Yan, Shilin and Wang, Fengxiang and Zhang, Yi-Fan and Wang, Haotian and Zhang, Yuanxing and Wan, Pengfei},
  journal={arXiv preprint arXiv:2605.18984},
  year={2026}
}

@article{liu2026longav,
  title={LongAV-Compass: Towards Unified Evaluation of Minute-Scale Audio-Visual Generation Across T2AV, I2AV, and V2AV},
  author={Liu, Tengfei and Shi, Yang and Zhu, Xuanyu and Tang, Jiafu and Yang, Liu and Wang, Qixun and Zhang, Zhuoran and Tang, Yuqi and Wang, Fengxiang and Dong, Yuhao and others},
  journal={arXiv preprint arXiv:2605.26244},
  year={2026}
}

@article{xu2025qwen3omni,
  title   = {Qwen3-Omni Technical Report},
  author  = {Jin Xu and Zhifang Guo and Hangrui Hu and Yunfei Chu and Xiong Wang and Jinzheng He and Yuxuan Wang and Xian Shi and Ting He and Xinfa Zhu and Yuanjun Lv and Yongqi Wang and Dake Guo and He Wang and Linhan Ma and Pei Zhang and Xinyu Zhang and Hongkun Hao and Zishan Guo and Baosong Yang and Bin Zhang and Ziyang Ma and Xipin Wei and Shuai Bai and Keqin Chen and Xuejing Liu and Peng Wang and Mingkun Yang and Dayiheng Liu and Xingzhang Ren and Bo Zheng and Rui Men and Fan Zhou and Bowen Yu and Jianxin Yang and Le Yu and Jingren Zhou and Junyang Lin},
  journal = {arXiv preprint arXiv:2509.17765},
  year    = {2025},
  url     = {https://arxiv.org/abs/2509.17765}
}

@misc{qwen2026qwen35,
  title  = {{Qwen3.5}: Towards Native Multimodal Agents},
  author = {{Qwen Team}},
  month  = {February},
  year   = {2026},
  url    = {https://qwen.ai/blog?id=qwen3.5}
}

@article{li2025granted,
  title={{GRAN-TED}: Generating Robust, Aligned, and Nuanced Text Embedding for Diffusion Models},
  author={Li, Bozhou and Yang, Sihan and Guan, Yushuo and An, Ruichuan and Chen, Xinlong and Shi, Yang and Wan, Pengfei and Zhang, Wentao and others},
  journal={arXiv preprint arXiv:2512.15560},
  year={2025}
}

@article{li2026semanticrouting,
  title={Semantic Routing: Exploring Multi-Layer {LLM} Feature Weighting for Diffusion Transformers},
  author={Li, Bozhou and Guan, Yushuo and Li, Haolin and Zeng, Bohan and Ji, Yiyan and Ding, Yue and Wan, Pengfei and Gai, Kun and Zhang, Yuanxing and Zhang, Wentao},
  journal={arXiv preprint arXiv:2602.03510},
  year={2026}
}

@article{liu2026rever,
  title={Rubrics from the Task, Verdicts from Evidence: Pointwise Rubric Rewards with Tool-Augmented Judges},
  author={Liu, Tianci and Wang, Haoyu},
  journal={arXiv preprint arXiv:2610.18766},
  year={2026}
}

@article{lin2025forensicchat,
  title={Seeing Before Reasoning: A Unified Framework for Generalizable and Explainable Fake Image Detection},
  author={Lin, Kaiqing and Yan, Zhiyuan and Chen, Ruoxin and Ye, Junyan and Zhang, Ke-Yue and Zhou, Yue and Jin, Peng and Li, Bin and Yao, Taiping and Ding, Shouhong},
  journal={arXiv preprint arXiv:2509.25502},
  year={2025}
}

@article{he2024videoscore,
  title={{VideoScore}: Building Automatic Metrics to Simulate Fine-Grained Human Feedback for Video Generation},
  author={He, Xuan and Jiang, Dongfu and Zhang, Ge and Ku, Max and Soni, Achint and Siu, Sherman and Chen, Haonan and Chandra, Abhranil and Jiang, Ziyan and Arulraj, Aaran and others},
  journal={arXiv preprint arXiv:2406.15252},
  year={2024}
}

@article{wang2025unifiedreward,
  title={Unified Multimodal Chain-of-Thought Reward Model through Reinforcement Fine-Tuning},
  author={Wang, Yibin and Li, Zhimin and Zang, Yuhang and Wang, Chunyu and Lu, Qinglin and Jin, Cheng and Wang, Jiaqi},
  journal={arXiv preprint arXiv:2505.03318},
  year={2025}
}

@inproceedings{wu2023dover,
  title={Exploring Video Quality Assessment on User Generated Contents from Aesthetic and Technical Perspectives},
  author={Wu, Haoning and Zhang, Erli and Liao, Liang and Chen, Chaofeng and Hou, Jingwen and Wang, Annan and Sun, Wenxiu and Yan, Qiong and Lin, Weisi},
  booktitle={Proceedings of the IEEE/CVF International Conference on Computer Vision},
  pages={20144--20154},
  year={2023}
}

@article{team2025kling,
  title={Kling-omni technical report},
  author={Team, Kling and Chen, Jialu and Ci, Yuanzheng and Du, Xiangyu and Feng, Zipeng and Gai, Kun and Guo, Sainan and Han, Feng and He, Jingbin and He, Kang and others},
  journal={arXiv preprint arXiv:2512.16776},
  year={2025}
}

@article{team2025klingavatar,
  title={Klingavatar 2.0 technical report},
  author={Team, Kling and Chen, Jialu and Ding, Yikang and Fang, Zhixue and Gai, Kun and Gao, Yuan and He, Kang and Hua, Jingyun and Jiang, Boyuan and Lao, Mingming and others},
  journal={arXiv preprint arXiv:2512.13313},
  year={2025}
}

@article{team2026kling,
  title={Kling-MotionControl Technical Report},
  author={Team, Kling and Chen, Jialu and Ding, Yikang and Fang, Zhixue and Gai, Kun and He, Kang and He, Xu and Hua, Jingyun and Lao, Mingming and Li, Xiaohan and others},
  journal={arXiv preprint arXiv:2603.03160},
  year={2026}
}

@misc{googledeepmind2026gemini31pro,
  title  = {{Gemini 3.1 Pro}},
  author = {{Google DeepMind}},
  year   = {2026},
  url    = {https://deepmind.google/models/gemini/pro/},
  note   = {Accessed October 23, 2026}
}

@inproceedings{panayotov2015librispeech,
  title     = {{LibriSpeech}: An {ASR} Corpus Based on Public Domain Audio Books},
  author    = {Vassil Panayotov and Guoguo Chen and Daniel Povey and Sanjeev Khudanpur},
  booktitle = {2015 IEEE International Conference on Acoustics, Speech and Signal Processing},
  pages     = {5206--5210},
  year      = {2015},
  doi       = {10.1109/ICASSP.2015.7178964}
}

@article{hong2026worldsenseevaluatingrealworldomnimodal,
  title   = {{WorldSense}: Evaluating Real-world Omnimodal Understanding for Multimodal {LLMs}},
  author  = {Jack Hong and Shilin Yan and Jiayin Cai and Xiaolong Jiang and Yao Hu and Weidi Xie},
  journal = {arXiv preprint arXiv:2502.04326},
  year    = {2026}
}

@article{shao2026listeningtimeprecisetemporal,
  title   = {Listening with Time: Precise Temporal Awareness for Long-Form Audio Understanding},
  author  = {Mingchen Shao and Hang Su and Wenjie Tian and Bingshen Mu and Zhennan Lin and Lichun Fan and Zhenbo Luo and Jian Luan and Lei Xie},
  journal = {arXiv preprint arXiv:2604.22245},
  year    = {2026}
}

@article{li2025omnibenchfutureuniversalomnilanguage,
  title   = {{OmniBench}: Towards the Future of Universal Omni-Language Models},
  author  = {Yizhi Li and Yinghao Ma and Ge Zhang and Ruibin Yuan and Kang Zhu and Hangyu Guo and others},
  journal = {arXiv preprint arXiv:2409.15272},
  year    = {2025}
}

@article{sun2024videosalmonnspeechenhancedaudiovisuallarge,
  title   = {{video-SALMONN}: Speech-Enhanced Audio-Visual Large Language Models},
  author  = {Guangzhi Sun and Wenyi Yu and Changli Tang and Xianzhao Chen and Tian Tan and Wei Li and others},
  journal = {arXiv preprint arXiv:2406.15704},
  year    = {2024}
}

@article{wu2025ugcvideocaptioneromniugcvideo,
  title   = {{UGC-VideoCaptioner}: An Omni {UGC} Video Detail Caption Model and New Benchmarks},
  author  = {Peiran Wu and Yunze Liu and Zhengdong Zhu and Enmin Zhou and Junxiao Shen},
  journal = {arXiv preprint arXiv:2507.11336},
  year    = {2025}
}

@article{wang2026omnicapifbenchmarkingimprovinginstruction,
  title   = {{OmniCap-IF}: Benchmarking and Improving Instruction Following Abilities for Omni-Video Captioning},
  author  = {Jiahao Wang and An Ping and Yanghai Wang and Yuanxing Zhang and Shihao Li and Hanyan Bian and others},
  journal = {arXiv preprint arXiv:2606.08572},
  year    = {2026}
}

@article{ma2025ivbenchbenchmarkimagegroundedvideo,
  title   = {{IV-Bench}: A Benchmark for Image-Grounded Video Perception and Reasoning in Multimodal {LLMs}},
  author  = {David Ma and Yuanxing Zhang and Jincheng Ren and Jarvis Guo and Yifan Yao and Zhenlin Wei and others},
  journal = {arXiv preprint arXiv:2504.15415},
  year    = {2025}
}

@article{peng2025mvuevalmultivideounderstandingevaluation,
  title   = {{MVU-Eval}: Towards Multi-Video Understanding Evaluation for Multimodal {LLMs}},
  author  = {Tianhao Peng and Haochen Wang and Yuanxing Zhang and Zekun Wang and Zili Wang and Ge Zhang and others},
  journal = {arXiv preprint arXiv:2511.07250},
  year    = {2025}
}

@article{liu2024mmbenchmultimodalmodelallaround,
  title   = {{MMBench}: Is Your Multi-modal Model an All-around Player?},
  author  = {Yuan Liu and Haodong Duan and Yuanhan Zhang and Bo Li and Songyang Zhang and Wangbo Zhao and others},
  journal = {arXiv preprint arXiv:2307.06281},
  year    = {2024}
}

@article{liu2024ocrbenchhiddenmysteryocr,
  title   = {{OCRBench}: On the Hidden Mystery of {OCR} in Large Multimodal Models},
  author  = {Yuliang Liu and Zhang Li and Mingxin Huang and Biao Yang and Wenwen Yu and Chunyuan Li and others},
  journal = {Science China Information Sciences},
  year    = {2024},
  doi     = {10.1007/s11432-024-4235-6}
}

@article{zhu2026textpeckerrewardingstructuralanomaly,
  title   = {{TextPecker}: Rewarding Structural Anomaly Quantification for Enhancing Visual Text Rendering},
  author  = {Hanshen Zhu and Yuliang Liu and Xuecheng Wu and An-Lan Wang and Hao Feng and Dingkang Yang and others},
  journal = {arXiv preprint arXiv:2602.20903},
  year    = {2026}
}

@article{lin2026perceptionbenchevaluatingatomicvisual,
  title   = {{PerceptionBench}: Evaluating Atomic Visual Perception in Multimodal Large Language Models},
  author  = {Zichao Lin and Yifeng Xie and Bowen Qu and Haiming Wang and Jia Li and Haoning Wu and others},
  journal = {arXiv preprint arXiv:2607.24957},
  year    = {2026}
}

@article{wu2024longvideobenchbenchmarklongcontextinterleaved,
  title   = {{LongVideoBench}: A Benchmark for Long-context Interleaved Video-Language Understanding},
  author  = {Haoning Wu and Dongxu Li and Bei Chen and Junnan Li},
  journal = {arXiv preprint arXiv:2407.15754},
  year    = {2024}
}

@article{shi2026mmevideoocrevaluatingocrbasedcapabilities,
  title   = {{MME-VideoOCR}: Evaluating {OCR}-Based Capabilities of Multimodal {LLMs} in Video Scenarios},
  author  = {Yang Shi and Huanqian Wang and Wulin Xie and Huanyao Zhang and Lijie Zhao and Yi-Fan Zhang and others},
  journal = {arXiv preprint arXiv:2505.21333},
  year    = {2026}
}

@article{yang2025thinkingspacemultimodallarge,
  title   = {Thinking in Space: How Multimodal Large Language Models See, Remember, and Recall Spaces},
  author  = {Jihan Yang and Shusheng Yang and Anjali W. Gupta and Rilyn Han and Li Fei-Fei and Saining Xie},
  journal = {arXiv preprint arXiv:2412.14171},
  year    = {2025}
}

@article{ge2025archunyuanvideo7bstructuredvideocomprehension,
  title   = {{ARC-Hunyuan-Video-7B}: Structured Video Comprehension of Real-World Shorts},
  author  = {Yuying Ge and Yixiao Ge and Chen Li and Teng Wang and Junfu Pu and Yizhuo Li and others},
  journal = {arXiv preprint arXiv:2507.20939},
  year    = {2025}
}

@article{zhang2025dsibenchbenchmarkdynamicspatial,
  title   = {{DSI-Bench}: A Benchmark for Dynamic Spatial Intelligence},
  author  = {Ziang Zhang and Zehan Wang and Guanghao Zhang and Weilong Dai and Yan Xia and Ziang Yan and others},
  journal = {arXiv preprint arXiv:2510.18873},
  year    = {2025}
}

@article{du2024eventorientedlongvideounderstanding,
  title   = {Towards Event-oriented Long Video Understanding},
  author  = {Yifan Du and Kun Zhou and Yuqi Huo and Yifan Li and Wayne Xin Zhao and Haoyu Lu and others},
  journal = {arXiv preprint arXiv:2406.14129},
  year    = {2024}
}

@article{hong2026motionbenchbenchmarkingimprovingfinegrained,
  title   = {{MotionBench}: Benchmarking and Improving Fine-grained Video Motion Understanding for Vision Language Models},
  author  = {Wenyi Hong and Yean Cheng and Zhuoyi Yang and Weihan Wang and Lefan Wang and Xiaotao Gu and others},
  journal = {arXiv preprint arXiv:2501.02955},
  year    = {2026}
}

@article{zhang2026timelensrethinkingvideotemporal,
  title   = {{TimeLens}: Rethinking Video Temporal Grounding with Multimodal {LLMs}},
  author  = {Jun Zhang and Teng Wang and Yuying Ge and Yixiao Ge and Xinhao Li and Ying Shan and Limin Wang},
  journal = {arXiv preprint arXiv:2512.14698},
  year    = {2026}
}

@article{feng2026geometryguidedcameramotionunderstanding,
  title   = {Geometry-Guided Camera Motion Understanding in {VideoLLMs}},
  author  = {Haoan Feng and Sri Harsha Musunuri and Guan-Ming Su},
  journal = {arXiv preprint arXiv:2603.13119},
  year    = {2026}
}

@inproceedings{sigurdsson2016hollywoodhomescrowdsourcingdata,
  title     = {Hollywood in Homes: Crowdsourcing Data Collection for Activity Understanding},
  author    = {Gunnar A. Sigurdsson and G{\"u}l Varol and Xiaolong Wang and Ali Farhadi and Ivan Laptev and Abhinav Gupta},
  booktitle = {European Conference on Computer Vision},
  year      = {2016}
}

@inproceedings{lei2021qvhighlightsdetectingmomentshighlights,
  title     = {{QVHighlights}: Detecting Moments and Highlights in Videos via Natural Language Queries},
  author    = {Jie Lei and Tamara L. Berg and Mohit Bansal},
  booktitle = {Advances in Neural Information Processing Systems},
  year      = {2021}
}

@inproceedings{krishna2017densecaptioningeventsvideos,
  title     = {Dense-Captioning Events in Videos},
  author    = {Ranjay Krishna and Kenji Hata and Frederic Ren and Li Fei-Fei and Juan Carlos Niebles},
  booktitle = {Proceedings of the IEEE International Conference on Computer Vision},
  year      = {2017}
}

@article{bhattacharya2025benchmarking,
  title={Benchmarking and Confidence Evaluation of LALMs For Temporal Reasoning},
  author={Bhattacharya, Debarpan and Kulkarni, Apoorva and Ganapathy, Sriram},
  journal={arXiv preprint arXiv:2505.13115},
  year={2025}
}

@article{gurnee2023language,
  title={Language Models Represent Space and Time},
  author={Gurnee, Wes and Tegmark, Max},
  journal={arXiv preprint arXiv:2310.02207},
  year={2023}
}

@article{bradley1952rank,
  title={Rank Analysis of Incomplete Block Designs: I. The Method of Paired Comparisons},
  author={Bradley, Ralph Allan and Terry, Milton E.},
  journal={Biometrika},
  volume={39},
  number={3/4},
  pages={324--345},
  year={1952}
}

@misc{internlm2026interndecision,
  author       = {{InternLM Team}},
  title        = {Intern-Decision: Typed Decisions with Probabilities from a Single Forward Pass},
  year         = {2026},
  howpublished = {\url{https://huggingface.co/collections/internlm/intern-decision}},
  note         = {Model weights, training and inference code. Accessed 2026-11-01},
}

@article{liu2026refcaptioner,
  title={RefCaptioner: Multi-Reference Image-Grounded Video Captioning},
  author={Liu, Tengfei and Shi, Yang and Wang, Yuran and Zhang, Xiaohan and Wen, Yuqing and Tang, Yuqi and Wang, Qixun and Zhang, Zhuoran and Zhu, Xuanyu and Lin, Weihong and others},
  journal={arXiv preprint arXiv:2607.28509},
  year={2026}
}

@article{zhu2026wanpe,
  title={WanPE: Towards Cinematic Prompt Enhancement for Modern Text-to-Video Generation},
  author={Zhu, Yubo and Shao, Yawen and Dai, Ziyun and Fang, Zixun and Zhu, Kai and Sun, Siyang and Xue, Haolan and Wang, Chuxin and Weng, Tingyu and Luo, Jingming and others},
  journal={arXiv preprint arXiv:2609.30221},
  year={2026}
}

\section*{Contributors}
\addcontentsline{toc}{section}{Contributors}

Names within each group are listed in alphabetical order by surname.

\medskip
\noindent\textbf{Contributors}

\smallskip
\noindent
Dong Chen, Jialu Chen, Xinlong Chen, Yifan Dai, Haotian Dong,
Binzong Geng, Yushuo Guan, Minghui Guo, Guanlin He, Guanxiao He, Jingyun Hua,
Yanwei Huang, Yiyan Ji, Xueping Jia, Yizhuo Jia, Bozhou Li, Jiaze Li, Min Li,
Shihao Li, Shujie Li, Shengyi Liao, Weihong Lin, Shiyuan Liu, Xinyu Liu,
Siyang Mao, Xiaoru Qu, Dianmo Sheng, Yang Shi, Jiafu Tang,
Longxiang Tang, Yuqi Tang, Chengzhuo Tong, Kuo Wang, Qixun Wang, Rui Wang,
Tianduo Wang, Yuran Wang, Zekun Wang, Zihao Wang, Boyang Xia,
Yifan Xu, Liu Yang, Zhenlong Yuan, Bohan Zeng, Mujin Zhang,
Yuanxing Zhang\footnotemark[2], Zhicheng Zhang, Zhuoran Zhang, Yuanhong Zheng
\footnotetext[2]{Project Lead.}

\medskip
\noindent\textbf{Project Sponsors}

\smallskip
\noindent
Kun Gai, Chengru Song, Pengfei Wan

\clearpage
\newpage

\end{document}